\documentclass{article}

\usepackage{arxiv}

\usepackage[utf8]{inputenc} 
\usepackage[T1]{fontenc}    
\usepackage{hyperref}       
\usepackage{url}            
\usepackage{booktabs}       
\usepackage{amsfonts}       
\usepackage{nicefrac}       
\usepackage{microtype}      
\usepackage{lipsum}
\usepackage{graphicx}
\graphicspath{ {./images/} }

\usepackage[utf8]{inputenc} 
\usepackage[T1]{fontenc}    
\usepackage{hyperref}       
\usepackage{url}            
\usepackage{booktabs}       
\usepackage{amsfonts}       
\usepackage{nicefrac}       
\usepackage{microtype}      
\usepackage{xcolor}         

\usepackage{subfigure}

\usepackage{bm}
\usepackage{graphicx}
\usepackage{mathptmx}      
\usepackage[T1]{fontenc}
\usepackage[utf8]{inputenc}

\usepackage{amsmath,amsfonts,amsthm}
\usepackage{amssymb}
\usepackage{systeme}
\usepackage{longtable,tabularx}
\usepackage{graphicx}
\usepackage{caption}
\usepackage{floatrow}   
\usepackage{graphicx,xcolor} 
\usepackage[framemethod=tikz]{mdframed}
\usepackage{caption}

\usepackage{algorithm}%
\usepackage{algorithmic}%

\floatname{algorithm}{Algorithm}

\usepackage{xcolor}
\def\mathbi#1{\textbf{\em #1}}
\usepackage{caption}
\usepackage{sidecap}
\usepackage{breqn}

\usepackage[square,numbers]{natbib}

\usepackage{floatrow}
\def\mathbi#1{\textbf{\em #1}}
\newcommand{\DesignVar}{\mathbi{x}}

\newcommand{\DesignSpace}{\chi}
\newcommand{\FidSpace}{\mathcal{L}}
\newcommand{\Dataset}{\mathcal{D}}
\newcommand{\ObjFun}{f}
\newcommand{\MinObjFun}{f^{*}}
\newcommand{\MinDesignVar}{x^{*}}
\newcommand{\LevFid}{l}
\newcommand{\MaxLevFid}{L}
\newcommand{\Dim}{D}
\newcommand{\IterOpt}{i}
\newcommand{\NoisyObserv}{y}
\newcommand{\IndNumObs}{n} 
\newcommand{\NumObs}{N} 

\newcommand{\StandDevNoise}{\sigma_{\epsilon}}
\newcommand{\NorDist}{\mathcal{N}}
\newcommand{\MeanFunGP}{\mu}
\newcommand{\CovFunGP}{\kappa}

\newcommand{\StandDevGP}{\sigma}

\newcommand{\ConsFact}{\varrho}
\newcommand{\Discrep}{\zeta}
\newcommand{\AF}{U}
\newcommand{\EI}{EI}
\newcommand{\PI}{PI}
\newcommand{\MFEI}{MFEI}
\newcommand{\MFPI}{MFPI}
\newcommand{\MFMES}{MFMES}
\newcommand{\MFES}{MFES}
\newcommand{\Improv}{I}
\newcommand{\CompCost}{\lambda}

\newcommand{\CumDistFun}{\Phi}
\newcommand{\ProbDensFun}{\phi}
\newcommand{\Budget}{B}

\newcommand{\Expectation}{\mathbb{E}}

\newcommand{\ES}{ES}
\newcommand{\MES}{MES}
\newcommand{\Entropy}{H}
\newcommand{\Probability}{p}
\newcommand{\Integral}{I}
\newcommand{\DimFun}{d_\DesignVar}
\newcommand{\Distribution}{\Tilde{\pi}}
\newcommand{\TargetDistribution}{\pi}
\newcommand{\ProposalPDF}{q}
\newcommand{\NumSamples}{N}
\newcommand{\IndNumSamples}{n}
\newcommand{\ImpWeight}{w}
\newcommand{\ParamVector}{\theta}
\newcommand{\IndSimSamples}{k}
\newcommand{\SimSamples}{K}
\newcommand{\Dirac}{\delta}
\newcommand{\points}{\DesignVar}
\newcommand{\dimpoints}{N}
\newcommand{\NumPoints}{P}
\newcommand{\RandSpace}{\Gamma}
\newcommand{\pdf}{\rho}
\newcommand{\IntPolynomial}{\mathcal{I}}

\newcommand{\IntRules}{\mathcal{U}}
\newcommand{\NumNodes}{n}
\newcommand{\NodeLevel}{i}
\newcommand{\NodalSet}{\vartheta}
\newcommand{\LagrangePol}{\mathcal{L}}
\newcommand{\SparsenessParam}{q}
\newcommand{\Grid}{\mathcal{A}}
\newcommand{\Nodes}{\mathcal{H}}

\newcommand{\ResponseVec}{\mathbi{R}}

\title{Active Learning and Bayesian Optimization: a Unified Perspective to Learn with a Goal 

}

\author{
 Francesco Di Fiore \\
  Politecnico di Torino\\
  \texttt{francesco.difiore@polito.it} \\
   \And
  Michela Nardelli \\
  Politecnico di Torino\\
  \texttt{michela.nardelli@polito.it} \\
   \And
 Laura Mainini \\
  Imperial College London\\
  Politecnico di Torino\\
  Massachusetts Institute of Technology\\
  \texttt{l.mainini@imperial.ac.uk} \\
}

\begin{document}

\maketitle


\begin{abstract}
Science and Engineering applications are typically associated with expensive optimization problems to identify optimal design solutions and states of the system of interest. Bayesian optimization and active learning compute surrogate models through efficient adaptive sampling schemes to assist and accelerate this search task toward a given optimization goal. Both those methodologies are driven by specific infill/learning criteria which quantify the utility with respect to the set goal of evaluating the objective function for unknown combinations of optimization variables. While the two fields have seen an exponential growth in popularity in the past decades, their dualism and synergy have received relatively little attention to date. This paper discusses and formalizes the synergy between Bayesian optimization and active learning as symbiotic adaptive sampling methodologies driven by common principles. In particular, we demonstrate this unified perspective through the formalization of the analogy between the Bayesian infill criteria and active learning criteria as driving principles of both the goal-driven procedures. To support our original perspective, we propose a general classification of adaptive sampling techniques to highlight similarities and differences between the vast families of adaptive sampling, active learning, and Bayesian optimization. Accordingly, the synergy is demonstrated mapping the Bayesian infill criteria with the active learning criteria, and is formalized for searches informed by both a single information source and multiple levels of fidelity. In addition, we provide guidelines to apply those learning criteria investigating the performance of different Bayesian schemes for a variety of benchmark problems to highlight benefits and limitations over mathematical properties that characterize real-world applications.
\end{abstract}

\let\thefootnote\relax\footnotetext{Published Research Paper: Di Fiore, Francesco, Michela Nardelli, and Laura Mainini. "Active learning and bayesian optimization: a unified perspective to learn with a goal." Archives of Computational Methods in Engineering (2024): 1-29. https://doi.org/10.1007/s11831-024-10064-z}

\section{Introduction}  \label{s:intro}

In science and engineering, the development of advanced technologies involves the formalization and solution of optimization problems to identify both optimal designs capable of satisfying competing requirements of performance \cite{martins2021engineering}, and states of the system to monitor their health status during the operational life \cite{kim2017prognostics}. Depending on the specific application, the identification of optimal solutions requires the minimization of an objective function that measures the goodness of design configurations with respect to the requirements, or the accuracy of the estimated health status of the system with respect to the measurements. Typically, the scale of complexity of engineering systems requires several evaluations of this objective function through accurate computer simulations -- e.g. Computational Fluid Dynamics (CFD) or Computational Structural Dynamics (CSD) -- or physical experiments -- e.g. lab-scale test benches or real-world testing -- before assessing an optimal solution. The use of highly complicate representations of those systems leads to a significant bottleneck: the demand for resources to evaluate the objective function for all the combinations of optimization variables is difficult to be adequately satisfied. Indeed, the acquisition of data from these high-fidelity models involves huge non-trivial computational and economical costs that could arise from the computation of the objective function and its derivatives over ideally the entire optimization domain.

Surrogate models are computed on evaluations of the objective function acquired through computer codes and/or physical experiments of the system: these sources of information are mostly treated as purely input/output black-box relationships whose analytical form is unknown and not directly accessible to the optimizer. Thus, the accuracy and efficiency of the resulting surrogate are highly dependent on the sampling approach adopted to select informative combinations of optimization variables for the acquisition of data. Among the numerous sampling schemes available in literature, it is possible to identify two major families: one-shot, and sequential schemes. The one-shot strategy defines a grid of samples over the domain all at once. Examples include Latin Hypercube \cite{mckay2000comparison}, factorial and fractional factorials designs \cite{gunst2009fractional, montgomery2017design}, Placket-Burmann \cite{gustafsson2013conjoint}, and D-optimal \cite{myers2016response}. However, it is very hard to identify a priori the best design of those experiments to efficiently compute the most informative surrogate. To overcome these limitations, sequential sampling selects samples over the domain through an iterative process \cite{chernoff1959sequential, jin2002sequential}. Among these, adaptive sampling \cite{provost1999efficient} provides resource-efficient techniques that seek to reduce as much as possible the evaluations of the objective function, and targets the improvement of the fitting quality across the domain and/or the acceleration of the optimization search \cite{viana2014special, liu2018survey, dias2019adaptive}. Popular adaptive samplings to address black-box optimization problems characterized by the expensive evaluation of the objective function are those realized through the Bayesian Optimization (BO) methodology \cite{shahriari2015taking, frazier2018tutorial}. BO aims at efficiently elicit valuable data from models of the system to contain the computational expense of the optimization procedure.The Bayesian routine iteratively computes a surrogate model of the objective function, and defines a goal-driven sampling process through an acquisition function computed on the surrogate information. This acquisition function measures the merit of samples according to certain infill criteria, and permits to select the next sample that maximizes the query utility with respect to the given optimization goal.

The popular paradigms for Bayesian optimization show substantial synergy with active learning schemes which has not been explicitly discussed and formally described in literature to date. This paper proposes the explicit formalization of this synergy through an original perspective of Bayesian optimization and active learning as symbiotic expressions of adaptive sampling schemes. The aim of this unifying viewpoint is to support the use of those methodologies, and point out and discuss the analogies via their mathematical formalization. This unified interpretation is based on the formulation and demonstration of the analogy between the Bayesian infill criteria and the active learning criteria as the elements responsible for the decision on how to learn from samples to reach the given goal. In support of this unified perspective, this paper first clarifies the concept of goal-driven learning, and proposes a general classification of adaptive sampling methods that recognizes Bayesian optimization and active learning as methodologies characterized by goal-oriented search schemes. Thus, we elucidate the synergy between Bayesian optimization and active learning mapping the Bayesian learning features on the active learning properties. The mapping is discussed through the analysis of three popular Bayesian frameworks for both the case of a single information source, and when a spectrum of multiple sources are available to the search. In addition, we observe the capabilities introduced by the different learning criteria over a comprehensive set of benchmark problems specifically defined to stress test and validate goal-driven approaches \cite{mainini2022analytical}. The objective is to discuss opportunities and limitations of different learning principles over a variety of challenging mathematical properties of optimization problems frequently encountered in complex scientific and engineering applications.

This manuscript is organized as follows. Section \ref{s:GoalDriven_ProbStatem} discusses goal-driven learning procedures and defines the concept of goal-driven learner according to surrogate modeling and optimization. In Section \ref{s: AS_Background}, we recognize that Bayesian optimization, active learning and adaptive sampling are not fully superimposable concepts, and propose a general classification to position Bayesian optimization and active learning with respect to the adaptive sampling methodologies. Then, Section \ref{s:BOFrameworks} provides an overview on Bayesian optimization and multifidelity Bayesian optimization. Section \ref{s:An Active Learning Perspective} presents our perspective on the symbiotic relationship between Bayesian optimization and active learning. Then, in Section \ref{s:Experiments} popular Bayesian optimization and multifidelity Bayesian optimization algorithms are numerically investigated over a variety of benchmark problems. Finally, Section \ref{s:concluding remarks} provides concluding remarks.

\section{Goal-Driven Learning}
\label{s:GoalDriven_ProbStatem}

Goal-driven learning is a decision-making process in which each decision is made to acquire specific information about the system of interest that contribute the most to achieve a given goal \cite{ram1995goal,oden2000estimation, bui2007goal, lieberman2013goal, di2022non, grassi2023raal}. This learning goal can be the increase of the knowledge of the system behaviour over all the domain of application, or the acquisition of specific knowledge to enhance and accelerate the identification of optimization solutions. Accordingly, a goal-driven learner selects what to learn considering both the current knowledge and information needed, and determines how to learn quantifying the relative utility of alternative options in the current circumstances. 

This paper focuses on Bayesian optimization and active learning as goal-driven procedures where a surrogate model is built to accurately represent the behaviour of a system or effectively inform an optimization procedure to minimize given objectives. This goal-driven process is guided by learning principles that determine the "best" location of the domain to acquire information about the system, and refine the surrogate model toward the goal -- improve the accuracy of the surrogate or minimize an objective function over the domain. Formally, these surrogate based modeling and optimization problems can be formulated as a minimization problem of the following form: 

\begin{equation} \label{e: ProbStatem}
\MinDesignVar = \text{arg} \min_{\DesignVar \in \DesignSpace} \ObjFun(\ResponseVec(\DesignVar))
\end{equation}

\noindent where $\ObjFun(\ResponseVec(\DesignVar))$ denotes the objective function evaluated at the location $\DesignVar \in \DesignSpace$ of the domain $\DesignSpace$. The objective function is of the general form $\ObjFun = \ObjFun(\ResponseVec(\DesignVar))$, where $\ResponseVec(\DesignVar)$ represents the response of the system of interest evaluated through a model -- e.g. computer-based numerical simulations or real-world experiments. In surrogate based modeling, the objective function can be represented as the error between the approximation of the surrogate model and the response of the system: the goal is to minimize such error to improve the accuracy of the surrogate over all the domain. In surrogate based optimization, the objective function represents a performance indicator dependent on the system response: the goal is to minimize this indicator to improve the capabilities of the system according to given performance requirements. Goal-driven techniques address Equation \eqref{e: ProbStatem} through a decision-making iterative process where learning principles tailor the acquisition of specific knowledge about the objective function -- evaluation of $\ObjFun$ at certain domain location $\DesignVar$ -- currently needed to update the surrogate and inform the learner toward the given goal.

In this context, the goal-driven learner is the agent that makes decisions based on the current knowledge of the system of interest, and acquires new information to accomplish a given goal while augmenting the awareness about the system itself. In practice, the learner queries the sample that maximizes the utility to achieve the desired goal: specific learning principles quantify this utility based on the surrogate estimate and in response to information needs. At the same time, the surrogate model is dynamically updated once new information are acquired, and informs the learner to focus and tailor on the fly the elicitation of samples to further overarching the goal. Thus, the distinguishing element of a goal-driven learning procedure is represented by the mutual exchange of information between the learner and the surrogate model: the learner assimilates the information from the surrogate to make a decision aimed at achieving the goal, and the approximation/prediction of the surrogate is enriched by the result of this decision.

\section{Adaptive Sampling Classification} \label{s: AS_Background}

\begin{figure}[!t]
\centering
\includegraphics[width=0.45\linewidth,trim=220 0 245 0, clip]{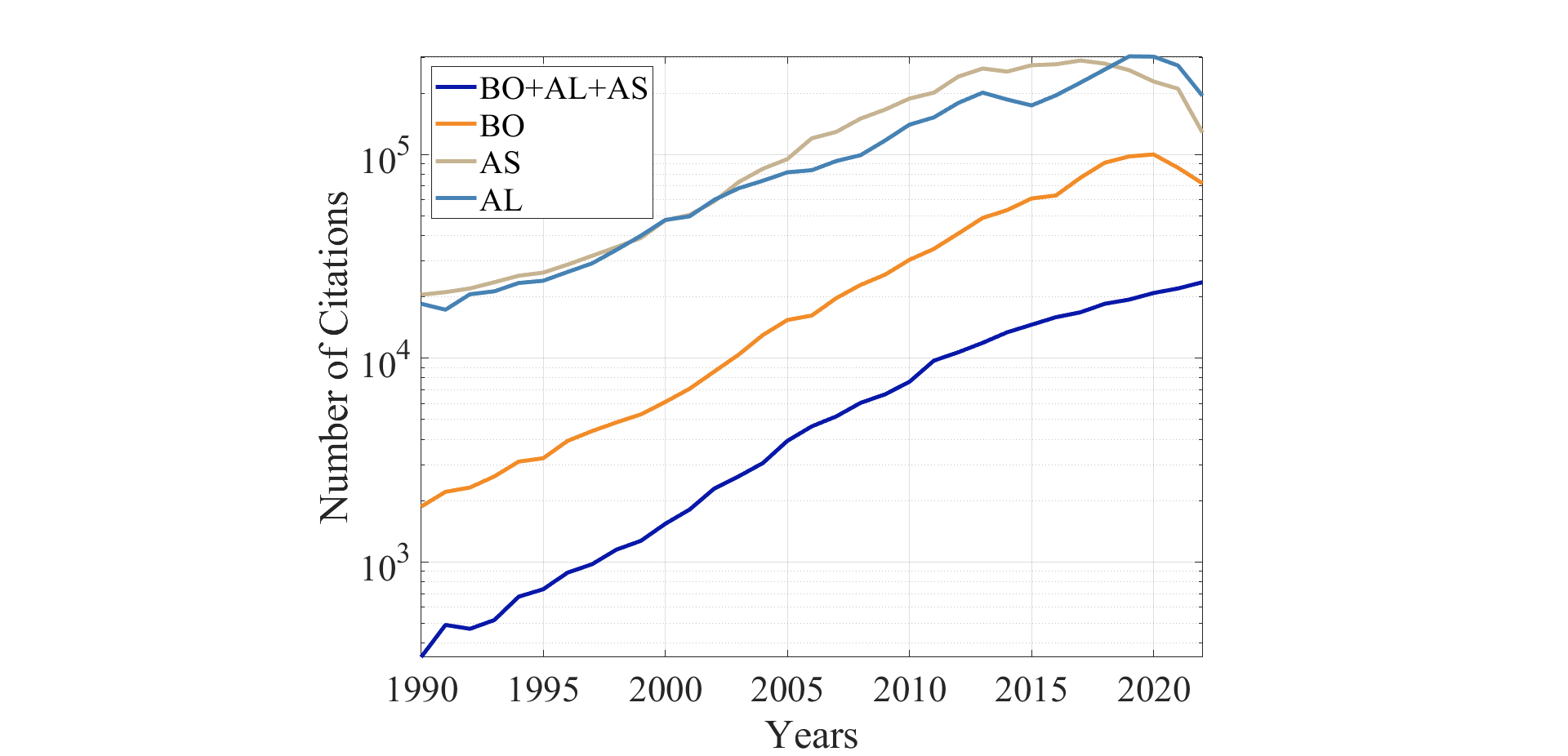}
\caption{Citations of Bayesian Optimization (BO), Active Learning (AL), Adaptive Sampling (AS) and the three terms combined (BO+AL+AS).}

\label{fig:PaperCitations} 
\end{figure}

Bayesian optimization and active learning realize adaptive sampling schemes to efficiently accomplish a given goal while adapting to the previously collected information. In recent years, there has been a profusion of literature devoted to the general topic of adaptive sampling but arguably a blurring of focus: many contributions from different field provided a deal of interesting advancements, but also led to some degree of confusion around the concepts of adaptive sampling, active learning and Bayesian optimization. Figure \ref{fig:PaperCitations} illustrates the use of the words "adaptive sampling", "active learning", and "Bayesian optimization" from 1990 to 2022. In addition, we report the combined use of all the three words over the same period of time. It can be appreciated both the general increasing trend of use of the three techniques and the associated increase of the use of the three terms combined. Many times the three concepts have been used as complete synonyms, with some growing abuse motivated by the difficulties to map the (shaded) boundaries. 

\begin{figure*}[!t]
\centering
\includegraphics[width=0.65\textwidth]{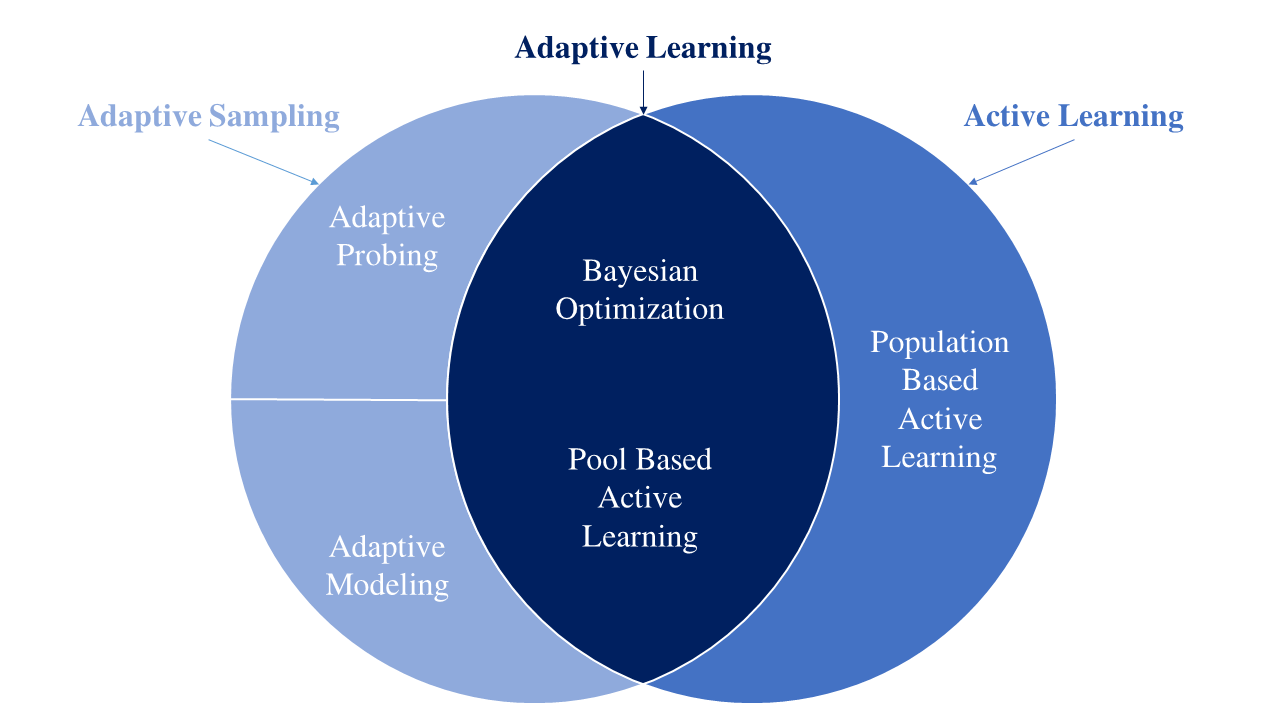}
\caption{Where adaptive sampling and active learning meet: this work focuses on the synergies between Bayesian optimization and active learning as goal-driven learning procedures driven by common learning principles.}
\label{fig:venn} 
\end{figure*}

Stemming from these considerations, this paper recognizes that adaptive sampling is not always superimposable with active learning and Bayesian optimization. Figure \ref{fig:venn} illustrates the relationships between those three methodologies. We propose a classification of adaptive sampling techniques in three main families, namely adaptive probing (Section \ref{s: adaptProb}), adaptive modeling (Section \ref{s: adaptModel}) and adaptive learning (Section \ref{s: adaptLear}). This classification is based on the concept of goal-driven learning as the distinctive element of adaptive learning methodologies: the learner assimilates the information from the surrogate model to make a decision aimed at achieving a goal, and the surrogate is enriched by the result of this decision following a mutual exchange of information. Conversely, adaptive probing and adaptive modeling classes do not realize a goal-driven learning: the former does not rely on a surrogate model to assist the sampling procedure while the latter computes a surrogate model that is not used to inform the search task. This classification permits to clarify the reciprocal positions between adaptive sampling, active learning and Bayesian optimization. 

Accordingly, adaptive sampling and active learning do not completely overlap. Active learning strategies are categorized into population-based and pool-based algorithms according to the nature of the search procedure \cite{Sugiyama&Nakajima2009, Wu2018}. In population-based active learning, the distribution of the objective function is available: the learner seeks to determine the optimal training input density to generate training points without relying on a surrogate model of the objective function. Conversely, pool-based active learning computes a surrogate model of the unknown objective function that is used to inform the learner toward a given goal, and is updated during the procedure to refine the informative content supporting the learning procedure. Thus, pool-based active learning methods realize goal-driven learning schemes and can be collocated in the adaptive learning class while population-based active learning techniques can not be considered as adaptive samplings. Following the proposed classification, Bayesian optimization represents the logic intersection between active learning and adaptive sampling since (i) BO realizes an adaptive sampling scheme toward a given goal, and (ii) the BO goal-driven learning procedure is guided by learning principles also traceable in active learning schemes. This synergy between Bayesian optimization and active learning is the main focus of our work, and the remaining of this manuscript is dedicated to formalize and discuss this dualism. To support this discussion, we provide additional details of the proposed classification for adaptive sampling, and review some popular approaches for each of the three classes. The literature on adaptive sampling is vast, and a complete review goes beyond the purpose of this work. Although our discussion will not be comprehensive, the objective is to highlight the distinguishing features of each class and clarify the relative positions of adaptive sampling, active learning and Bayesian optimization.

\subsection{Adaptive Probing} \label{s: adaptProb}

Adaptive probing schemes exploit the observations of previous samples without computing any surrogate model. These sampling procedures are informed exclusively from the collected data to guide the selection of the next location to query, and exclude the adoption of emulators to support the search. Several adaptive probing frameworks have been developed based on the Monte Carlo method \cite{shapiro2003monte, owen2003quasi}. Among these, adaptive importance samplings \cite{karamchandani1989adaptive, bugallo2017adaptive, peherstorfer2016multifidelity} and adaptive Markov Chain Monte Carlo samplings \cite{atchade2005adaptive, atchade2011adaptive} represent popular methodologies adopted in different practical scenarios, from signal processing \cite{bugallo2015adaptive, zhou2011abrupt} to reliability analysis of complex systems \cite{xiao2020new, jensen2020adaptive}. Adaptive importance sampling uses previously observed samples to adapt the proposal densities and locate the regions from which samples should be drawn; this strategy permits to iteratively improve the quality of the samples distribution and enhance the accuracy of the relative inference from these observations. Adaptive Markov Chain Monte Carlo (MCMC) determines the parameters of the MCMC transition probabilities on the fly through already collected information. This adaptively generates new samples from an usually complex and high-dimensional distribution, and enhances the overall computational efficiency and reliability of the procedure. In the next paragraph, we report the mathematical formulation of adaptive importance sampling to illustrate the properties of adaptive probing methodologies and the elements that differentiate them from active learning paradigms.

\subsubsection{Adaptive Importance Sampling} \label{s: ASearch}

Adaptive Importance Sampling (AIS) usually considers a generic inference problem characterized by a certain probability density function (pdf) $\Distribution(\DesignVar)$ of a $\DimFun$-dimensional vector of unknown statistic real parameters $\DesignVar \in \DesignSpace$. AIS frameworks aim to provide a numerical approximation of some particular moment of $\DesignVar$: 

\begin{equation} \label{e: AIS1}
    \Integral(\ObjFun) = \Expectation_{\Distribution}\left[ \ObjFun(\DesignVar)\right] = \int \ObjFun(\DesignVar)\Distribution(\DesignVar)\,d\DesignVar
\end{equation}

\noindent where $\ObjFun: \DesignSpace \rightarrow \mathbb{R}$ can be any function of $\DesignVar$ integrable with respect to the pdf $\Distribution(\DesignVar)$ 

The integral $\Integral(\ObjFun)$ is representative of different mathematical problems, from Bayesian inference \cite{robert1999monte} to the estimate of rare events \cite{hesterberg1995weighted}. In many practical scenarios, the integral $\Integral(\ObjFun)$ can not be computed in closed form. Adaptive importance sampling provides an algorithmic framework to efficiently address this problem. 

Let us define a proposal probability density function $\ProposalPDF(\DesignVar)$ to simulate samples under the restriction that $\ProposalPDF(\DesignVar)>0$ for all $\DesignVar$ where $\Distribution(\DesignVar) \ObjFun(\DesignVar) \neq 0$. AIS provides an iterative procedure that improves the quality of one or multiple proposals $\ProposalPDF(\DesignVar)$ to approximate a non-normalized non-negative target function $\TargetDistribution(\DesignVar)$. At the beginning, AIS initializes $\NumSamples$ proposals $\{\ProposalPDF_{\IndNumSamples}(\DesignVar | \ParamVector_{\IndNumSamples,1} ) \}_{\IndNumSamples=1}^{\NumSamples}$ parameterized through the vector $\ParamVector_{\IndNumSamples,1}$. Then, the procedure simulates $\SimSamples$ samples from each proposal $\DesignVar_{\IndNumSamples,1}^{(\IndSimSamples)}, \;\; \IndNumSamples = 1,...,\NumSamples, \, \IndSimSamples = 1,...,\SimSamples$, and assigns to each sample an associated importance weight formalized as follows:

\begin{equation}
    \ImpWeight_{\IndNumSamples} = \frac{\TargetDistribution(\DesignVar_{\IndNumSamples})}{\ProposalPDF(\DesignVar_{\IndNumSamples})} \, , \;\;\;\; \IndNumSamples = 1,...,\NumSamples
\end{equation}

These importance weights measure the representativeness of each sample simulated from the proposal pdf $\ProposalPDF(\DesignVar)$ with reference to the distribution of random variables $\Distribution(\DesignVar)$. 

At this point, this set of $\NumSamples$ weighted samples $( \DesignVar_{\IndNumSamples,1}^{(\IndSimSamples)},\ImpWeight_{\IndNumSamples,1}^{(\IndSimSamples)} ), \;\;\;\; \IndNumSamples = 1,...,\NumSamples, \, \IndSimSamples = 1,...,\SimSamples$ is used to define a self-normalized estimator: 

\begin{equation}
    \hat{\Integral}^{\NumSamples}(\ObjFun) = \sum_{\IndNumSamples=1}^{\NumSamples} \bar{\ImpWeight}_{\IndNumSamples} \ObjFun(\ImpWeight_{\IndNumSamples})
\end{equation}

\noindent where $\bar{\ImpWeight}_{\IndNumSamples} =  \ImpWeight_{\IndNumSamples}/\sum_{j = 1}^{\NumSamples} \ImpWeight_j$ are the normalized weights. This permits to approximate the target function distribution as follows: 

\begin{equation}
    \Distribution^{\NumSamples}(\DesignVar) = \sum_{\IndNumSamples=1}^{\NumSamples} \bar{\ImpWeight}_{\IndNumSamples} \Dirac (\DesignVar - \DesignVar_{\NumSamples})
\end{equation}

\noindent where $\Dirac$ represents the Dirac measure. 

Finally, AIS realizes the adaptation phase and updates the parameters of the $\IndNumSamples$-th proposals from $\ParamVector_{\IndNumSamples,1}$ to $\ParamVector_{\IndNumSamples,2}$ using the last set of drawn parameters \cite{martino2015adaptive} or all the parameters evaluated so far \cite{el2019efficient}. The whole procedure is repeated until a certain termination criteria is met (e.g. maximum number of iterations). 

This adaptive policy permits to gradually evolve the single or multiple proposal densities to accurately approximate the target pdf. The generation of new samples is uniquely driven by the measurement of the importance of previous samples (weighting) that supports the updating of the proposal parameters (adaptation). Thus, AIS adaptively locates promising regions to query without benefiting from an overall quantification of the goodness of all the spectrum of samples available in the domain -- e.g. through the construction of a surrogate model. On these basis, AIS and the general class of adaptive probing strategies is not considerable as a learning procedures since the adaptation phase is not informed by a surrogate model updated on the fly during the procedure, and is not guided by a "learner" that assimilates information from this emulator and adapts the next queries to achieve a given goal.

\subsection{Adaptive Modeling} \label{s: adaptModel}

Adaptive modeling paradigms sample the domain supported by the information from previous queries, and use the collected data to build a surrogate model. However, the informative content encoded in the emulator is not used to guide the sampling and decide the next point to evaluate. Adaptive modeling approaches have been extensively developed for the reliable propagation and quantification of uncertainties \cite{jakeman2012local, jakeman2020adaptive}, analysis of ordinary or partial differential equations \cite{gunzburger2014adaptive, eigel2022convergence}, and inverse problems \cite{marzouk2009stochastic, ma2009efficient}. One common approach is represented by adaptive stochastic collocation methodologies, which use an adaptive sparse grid approximation scheme to construct an interpolant polynomial in a multi-dimensional random space \cite{haji2016multi, lang2020fully}. The adaptive selection of collocation points is driven by an error indicator \cite{gerstner2003dimension} or estimator \cite{guignard2018posteriori} that evaluates a certain number of sparse admissible subspaces of the domain: the subspace that exhibits the higher error is included in the grid and the new set of subspaces is identified. Other well-known adaptive modeling approaches are residual-based samplings distribution \cite{wu2023comprehensive}. This family of techniques is mostly applied to improve the training efficiency of Physics-Informed Neural Networks (PINN) surrogate models. Residual-based approaches enhance the distribution of residual points by placing more samples according to certain properties of the residuals during the training of PINN. This decision can be made on the basis of locations where the residual of the partial differential equation is large \cite{lu2021deepxde}, according to a probability density function of the residual points \cite{nabian2021efficient}, and hybrid approaches of the above \cite{wu2023comprehensive}. This permits to achieve a better accuracy of the final PINN surrogate model while containing the computational burden associated with computations. Both stochastic collocation and residual based samplings are intended to build an efficient and accurate surrogate model over the domain of samples. However, the sampling procedure is adapted uniquely to previous evaluated samples without a learning procedure from data: the surrogate model is not used to inform the decision on where to sample, and is not progressively updated with previous information. In the following, we provide general mathematical details about adaptive stochastic collocation to analyze the peculiarities of the adaptive modeling class, and underline the absence of a learning process during the construction of the surrogate model.

\subsubsection{Adaptive Stochastic Collocation} \label{s: AModeling}

Adaptive Stochastic Collocation (ASC) builds an interpolation function to approximate the outputs from a model of interest. This emulator is constructed on the evaluations of the model at valuable collocation points of the stochastic inputs to obtain the moments and the probability density function of the outputs.

Consider any point $\points$ contained in the random space $\RandSpace \subset \mathbb{R}^{\dimpoints}$ with probability distribution function $\pdf(\points)$. The goal of ASC is to find an interpolating polynomial $\IntPolynomial(\ObjFun)$ to approximate a smooth function $\ObjFun(\points) : \mathbb{R}^{\dimpoints} \rightarrow \mathbb{R}$: 

\begin{equation}
    \IntPolynomial(\ObjFun)(\points_k) = \ObjFun(\points_k) \, , \;\;\;\; 1 \leq k \leq \NumPoints
\end{equation}

\noindent for a given set of points $\{ \points_k \}_{k=1}^{\NumPoints}$. The selection of the collocation points majorly influences the capability of the interpolating polynomial to be close to the original function $\ObjFun$. For multivariate problems, the interpolation function is defined as follows using the tensor product grid: 

\begin{equation} \label{e: ASC1}
\begin{split}
     \IntPolynomial(\ObjFun) &= (\IntRules^{\NodeLevel_1} \otimes \cdots \otimes \IntRules^{\NodeLevel_\dimpoints})(\ObjFun)  \\ &= \sum_{j_1=1}^{\NumNodes_{\NodeLevel_1}} \cdots \sum_{j_\dimpoints=1}^{\NumNodes_{\NodeLevel_\dimpoints}} \ObjFun(\points_{j_1}^{\NodeLevel_1},...,\points_{j_\dimpoints}^{\NodeLevel_\dimpoints}) \cdot (\LagrangePol_{j_1}^{\NodeLevel_1} \otimes \cdots \otimes \LagrangePol_{j_\dimpoints}^{\NodeLevel_\dimpoints})
\end{split}
\end{equation}

\noindent where $\IntRules^{\NodeLevel_k}$ is the univariate interpolation function for the level $\NodeLevel_k$ in the $k$-th coordinate, $\points_{j_m}^{\NodeLevel_k}$ is the $j_m$-th node, and $\LagrangePol_{j_k}$ are the Lagrange interpolating polynomials. 


Equation \ref{e: ASC1} demands for $\NumNodes_{\NodeLevel_1} \times \cdots \times \NumNodes_{\NodeLevel_\dimpoints}$ nodes, which indicate an exponential rate of computational cost growth with the number of dimensions. Adaptive stochastic collocation targets the reduction of this computational effort through an adaptive sparse grid of collocation points: the objective is to wisely place more points of the grid in the important directions to prioritize the collection of highly informative data. This adaptive sparse grid is defined through a subset of the full tensor product grid as follows: 

\begin{equation} \label{e: ASC2}
\begin{split}
    \Grid_{\SparsenessParam, \dimpoints}(\ObjFun) &= \sum_{|\mathbi{\NodeLevel}| \leq \SparsenessParam} (\Delta \IntRules^{\NodeLevel_1} \otimes \cdots \otimes \Delta \IntRules^{\NodeLevel_\dimpoints})(\ObjFun) \\ &= \Grid_{\SparsenessParam-1, \dimpoints} (\ObjFun) + \sum_{|\mathbi{\NodeLevel}| = \SparsenessParam}(\Delta \IntRules^{\NodeLevel_1} \otimes \cdots \otimes \Delta \IntRules^{\NodeLevel_\dimpoints})(\ObjFun)
\end{split}
\end{equation}

\noindent where $\mathbi{\NodeLevel} = (\NodeLevel_1,...,\NodeLevel_{\dimpoints}) \in \mathbb{R}^{\dimpoints}$, $|\mathbi{\NodeLevel}| = \NodeLevel_1+...+\NodeLevel_{\dimpoints}$, $\SparsenessParam$ is the sparseness parameter, and the difference formulas are defined by $\IntRules^0 = 0$ and $\Delta \IntRules^{\NodeLevel} = \IntRules^{\NodeLevel} - \IntRules^{\NodeLevel-1}$. 

Equation \ref{e: ASC2} leverages the previous results to extend the interpolation from level $\SparsenessParam-1$ to $\SparsenessParam$ through the evaluation of the multivariate function on the sparse grid: 

\begin{equation}
\begin{split}
    \Nodes_{\SparsenessParam, \dimpoints} &= \bigcup_{|\mathbi{\NodeLevel}| \leq \SparsenessParam}(\Delta \NodalSet^{\NodeLevel_1} \times \cdots \times \Delta \NodalSet^{\NodeLevel_\dimpoints})\\ &= \Nodes_{\SparsenessParam-1, \dimpoints} + \bigcup_{|\mathbi{\NodeLevel}| = \SparsenessParam}(\Delta \NodalSet^{\NodeLevel_1} \times \cdots \times \Delta \NodalSet^{\NodeLevel_\dimpoints})
\end{split}
\end{equation}

\noindent where $\Delta \NodalSet^{\NodeLevel} = \NodalSet^{\NodeLevel} \backslash \NodalSet^{\NodeLevel-1}$ are the newly added set of univariate nodes $\NodalSet^{\NodeLevel_k}$ for level $\NodeLevel_k$ in the $k$-th coordinate.

This scheme adapts the sampling procedure through the knowledge acquired on the fly, and efficiently leverages data to improve the quality of the interpolation function. In this case, the selection of the collocation points is intended to compute an emulator of the target function, but the adaptive sampling is not driven by the information acquired from this emulator. In addition, the acquisition of data is not used to learn and update the surrogate model. These considerations on ASC can be extended to the general class of adaptive modeling methods: even if the sampling scheme is conceived to construct surrogate models, the selection of promising locations to query is not delegated to a goal-driven learner that leverages a mutual exchange of information with the surrogate.

\subsection{Adaptive Learning} \label{s: adaptLear}

Adaptive learning methodologies realize goal-driven learning processes characterized by the mutual exchange of information between the surrogate model and the goal-driven learner: the former is updated and refined after new evaluations of samples while the latter decides the next query based on the updated approximation given by the emulator. Bayesian optimization and pool-based active learning belong to this specific class of adaptive sampling techniques. Bayesian frameworks constitute a learning process driven by the mutual informative assimilation between an acquisition function -- learner -- and a surrogate model \cite{Mockus2012, frazier2018tutorial}. The acquisition function commensurates the benefit of evaluating samples based on the prediction of the surrogate model, and selects the most useful sample to query toward the given goal -- either to improve the accuracy of the surrogate over the domain or to effectively inform the optimization search; at the same time, the emulator is enriched with the data from the new query, and is updated to refine the approximation of the objective function over the domain. Similarly, pool-based active learning methods search the domain through a goal-driven learner informed by a classification model of samples \cite{Settles2009, ZhanAl2021}. This process is characterized by the reciprocal flow of information between the learner and the emulator: the classification model is updated through the new evaluations of unsampled locations, and the learner uses these information to select the next query. Mathematical details about pool-based active learning are provided in the following section to better clarify the distinction between this class of adaptive learners, and the other classes which do not realize a goal-driven learning procedure.

\subsubsection{Pool-Based Active Learning} \label{s: ALearning}

Pool-based active learning commonly defines an optimal sampling strategy to improve the accuracy of a surrogate model adopted to classify data-points from a target distribution of labels over the domain of samples $\DesignSpace$. Considering this general classification task, pool-based active learning routine is grounded on a probabilistic estimate of the distribution of features $\ObjFun$ over the entire domain $\DesignSpace$ through a surrogate model $\hat{\ObjFun}$. This emulator is trained on a set of collected data-points, and maps features to labels $\ObjFun_\NumSamples(\DesignVar_\IndNumSamples) = \hat{\ObjFun}_\IndNumSamples$ through a predicted probability $\Probability_\NumSamples (\ObjFun_\IndNumSamples = \ObjFun | \DesignVar_\IndNumSamples)$ that estimates the distribution of features over the domain. Suppose we have collected from a large pool of unlabelled data $\DesignSpace$ the -- small-- dataset $\Dataset_\NumSamples\{ \DesignVar_\IndNumSamples, \ObjFun(\DesignVar_\IndNumSamples) \}_{\IndNumSamples=1}^{\NumSamples}$ observing the label values $\ObjFun(\DesignVar_{\IndNumSamples})$ in output from an observation model or oracle at some informative locations $\DesignVar_{\IndNumSamples}$. Based on this dataset, the goal-driven procedure learns a surrogate model $\hat{\ObjFun}_\NumSamples$ whose predictive framework emulates the behaviour of samples over the domain based on the previous collected information.

At this point, an utility function acts as the goal-driven learner informed by the surrogate model, and identifies the most promising sample to be labelled by the oracle according to a measure of utility with respect to the given goal -- improve the accuracy of the classifier. The next query augments the dataset $\Dataset_{\NumSamples+1} = \Dataset_{\NumSamples} \bigcup \{ \DesignVar_{\NumSamples+1}, \ObjFun_{\NumSamples+1} \}$ and the surrogate model is updated. This utility function defines a learning policy that maps the current predictive distribution to a decision/action on where to sample in the next iteration as follows:

\begin{equation} \label{e: PBAL1}
    \DesignVar_{\NumSamples+1} = \text{arg}\,\max \AF(\Probability_\NumSamples (\ObjFun_\IndNumSamples = \ObjFun | \DesignVar_\IndNumSamples))
\end{equation}

Equation \eqref{e: PBAL1} mathematically formalizes the concept of goal-driven learning procedure: the learner leverages the predicted probability of the surrogate $\Probability_\NumSamples (\ObjFun_\IndNumSamples = \ObjFun | \DesignVar_\IndNumSamples)$ to make an action $\DesignVar_{\NumSamples+1}$; at the same time, the decision is used to enrich the dataset $\Dataset\{ \DesignVar_\IndNumSamples, \ObjFun(\DesignVar_\IndNumSamples) \}_{\IndNumSamples=1}^{\NumSamples+1}$ and update the predicted probability $\Probability_{\NumSamples+1}$. This mutual exchange and assimilation between the learner and the surrogate represents the key aspect that defines a goal-driven learning process and the whole class of adaptive learning sampling schemes.

\section{Bayesian Frameworks} \label{s:BOFrameworks}

Bayesian optimization constitutes the mid-point between adaptive sampling and active learning. This intersection represents the focal point of our work, and motivates the substantial synergy between Bayesian optimization and active learning as adaptive sampling schemes capable of learning from data and accomplish a certain learning goal. The remaining of this section is dedicated to the general overview of Bayesian optimization considering both a single source of information (Section \ref{s:BO}) and when multiple sources are available to the learning procedure (Section \ref{s:MFBO}). This will guide the reader into the next sections that make explicit the symbiosis between Bayesian frameworks and active learning through our original perspective of Bayesian optimization as a way to actively learn with acquisition functions (Section \ref{s:An Active Learning Perspective}).

\subsection{Bayesian Optimization}  \label{s:BO}

The birth of Bayesian optimization can be retraced in 1964 with the work of Kushner \cite{kushner1964new} where unconstrained one-dimensional optimization problems are addressed through a predictive framework based on the Wiener process surrogate model, and a sampling scheme guided by the probability of improvement acquisition function. Further contributions have been proposed by Zhilinskas \cite{zhilinskas1975single} and Mockus \cite{movckus1975bayesian}, and the methodology has been extended to high dimensional optimization problems in the works of Stuckman \cite{stuckman1988global} and Elder \cite{elder1992global}. Bayesian optimization achieved resounding success after the introduction of the Efficient Global Optimization (EGO) algorithm by Jones et al. \cite{jones1998}. EGO uses a Kriging surrogate model to predict the distribution of the objective function, and adopts the expected improvement acquisition function to measure the improvement of the optimization procedure obtained evaluating unknown samples.

The EGO methodology paves the way to the application of Bayesian optimization over a wide range of problems in science and engineering. These research fields demand for the efficient management of the information from black-box representations of the objective function -- the procedure is only aware of the input and output without a priori knowledge about the function  -- to guide the optimization search. Engineering has been a pioneer in the adoption of Bayesian optimization: the design optimization of complex systems is frequently characterized by computationally intensive black-box functions which require efficient global optimization methods. Early applications relate to engineering design optimization \cite{zang2002needs}, computer vision \cite{Wu2005ABayesian} and combinatorial problems \cite{li2003bayesian}. Nowadays, the Bayesian framework becomes widely adopted in many fields including and not limited to engineering \cite{frazier2016bayesian, LamAl2018, kong2020energy, priem2020efficient}, robotics and reinforcement learning \cite{berkenkamp2021bayesian, balakrishnan2020efficient, young2020distributed}, finance and economics \cite{gonzalvez2019financial, pour2022cryptocurrency}, automatic machine learning \cite{victoria2021automatic, turner2021bayesian}, and preference learning \cite{dudley2019crowdsourcing, koyama2020sequential}. In addition, significant advances have been made in the expansion of BO methodologies to higher-dimensional search spaces frequently encountered in science and engineering, where the effectiveness of the search procedure is usually correlated to an exponential growth of the required observations of the objective function and associated demand for computational resources and time. Within this context, BO techniques have been scaled to approach high-dimensional problems exploiting potential additive structures of the objective function \cite{kandasamy2015high, wang2018batched}, mapping high-dimensional search spaces into low-dimensional subspaces \cite{nayebi2019framework, wang2016bayesian}, learning from observations of multiple input points evaluated through parallel computing \cite{shah2015parallel, wang2020parallel}, and through simultaneous local optimization approaches \cite{eriksson2019scalable}.

Given a black-box expensive objective function $\ObjFun : \DesignSpace \rightarrow \mathbb{R}$, Bayesian optimization seeks to identify the input $\MinDesignVar \in \min_{\DesignVar \in \DesignSpace} \ObjFun(\DesignVar)$ that minimizes the objective $\ObjFun$ over an admissible set of queries $\DesignSpace$ with a reduced computational cost. To achieve this goal, Bayesian optimization relies on an adaptive learning scheme based on a surrogate model that provides a probabilistic representation of the objective $\ObjFun$, and uses this information to compute an acquisition function $\AF(\DesignVar) : \DesignSpace \rightarrow \mathbb{R}^{+}$ that drives the selection of the most promising sample to query. Let us consider the available information regarding the objective function $\ObjFun$ stored in the dataset $\Dataset_{\NumObs} = \{ (\DesignVar_1,\NoisyObserv_1),..., (\DesignVar_{\IndNumObs}, \NoisyObserv_{\IndNumObs}) \}$ where $\NoisyObserv_\IndNumObs \sim \NorDist (\ObjFun (\DesignVar_\IndNumObs), \StandDevNoise(\DesignVar_\IndNumObs))$ are the noisy observations of the objective function and $\StandDevNoise$ is the standard deviation of the normally distributed noise. 

At each iteration of the optimization procedure, the surrogate model depicts possible explanations of $\ObjFun$ as $\ObjFun \sim p(\ObjFun | \Dataset_{\NumObs})$ applying a joint distribution over its behaviour at each sample $\DesignVar \in \DesignSpace$. Typically, Gaussian Processes (GPs) have been widely used as the surrogate model for Bayesian optimization \cite{Rasmussen2003, OsborneAl2009}. In GP regression, the prior distribution of the objective $p(\ObjFun)$ is combined with the likelihood function $p(\Dataset_{\NumObs}|\ObjFun)$ to compute the posterior distribution $p(\ObjFun|\Dataset_{\NumObs}) \propto  p(\Dataset_{\NumObs}|\ObjFun)p(\ObjFun)$, representing the updated belief about $\ObjFun$. The GP posterior is a joint Gaussian distribution $p(\ObjFun|\Dataset_{\NumObs}) = \NorDist(\MeanFunGP(\DesignVar), \CovFunGP (\DesignVar, \DesignVar'))$ completely specified by its mean $\MeanFunGP(\DesignVar) = \Expectation \left[ \ObjFun(\DesignVar) \right]$ and covariance (also referred as kernel) function $\CovFunGP(\DesignVar, \DesignVar') = \Expectation \left[ (\ObjFun(\DesignVar) - \MeanFunGP(\DesignVar)) (\ObjFun(\DesignVar') - \MeanFunGP(\DesignVar')) \right]$, where $\MeanFunGP(\DesignVar)$ represents the prediction of the GP model at $\DesignVar$ and $\CovFunGP(\DesignVar, \DesignVar')$ the associated uncertainty. 

BO uses this statistical belief to make the decision on where to sample assisted by an acquisition function $\AF$, which identifies the most informative sample $\DesignVar_{new} \in \DesignSpace$ that should be evaluated via maximization $\DesignVar_{new} \in \max_{\DesignVar \in \DesignSpace} \AF(\DesignVar)$. Then, the objective function is evaluated at $\DesignVar_{new}$ and this information is used to update the dataset  $\Dataset_{\NumObs} = \Dataset_{\NumObs} \cup (\DesignVar_{new},\NoisyObserv(\DesignVar_{new}))$. Acquisition functions are designed to guide the search for the optimum solution according to different infill criteria which provide a measure of the improvement that the next query is likely to provide with respect to the current posterior distribution of the objective function. In engineering applications, we could retrieve different implementations proposed for the acquisition function, which differ for the infill schemes adopted to sample pursuing the optimization goal. Examples include the Probability of Improvement (PI) \cite{kushner}, Expected Improvement (EI) \cite{jones1998}, Entropy Search (ES) \cite{hennig&Schuler2012} and Max-Value Entropy Search (MES) \cite{wang_mes}, Knowledge-Gradient (KG) \cite{scott2011correlated}, and non-myopic acquisition functions \cite{lam2017lookahead, wu2019practical}. 

The Probability of Improvement (PI) acquisition function encourages the selection of samples that are likely to obtain larger improvements over the current minimum predicted by the surrogate model, while the Expected Improvement (EI) considers not only the PI but also the expected gain in the solution of the optimization problem achieved evaluating a certain sample. Other popular schemes are entropy-based acquisition functions such as the Entropy Search (ES) and Max-Value Entropy Search (MES), which rely on estimating the entropy of the location of the optimum and the minimum function value, respectively, to maximize the mutual information between the samples and the location of the global optimum. Knowledge-gradient sampling procedures are conceived for applications where the evaluations of the objective function are affected by noise, recommending the location that maximizes the increment of the expected value that would be acquired by taking a sample from the location. Through the adoption of non-myopic acquisition functions, the learner maximizes the predicted improvement over future iterations of the optimization procedure, overcoming myopic schemes where the improvement of the solution is measured at the immediate step ahead.

\subsection{Multifidelity Bayesian Optimization}  \label{s:MFBO}

The evaluation of black-box functions in engineering and science frequently requires time-consuming lab experiments or expensive computer-based models, which would dramatically increase the computational burden for the optimization procedure. This is the case of large-scale design optimization problems, where the evaluation of the objective function for enough samples can not be afforded in practice. In many real-world applications, the objective function can be computed using multiple representations at different levels of fidelity $\{\ObjFun^{(1)},...,\ObjFun^{(\MaxLevFid)}\}$, where the lower the level of fidelity the less accurate but also less time-consuming the evaluation procedure. Multifidelity methods recognize that different representative levels of fidelity and associated costs can be used to accelerate the optimization process, and enable a flexible trade-off between computational cost and accuracy of the solution. In particular, multifidelity optimization leverages low-fidelity data to massively query the domain, and uses a reduced number of high-fidelity observations to refine the belief about the objective function toward the optimum \cite{ForresterAl2007, PeherstorferAl2018, BeranAl2020}.

Accordingly, Multifidelity Bayesian Optimization (MFBO) learns a surrogate model that synthesizes through stochastic approximation the multiple levels of fidelity available, and uses an acquisition function as the learner that selects the most promising sample and associated level of fidelity to interrogate. This learning procedure provides potential accelerations of the optimization procedure that is reflected in the likely improvement of the surrogate accuracy. According to Godino et al. \cite{giselle2019issues}, the improvement in performance occurs usually if the acquisition of large amount of high-fidelity data is hampered by the computational expense, the correlation between high-fidelity and low-fidelity data is high, and low-fidelity models are sufficiently inexpensive; Under different circumstances, multifidelity optimization might not deliver substantial accelerations and quality of the surrogate: the relationship between dimension of the training set and surrogate accuracy is not monotonically increasing, as evidenced by \cite{davis2018efficient}. In recent years, multifidelity Bayesian optimization has been successfully adopted for optimization problems ranging from engineering design optimization \cite{BonfiglioAl2018, meliani2019multi, grassi2023raal, di2021multifidelity, di2023nm, serani2022resistance}, automatic machine learning \cite{wu2020, kandasamy2017multi}, applied physics \cite{irshad2023multi, winter2023multi}, and medical applications \cite{perdikaris2016model, PezzutoAl2022}. In the context of high-dimensional problems, multifidelity Bayesian optimization capitalizes from fast low-fidelity models to alleviate the computational burden associated with the required numerous observations of the objective function to effectively direct the search toward the given goal, and achieved promising results in terms of accuracy and efficiency for applications in quantum control \cite{lazin2023high}, aerospace engineering \cite{sarkar2019multifidelity}, and reinforcement learning \cite{imani2021scalable}.



Multifidelity Bayesian optimization determines a learning procedure informed by the surrogate model of the objective function constructed on the dataset of noisy objective observations $\Dataset_\NumObs=\{ (\DesignVar_1, \NoisyObserv^{(\LevFid_1)}_1), ..., (\DesignVar_\IndNumObs, \NoisyObserv^{(\LevFid_\IndNumObs)}_\IndNumObs) \}$, where $\NoisyObserv^{(\LevFid_\IndNumObs)}_\IndNumObs \sim \NorDist (\ObjFun^{(\LevFid_\IndNumObs)} (\DesignVar_\IndNumObs), \StandDevNoise(\DesignVar_\IndNumObs))$ and $\StandDevNoise$ has the same distribution over the fidelities. This multifidelity surrogate model defines an approximation of the objective $\ObjFun^{(\LevFid)} \sim p(\ObjFun^{(\LevFid)} | (\DesignVar, \LevFid), \Dataset_\NumObs)$ at different level of fidelity, and represents the belief about the distribution of the objective function over the domain $\DesignSpace$ based on data. A popular practice for MFBO is to extend the Gaussian process surrogate model to a multifidelity setting through an autoregressive scheme \cite{kennedy2000predicting}: 

\begin{equation} \label{e:AutoRegScheme}
\ObjFun^{(\LevFid)} = \ConsFact^{(\LevFid - 1)} \ObjFun^{(\LevFid - 1)} \left( \mathbi{\DesignVar} \right) + \Discrep^{(\LevFid)} \left( \mathbi{\DesignVar} \right) \quad \LevFid = 2,...,\MaxLevFid
\end{equation} 

\noindent where $\ConsFact^{(\LevFid - 1)}$ is a constant scaling factor that includes the contribution of the previous fidelity with respect to the following one, and $\Discrep^{(\LevFid)} \sim GP(0, \CovFunGP^{(\LevFid)} \left( \mathbi{\DesignVar}, \mathbi{\DesignVar}'\right))$ models the discrepancy between two adjoining levels of fidelity. The posterior of the multifidelity Gaussian process is completely specified by the multifidelity mean function $\MeanFunGP^{(\LevFid)}(\DesignVar, \LevFid) = \Expectation \left[ \ObjFun^{(\LevFid)}(\DesignVar) \right]$ that represents the approximation of the objective function at different levels of fidelity, and the multifidelity covariance function $\CovFunGP^{(\LevFid)}((\DesignVar, \LevFid), (\DesignVar', \LevFid)) = \Expectation \left[ (\ObjFun^{(\LevFid)}(\DesignVar, \LevFid) - \MeanFunGP^{(\LevFid)}(\DesignVar, \LevFid)) (\ObjFun^{(\LevFid)}(\DesignVar', \LevFid) - \MeanFunGP^{(\LevFid)}(\DesignVar', \LevFid)) \right]$ that defines the associated uncertainty for each level of fidelity.

The availability of multiple representations of the objective function poses a further decision task that has to be accounted by the learner during the sampling of unknown locations: the selection of the most promising sample is effected with the simultaneous designation of the information source to be evaluated. This is obtained through a learner represented by the multifidelity acquisition function $\AF(\DesignVar, \LevFid)$ that extends the infill criteria of Bayesian optimization, and selects the pair of sample and the associated level of fidelity to query $(\DesignVar_{new}, \LevFid_{new}) \in \max_{\DesignVar \in \DesignSpace, \LevFid \in \FidSpace} \AF(\DesignVar, \LevFid)$ that is likely to provide higher gains with a regard for the computational expenditure. Among different formulations, well known multifidelity acquisition functions to address optimization problems are the Multifidelity Probability of Improvement (MFPI) \cite{RuanAl2020}, Multifidelity Expected Improvement (MFEI) \cite{HuangAl2006}, Multifidelity Predictive Entropy Search (MFPES) \cite{ZhangAl2017mf}, Multifidelity Max-Value Entropy Search (MFMES) \cite{TakenoAl2020}, and non-myopic multifidelity expected improvement \cite{di2022non}. These formulations of the acquisition function define adaptive learning schemes that retain the infill principles characterizing the single-fidelity counterpart, and account for the dual decision task balancing the gains achieved through accurate queries with the associated cost during the optimization procedure.

\section{An Active Learning Perspective} \label{s:An Active Learning Perspective}

Bayesian frameworks and Active learning schemes exhibit a strong synergy: in both cases the learner seeks to design an efficient sampling policy to accomplish the learning goal, and is guided by a surrogate model that informs the learner and is continuously updated during the learning procedure. Active learning literature is vast an include a multitude of approaches \cite{Abe1998, BurbidgeAl2007, CaiAl2016, Demir&Bruzzone2014, Raychaudhuri&Hamey1995, Settles&Burr2008, Seung&Opper1992, Wu2018}. According to the well accepted classification proposed by Sugiyama and Nakajima \cite{Sugiyama&Nakajima2009}, active learning strategies can be categorized in population-based and pool-based active learning frameworks according to the nature of the sampling scheme defined by the learner. Population-based active learning targets the identification of the best optimal density of the samples for training known the target distribution. Conversely, pool-based active learning defines an efficient sampling scheme to improve the efficiency of a surrogate model of the unknown target distribution over the domain of samples.

This paper explicitly formalizes and discusses Bayesian frameworks as an active learning procedure realized through acquisition functions. In particular, pool-based active learning shows in essence a strong dualism with Bayesian frameworks. We emphasize this synergy through the dissertation on the correspondence between learning criteria and infill criteria; the former drive the sampling procedure in pool-based active learning, while the latter guide the search in Bayesian schemes through the acquisition function. This symbiosis is evidenced for the case of a single source of information adopted to query samples, and when multiple sources are at disposal of the learner to evaluate new input. Accordingly, we review and discuss popular sampling policies commonly adopted in pool-based active learning, and discern the learning criteria to accomplish a specific learning goal (Section \ref{s:learning criteria}). Then, the attention is dedicated to the identification of the infill criteria realized through popular acquisition functions in Bayesian optimization (Section \ref{s: acquisition functions }). The objective is to explicitly formalize the synergy between Bayesian frameworks and Active learning as adaptive sampling schemes guided by common principles. The same avenue is followed to formalize this dualism for the case of multiple sources of information available during the learning procedure. In particular, we identify the learning criteria adopted in pool-based active learning with multiple oracles (Section \ref{s:Active learning with multiple oracles}), and compare them with the infill criteria specified by well-established multifidelity acquisition functions in multifidelity Bayesian optimization (Section \ref{s: multifidelity acquisition function}). The objective is to clarify the shared principles and the mutual relationship that characterize the two adaptive learning schemes when the decision of the sample to query requires also the selection of the appropriate source of information to be evaluated.

\subsection{Learning Criteria} 
\label{s:learning criteria}

Pool-based active learning determines a tailored sampling policy to ensure the maximum computational efficiency of the adaptive sampling procedure -- limited and well selected amount of samples to query. This adaptive learning demands for principled guidelines to decide whether to evaluate or not a certain sample based on a measure of its goodness. Learning criteria permit to establish a metric to quantify the gains of all the possible learner decisions, and prescribe an optimal decision based on the information acquired from the surrogate model. The vast majority of the literature concerning pool-based active learning identifies three essential learning criteria: informativeness, representativeness and diversity \cite{HeAl2014, ShenAl2004, WuAl2016, Wu2018, Monarch2021, ZhanAl2021}:

\begin{figure*}[t!]
    \centering
     \subfigure[Informativeness]{%
        \includegraphics[width=0.3\linewidth,trim=200 0 200 0, clip]{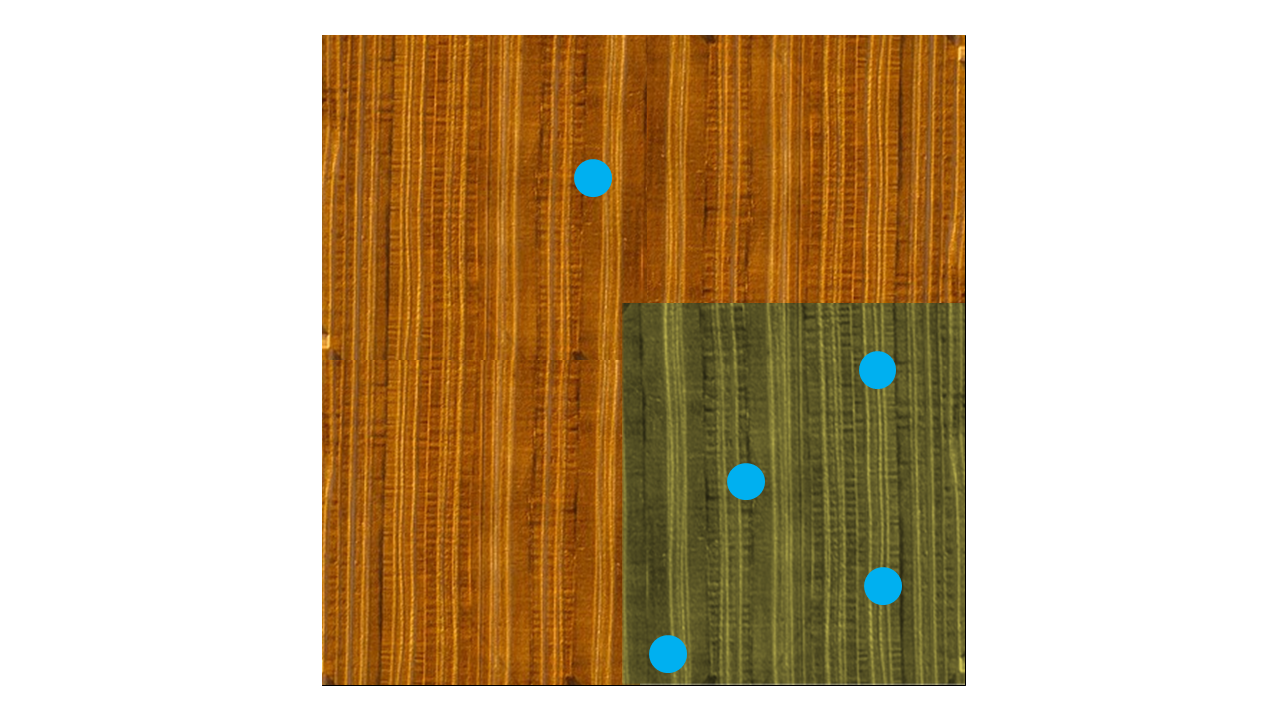} 
        \label{fig:informativeness}}
        \subfigure[Representativeness]{%
        \includegraphics[width=0.3\linewidth,trim=200 0 200 0, clip]{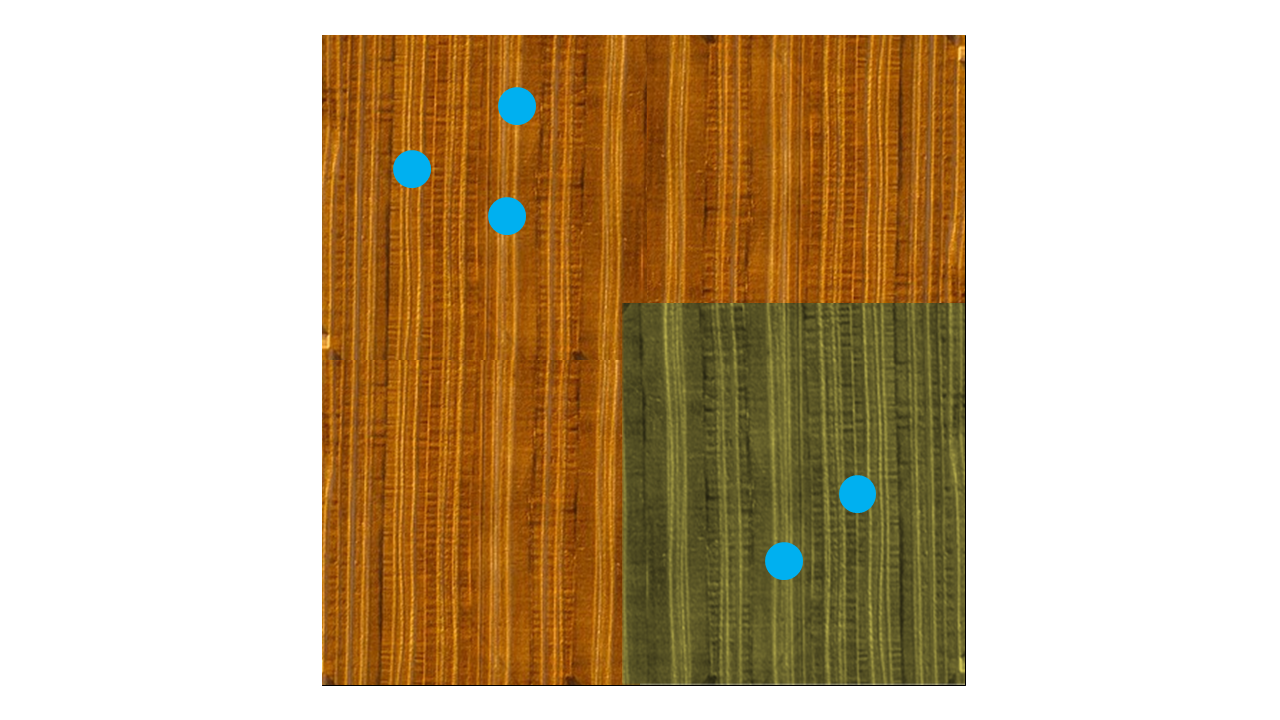}
        \label{fig:representativeness}}
        \subfigure[Diversity]{%
        \includegraphics[width=0.3\linewidth,trim=200 0 200 0, clip]{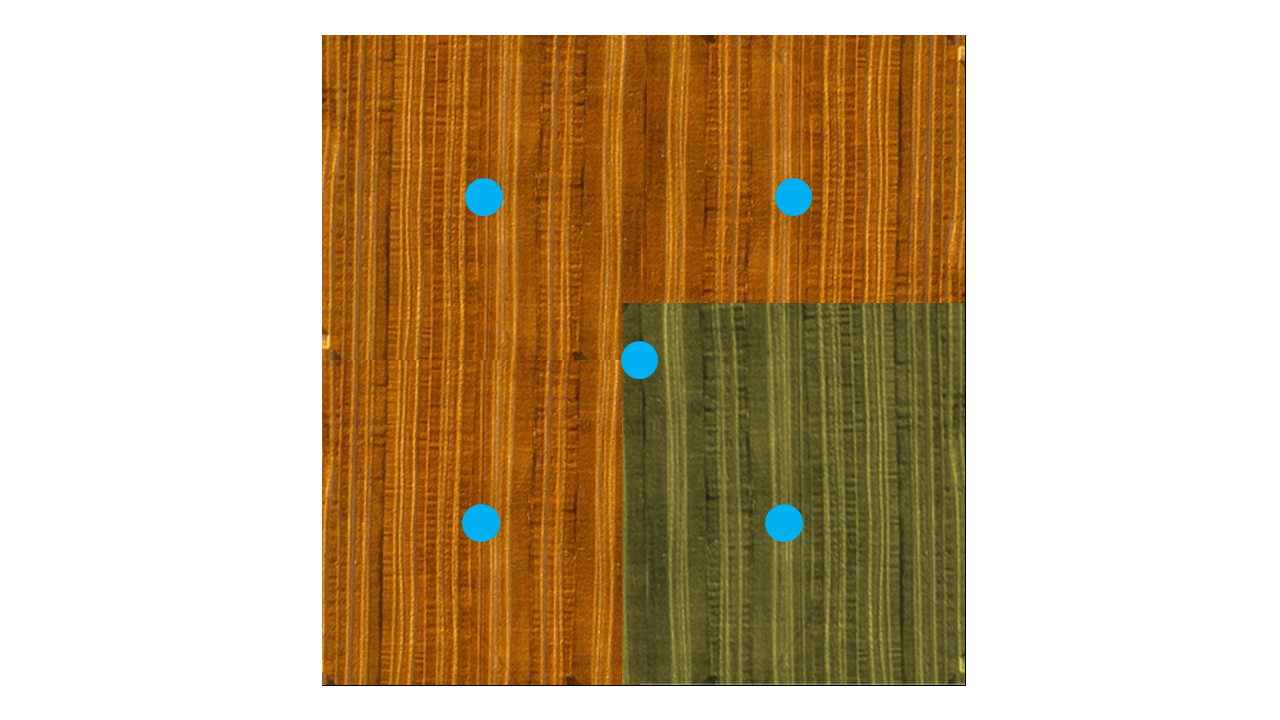}
        \label{fig:diversity}}
    \caption{Learning criteria: watering optimization problem.}\label{fig: learningcriteria}
\end{figure*}

\begin{enumerate}

    \item \textbf{Informativeness} measures the amount of information encoded by a certain sample. This means that the sampling policy is driven by the maximum likely contribution of queries that would significantly benefit the objective of the learning procedure.

    \item \textbf{Representativeness} quantifies the similarity of a sample or a group of samples with respect to a target sample representative of the target distribution. Thus, the sampling policy exploits the structure underlying the domain to direct the queries in locations where a sample can represent a large amount of neighbouring samples.

    \item \textbf{Diversity} estimates how well the queries are disseminated over the domain of samples. This is reflected in a sampling policy that selects samples scattering across the full domain, and prevents the concentration of queries in small local regions.

\end{enumerate}

Figure \ref{fig: learningcriteria} illustrates a watering optimization problem that attempts to clarify the peculiarities of each learning criteria. This simple toy problem requires to identify the areas of a wheat field where the crop is ripe and where it is still unripe for irrigation purposes. The learning goal is formalized as the identification of the area where the wheat is lower, which means an unripe cultivation and maximum requirements for irrigation. We assume that the learner can explore a maximum of five sites on the field during the procedure. A learner driven by the pure informativeness criterion (Figure \ref{fig:informativeness}) would uniquely sample the regions of the wheat field that are likely to provide the maximum amount of information to accomplish the given learning goal; accordingly, observations are placed where the height of the wheat is minimum and the demand for water is maximum: this maximizes the information on where it is strictly necessary to irrigate, but nothing is known about the regions where the wheat is higher and irrigation is not a priority. Conversely, a purely representative sampling (Figure \ref{fig:representativeness}) would probe the field by agglomerating observations to ensure the representativeness of the samples. This allows to partially know even areas where copious irrigation is not necessary, but increases the overall uncertainty given the small amount of samples for each agglomeration. If the learner pursues only the diversity of queries (Figure \ref{fig:diversity}), samples would scatter the field minimizing the maximum distance between measurements. Although this allows the queries to be distributed across the entire domain, the uncertainty is high as only one sample covers a respective area of the field.

\begin{figure}[t]
\centering
\includegraphics[width=0.6\linewidth,trim=80 50 140 50, clip]{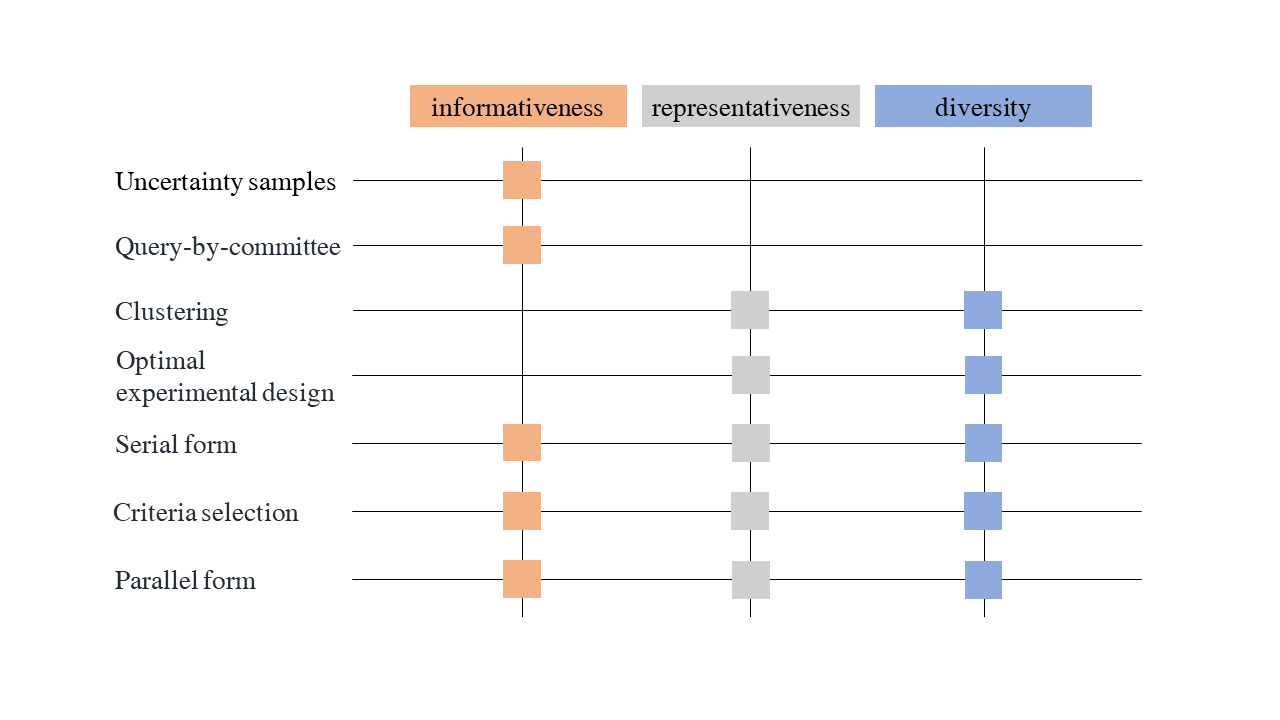}
\caption{Mapping methodologies to learning criteria.}
\label{fig:MappingLCtoIC} 
\end{figure}

The remaining of this section is dedicated to the revision and discussion of popular pool-based active learning schemes. We aim to provide a broad spectrum of approaches that exemplify the implementation of different learning criteria both individually and in combination. This permits to highlight the driving principles of learning procedures, and will help to better clarify the existing synergy between active learning and Bayesian optimization accounted in the following sections. Figure \ref{fig:MappingLCtoIC} summarizes the relationship between the methodologies reviewed in the following and the three learning criteria.

\subsubsection{Informativeness-Based} Learning procedures characterized by a pure informative criterion can be traced in uncertainty-based sampling policies. These approaches make the query decision based on the predictive uncertainty of the surrogate model, and seek to improve the density of samples in regions that exhibit the largest uncertainty with respect to a specific learning goal. Popular uncertainty-based active learning algorithms are uncertainty sampling and query-by-committee methods. Uncertainty sampling algorithms probe the domain to improve the overall accuracy of the surrogate model according to a measure of the predictive uncertainty. Examples include the quantification of the uncertainty associated with samples \cite{Lewis&Catlett1994}, and its alternatives as  margin-based \cite{BalcanAl2007}, least confident \cite{Li&Sethi2006} and entropy-based \cite{HolubAl2008} approaches. Other strategies define sampling policies which promote the minimization of the surrogate model predicted variance \cite{Cohn1993} to maximize, respectively, the decrease of loss augmenting the training set \cite{Settles2009}, and the gradient descend \cite{CaiAl2013}. Other uncertainty-based strategies are query-by-committee sampling schemes \cite{BurbidgeAl2007, ZhaoAl2006}, where the most informative sample to query is selected through the maximization of the disagreement between the predictions of a committee of surrogate models computed on subsets of the locations.

\subsubsection{Representativeness/Diversity-Based} Other pool-based active learning algorithms rely exclusively on representativeness and diversity learning frames: usually these learning criteria are implemented simultaneously in the learning procedure to drive the domain probing. This blend is justified by the mutual complementary relationship between representativeness and diversity: pure representativeness might concentrate the sampling in congregated representative domain regions without a proper dispersion of queries, while pure diversity might lead to the over-query of the domain and divert the learning procedure from the actual goals. The combination of both the learning criteria permits on one hand to leverage the representativeness of samples to accomplish a certain learning goal, on the other hand prevents the selection of redundant samples and the high densities of queries only in circumstanced regions of the domain. Representative/diversity-based algorithms include a multitude of approaches that are commonly classified in two main schemes: clustering methodology and optimal experimental design. The former clustering algorithms identify the most representative locations exploiting the underlying structures of the domain: the utility of samples is obtained as a function of their distance from the cluster centers. Popular examples include hierarchical clustering and k-center clustering. The former identifies a hierarchy of clusters based on the encoded information, and selects samples closer to the cluster centers \cite{Dasgupta&Hsu2008}; the latter determines a subset of k congruent clusters that together cover the sampling space and whose radius is minimized, and the best sample minimizes the maximum distance of any point to a center \cite{Sener&Savarese2017}. The latter optimal experimental design defines a sampling policy based on a transductive approach: the learning procedure conducts the queries through a data reconstruction framework that measures the samples representativeness based on the capacity to reconstruct the training dataset. The selection of the most representative sample comes from an optimization process that maximizes the local acquisition of information about the parameters of the surrogate model \cite{ChattopadhyayAl2013, FuAl2013, ReitmaierAl2015}.    

\subsubsection{Hybrid} Recent avenues explore the combination of both informativeness and representativeness/diversity learning criteria to combine the goal oriented query of the first, and the use of underlying structures preventing over-density of the second. Accordingly, combined-based algorithms integrates multiple learning criteria to improve the overall sampling performance. Those approaches are commonly classified into three main classes \cite{ZhaoAl2019, ZhanAl2021}: serial-form, criteria selection, and parallel-form approaches. Serial-form algorithms use a switching approach to take advantages from all the three learning criteria:  informativeness-based techniques are used to select a subset of highly informative samples, and then representativeness/diversity techniques identify the centers of the clusters on this subset as the querying locations \cite{ShenAl2004}. Criteria selection algorithms rely on a selection parameter informed by a measure of the learning improvement that suggests the appropriate learning criteria to be used during the procedure \cite{Hsu&Lin2015}. Both serial-form and criteria selection strategies combine the three learning criteria through a sequential approach where each criteria is used consecutively during the learning procedure. Parallel-form methods combine simultaneously multiple learning criteria: the utility of each sample is judged by weighting informativeness and representativeness/diversity at the same time; then, valuable samples are selected through a multi-objective optimization of the weights to maximize at the same time the improvement in terms of learning goals and the exploitation of potentially useful structures of the domain \cite{Li&Guo2013, Wang&Ye2015, TangAl2019}.

\subsection{Acquisition Functions and Infill Criteria}
\label{s: acquisition functions }

The synergy between active learning and Bayesian optimization relies on the substantial analogy between the learning criteria driving the active learning procedure and the infill criteria that characterize the Bayesian learning scheme. Infill criteria provide a measure of the information gain in terms of utility acquired evaluating a certain location of the domain. In Bayesian optimization, the acquisition function is formalized according to a certain infill criterion: this permits to quantify the merit of each sample with respect to a specific learning goal. Accordingly, the sample that maximizes the querying utility is observed to enrich the learning procedure toward this goal.

In particular, Bayesian learning schemes rely on two main infill criteria: global exploration ad local exploitation toward the optimum. The former exploration criterion concentrates the samples in regions of the domain where the uncertainty predicted by the surrogate is higher; this enhances the global awareness about the distribution of the objective function over the domain, but the resources might not be directed toward the goal of the procedure -- e.g. minimum of the objective function. The latter exploitation criterion condensates the samples in regions where the surrogate model indicates that the objective is likely to be located -- e.g. minimum of the Gaussian process mean function; exploitation realizes a goal-oriented sampling procedure that privileges the search for the objective without a potentially accurate knowledge of the overall distribution of interest. The dilemma between exploration and exploitation represents a key challenge to be carefully addressed. On one hand, a learning procedure based on pure exploration might use a large amount of samples to improve the overall accuracy of the surrogate model without searching toward the learning goal. On the other hand, an exploitation-based learner might anchor a high density of samples to a suboptimal local solution as a consequence of the information from an unreliable surrogate model. These extreme behaviours demonstrate the need to find a compromise between exploration and exploitation criteria.

In principle, infill criteria in Bayesian optimization are strongly related to the learning criteria commonly adopted in active learning. In particular: 

\begin{itemize}
    \item The concept of \textbf{exploration} is close to the \textbf{representativeness/diversity} criterion: both these learning schemes leverage underlying structures of the target distribution predicted by an accurate surrogate model to improve the awareness about the objective over the domain.  

    \item The concept of \textbf{exploitation} is close to the \textbf{informativeness} criterion: the learner directs the selection of samples toward the believed objective without considering the global behaviour of the objective over the domain.  
    
\end{itemize}

\begin{figure}
    \centering
\includegraphics[width=0.45\textwidth,trim=210 100 140 50, clip]{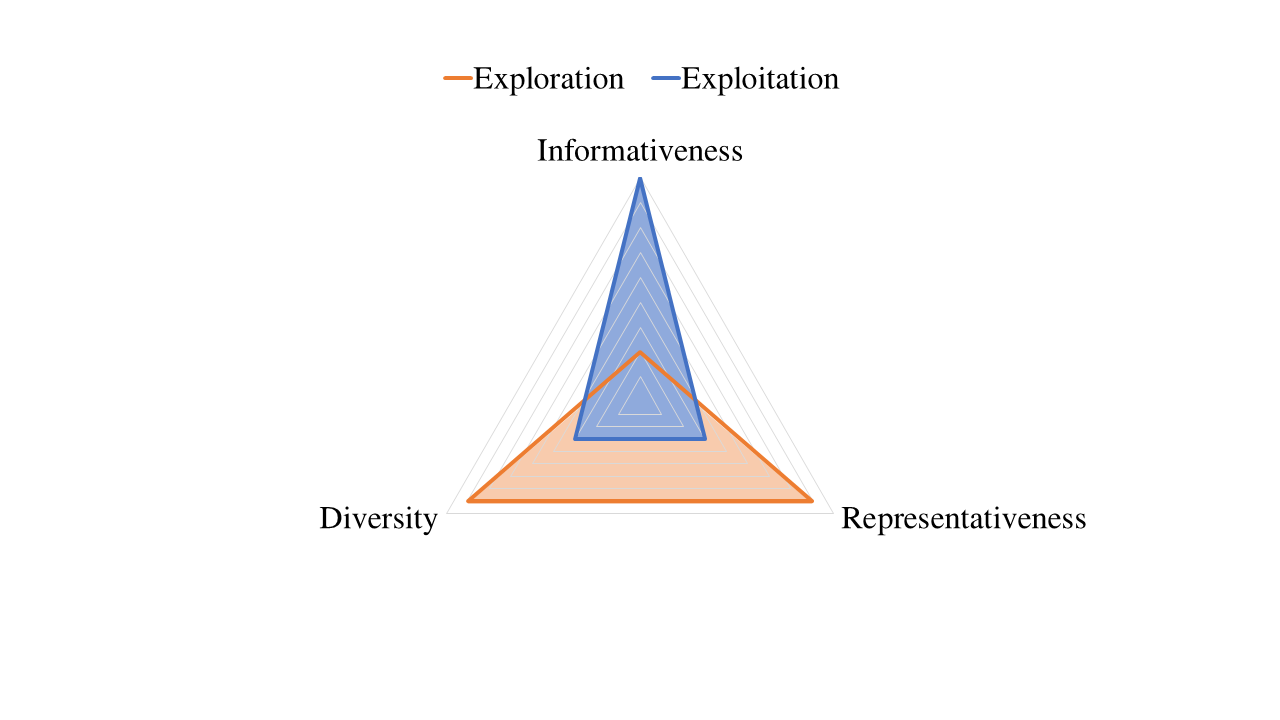}
\caption{Mapping of the learning criteria in active learning and infill criteria in Bayesian optimization. }
\label{fig:MappingLCandIC}
\end{figure}

Figure \ref{fig:MappingLCandIC} summarizes the mapping between infill criteria and learning criteria. The following sections discuss the formalization of (infill) active learning criteria for three popular formulations of Bayesian acquisition functions, namely the expected improvement (Section \ref{s:EI}), probability of improvement (Section \ref{s:PI}), and max-value entropy search (Section \ref{s: MES}).

\subsubsection{Expected Improvement}
\label{s:EI}

The Expected Improvement (EI) acquisition function quantifies the expected value of the improvement of the solution of the optimization problem achieved evaluating a certain location of the domain \cite{jones1998, mockus}. EI in the generic location $\DesignVar$ relies on the predicted improvement over the best solution of the optimization problem observed so far. Considering the Gaussian process as the surrogate model for Bayesian optimization, EI can be expressed as follows: 

\begin{equation} \label{e:EI}
\begin{split}
     \AF_{\EI}(\DesignVar) =  
     \StandDevGP(\DesignVar)[\Improv(\DesignVar)\CumDistFun(\Improv(\DesignVar)) + \ProbDensFun(\Improv(\DesignVar))]
\end{split}
\end{equation}
\noindent where $\Improv(\DesignVar) = (\ObjFun(\hat{\DesignVar}^*) - \MeanFunGP(\DesignVar))/\StandDevGP(\DesignVar)$ is the predicted improvement, $\hat{\DesignVar}^*$ is the current location of the best value of the objective sampled so far, $\MeanFunGP$ is the mean function and $\StandDevGP$ is the standard deviation of the GP, and $\CumDistFun(\cdot)$ and $\ProbDensFun(\cdot)$ are the cumulative distribution function and the probability density function of a standard normal distribution, respectively. The computation of $\AF_{\EI}(\DesignVar)$ requires limited computational resources and the first-order derivatives are easy to calculate: 

\begin{equation} \label{e:dEI1}
    \frac{\partial \AF_{\EI}(\DesignVar)}{\MeanFunGP(\DesignVar)} = - \CumDistFun (\Improv(\DesignVar))
\end{equation}

\begin{equation} \label{e:dEI2}
    \frac{\partial \AF_{\EI}(\DesignVar)}{\StandDevGP(\DesignVar)} = \ProbDensFun (\Improv(\DesignVar)).
\end{equation}

Both Equation \eqref{e:dEI1} and Equation \eqref{e:dEI2} demonstrate that $\AF_{\EI}(\DesignVar)$ is monotonic with respect to the increase of both the mean and the uncertainty of the GP surrogate model. This highlights a form of trade-off between exploration and exploitation: the formulation of the EI permits to balance the sampling in locations of the domain where is likely to have a significant improvement of the solution with respect to the current best solution, and the observations of regions where the improvement might be contained but the prediction is highly uncertain. In principle, it is possible to state that EI is driven by a combination of informativeness and representativeness/diversity criteria adopted in active learning. On one hand, the learner seeks to direct the computational resources toward the maximization of the learning contribution and achievement of the goal -- informativeness; on the other hand, the learner pursues the awareness of the objective distribution over the domain to improve the quality of the prediction and better drive the search -- representativeness/diversity. The predictive framework of the surrogate model regulates the learning thrusts privileging the one over the other on the basis of the information about the objective function acquired over the iterations.

\subsubsection{Probability of Improvement}
\label{s:PI}

The Probability of Improvement (PI) acquisition function targets the locations characterized by the highest probability of achieving the goal, based on the information from the current surrogate model \cite{kushner, jones2001}. PI measures the probability that the prediction of the surrogate model in the generic location is lower than the best observation of the objective function so far. Under the Gaussian process surrogate model, the PI acquisition function is computed in closed form as follows:

\begin{equation} \label{e:PI}
     \AF_{\PI}(\DesignVar) = \CumDistFun \left( \Improv(\DesignVar) \right)
\end{equation}

\noindent where $\CumDistFun(\cdot)$ is the cumulative distribution function of a standard normal distribution and $\DesignVar^*$ is the current location of the best value of the objective. Similarly to EI, also $\AF_{\PI}(\DesignVar)$ is inexpensive to compute and the evaluation of the first-order derivatives requires simple calculations: 
\begin{equation} \label{e:dPI1}
     \frac{\partial \AF_{\PI}(\DesignVar)}{\partial \MeanFunGP(\DesignVar)} = -\frac{1}{\StandDevGP(\DesignVar)} \ProbDensFun \left(\Improv(\DesignVar)\right)
\end{equation}

\begin{equation} \label{e:dPI2}
     \frac{\partial \AF_{\PI}(\DesignVar)}{\partial \StandDevGP(\DesignVar)} = -\frac{\Improv(\DesignVar)}{\StandDevGP(\DesignVar)} \ProbDensFun \left(\Improv(\DesignVar)\right)
\end{equation}

\noindent where $\ProbDensFun$ is the standard Gaussian probability density function. As demonstrated by Equation \eqref{e:dPI1}, regions of the input space characterized by lower values of the posterior mean of the GP are preferred for sampling, at fixed uncertainty of the surrogate. Moreover, Equation \eqref{e:dPI2} shows that if $\MeanFunGP(\DesignVar) < \ObjFun(\DesignVar^*)$ the regions characterized by lower uncertainty are preferred and, conversely, PI increases with uncertainty. Overall, the PI acquisition function can be considered as an exploitative scheme that determines the most informative location as the one that potentially produces a larger reduction of the minimum value of the objective function observed so far. This is achieved sampling regions where the surrogate model is reliable and characterized by lower levels of uncertainty. In principle, this sampling scheme makes PI in accordance with the informativeness criterion: the search toward the optimum is uniquely directed in regions of the domain that exhibit the higher probability of achieving the goal according to the emulator prediction.

\subsubsection{Entropy Search and Max-value Entropy Search}
\label{s: MES}

The Entropy Search (ES) acquisition function measures the differential entropy of the believed global minimum location of the objective function, and targets the reduction of uncertainty selecting the sample that maximizes the decrease of differential entropy \cite{hennig&Schuler2012}. The ES acquisition function is formulated as follows: 
 \begin{equation} \label{e:ES}
     \AF_{\ES} (\DesignVar) = \Entropy (\Probability(\MinDesignVar | \Dataset)) - \Expectation_{\ObjFun(\DesignVar) | \Dataset} [ \Entropy(\Probability(\MinDesignVar | \ObjFun(\DesignVar), \Dataset)) ]
 \end{equation}
 \noindent where $\Entropy (\Probability(\MinDesignVar))$ is the entropy of the posterior distribution at the current iteration in the location of the minimum of the objective function $\MinDesignVar$, and $\Expectation_{\ObjFun(\DesignVar)} [\cdot]$ is the expectation over $\ObjFun(\DesignVar)$ of the entropy of the posterior distribution at the next iteration on $\MinDesignVar$. Typically, the exact calculation of the second term of Equation \eqref{e:ES} is not possible and requires complex and expensive computational techniques to provide an approximation of $\AF_{\ES} (\DesignVar)$. 
 
 The Max-value entropy search (MES) \cite{wang_mes} acquisition function is derived from the ES acquisition function and allows to reduce the computational effort required to estimate Equation \eqref{e:ES} measuring the differential entropy of the minimum-value of the objective function:
 
  \begin{equation} \label{e:MES}
     \AF_{\MES} (\DesignVar) = \Entropy (\Probability(\ObjFun | \Dataset)) - \Expectation_{\ObjFun(\DesignVar) | \Dataset} [ \Entropy(\Probability(\ObjFun | \MinObjFun, \Dataset)) ]
 \end{equation}
 
\noindent where the first and the second term are now computed for the minimum value of the objective function $\MinObjFun$. This permits to simplify the computations and to approximate the second term through a Monte Carlo strategy \cite{wang_mes}. The analysis of the derivatives is not possible for the MES acquisition function since the formulation of the second term of Equation \eqref{e:MES} is intractable. 

As reported by Wang et al. \cite{wang_mes} in their experimental analysis, MES targets the balance between the exploration of locations characterized by higher uncertainty of the surrogate model, and the exploitation toward the believed optimum of the objective function. However, Nguyen et al. \cite{NguyenAl2022} demonstrate that MES might suffer from an imbalanced exploration/exploitation trade-off due to noisy observations of the objective function, and to the discrepancy in the computation of the mutual information in the second term of Equation \eqref{e:MES}. As a result, MES might over-exploit the domain in presence of noise in measurements, and over-explore when the discrepancy in the evaluation issue determines a pronounced sensitivity to the uncertainty of the surrogate model. Overall, the adaptive sampling scheme determined by the MES acquisition function follows both the informativeness and the representativeness/diversity learning criteria: the most promising sample is ideally selected targeting the balance between the search toward the believed minimum predicted by the emulator, and the decrease of uncertainty about the objective function distribution.

\subsection{Learning Criteria with Multiple Oracles}  \label{s:Active learning with multiple oracles}

Most of the active learning paradigms rely on a unique and supposed omniscient source of information about the target distribution. This oracle is iteratively queried by the learner to evaluate the value of the distribution in certain locations, and is assumed that its prediction is exact. In many other scenarios, the learner can elicit information from multiple imperfect oracles at different levels of reliability, accuracy and cost. Accordingly, the active learning community introduced a multitude of annotator-aware algorithms which are capable of efficiently learning from multiple sources of information. This requires to make an additional decision during the learning procedure: the learner has to select at each iteration the most useful sample and the associated information source to query. In this context, the original learning criteria of informativeness and representativeness/diversity (Section \ref{s:learning criteria}) evolve and extend to quantify the utility of querying the domain with a certain level of accuracy and associated cost:

\begin{enumerate}

    \item \textbf{Informativeness} seeks to maximize the amount of information to decide the sample and information source to query. Thus, the learner might privilege the evaluations from accurate and yet costly oracles to capitalize from high-quality information and potentially reach the objective.  

    \item \textbf{Representativeness} attempts to identify underlying structures of the domain to better inform the search procedure. In this case, the decision making process might prefer to interrogate less expensive sources of information to contain the required effort, especially if cheap predictions of the target distribution exhibit good correlation with the estimate of the accurate oracle.  

    \item \textbf{Diversity} scatters the sampling effort over the domain to pursue a proper distribution of evaluations and augment the awareness about the target distribution. This might be favored by a major use of less accurate predictions of the target distribution, which are more likely to well address the cost/effectiveness trade-off during the diversity sampling.   
    
\end{enumerate}

The remaining of this section provides an overview of different multiple oracles active learning methodologies to present and further clarify popular extensions of the learning criteria to a multi-oracle setting.

Typically, active learning paradigms are extended to the multiple-oracle setting through relabeling, repeating-labeling, probabilistic and transfer knowledge, and cost-aware algorithms. Relabeling approaches query samples multiple times using the library of sources of information available, and the final query is obtained via majority voting \cite{ZhaoAl2011}. Popular methodologies following this scheme pursue the identification of a subset of oracles according to the proximity of their upper confidence bound to the maximum upper confidence bound, and apply the majority voting technique only considering the queries of this informative subset \cite{DonmezAl2009}. Other multi-oracle active learning methods use a repeating-labeling procedure: the learner integrates the repeated -- often noisy -- prediction of the oracles to improve the quality of the evaluation process and the accuracy of the surrogate model learned from data \cite{IpeirotisAl2014}. Both relabeling and repeating-labeling approaches share a common drawback: the same unknown sample is evaluated multiple times with different oracles, which results in a sub-optimal usage of the available sources of information. Probabilistic and transfer learning methodologies attempt to overcome this limitation. Probabilistic frameworks rely on surrogate models specifically conceived for the multi-source scenario that provides a predictive framework to estimate the accuracy of each oracle in the evaluation of samples over the domain \cite{YanAl2011, YanAl2012}. Transfer knowledge approaches enhance the simultaneous selection of the most informative location to sample and the associated most profitable source to query; this is achieved through the transfer of knowledge from samples not evaluated in auxiliary domains to support the estimate of the oracle reliability \cite{FangAl2014}. Recent advancements in multiple oracles active learning are cost-effective algorithms, where the cost of an oracle is evaluated considering both the overall reliability of the prediction and the quality of samples in specific locations \cite{HuangAl2017, YuAl2020, GaoAl2020}. The cost-effectiveness property enhances the use of computational resources for the evaluation of samples, and targets the search toward the learning objectives while guarantees an optimal trade-off between evaluation accuracy and computational cost.

From the examined literature, the three learning criteria appear frequently coupled together during the learning procedure with multiple sources to query. This appears as a natural evolution of what has already been observed in the literature for active learning with single information source: the overall learning procedure usually benefits from a balanced learning scheme driven by informativeness and representativeness/diversity. In particular, informativeness permits to direct the search toward the learning goal, while representativeness/diversity augments the learner awareness about the target distribution over the domain; the combination of these learning criteria -- in different measures -- contributes to improve the performance of the active learning algorithms by efficiently using the computational resources and the information from multiple oracles.

\subsection{Multifidelity Acquisition Functions and Infill Criteria}
\label{s: multifidelity acquisition function}

This section further investigates and highlights the synergy between active learning and Bayesian optimization for the specific case of multiple sources of information used to accomplish the learning goal. 
Similarly to the single source setting, this symbiotic relationship is revealed through common principles characterizing the infill criteria in multifidelity Bayesian optimization and the learning criteria in active learning with multiple oracles. The multifidelity scenario imposes an additional decision to be made: the learner has to identify the appropriate information source to query according to an accuracy/cost trade-off. This is reflected in the formalization of infill criteria capable of defining an efficient and balanced sampling policy, targeting either the wise selection of the samples and the levels of fidelity which ensure the maximum benefits with the minimum cost. Accordingly, the multifidelity acquisition function formalizes an adaptive sampling scheme based on one or multiple infill criteria to quantify the utility of querying a location of the domain with a specific level of fidelity. 

Based on these considerations, the exploration and exploitation infill strategies are extended according to the peculiarities of the multifidelity setting: 

\begin{itemize}

    \item \textbf{Exploration} is close to the \textbf{representativeness/diversity} criterion and defines a sampling policy that incentivizes the overall reduction of the surrogate uncertainty. Accordingly, the selection of the appropriate level of fidelity is driven by a trade-off between accuracy and evaluation cost. This might be accomplished through less-expensive low-fidelity information to contain the demand for computational resources during exploration. 

    \item \textbf{Exploitation} is close to the \textbf{informativeness} criterion and concentrates the sampling process in the regions of the domain where optimal solutions are likely to be located. For this purpose, the learner might emphasize the use of accurate evaluations of the target function to refine the solution of the learning procedure toward the specific goal. 
    
\end{itemize}

Similarly to the acquisition functions in Bayesian optimization (Section \ref{s: acquisition functions }), the symmetry between informativeness and exploitation criteria, and between representativeness/diversity and exploration criteria is preserved in the multifidelity setting. The following sections are dedicated to the review and discussion of popular multifidelity acquisition function, namely the multifidelity expected improvement (Section \ref{s: MFEI}), multifidelity probability of improvement (Section \ref{s:MFPI}) and multifidelity max-value entropy search (Section \ref{s:MFMES}). The goal is to highlight the equivalent principles driving both the learning schemes, and further clarify the elements that encode the symbiotic relationship that exists between multifidelity Bayesian optimization and multi-oracle active learning.

\subsubsection{Multifidelity Expected Improvement}
\label{s: MFEI}

The Multifidelity Expected Improvement (MFEI) extends the expected improvement acquisition function to define a learning scheme in the multifidelity setting as follows \cite{HuangAl2006}: 

\begin{equation} \label{e:MFEI}
 \AF_{\MFEI}(\DesignVar, \LevFid) =  \AF_{\EI}(\DesignVar, \MaxLevFid) \alpha_1(\DesignVar,\LevFid) \alpha_2(\DesignVar,\LevFid)  \alpha_3(\LevFid)
\end{equation} 
\noindent where $\AF_{\EI}(\DesignVar, \MaxLevFid)$ is the expected improvement illustrated in Equation \eqref{e:EI} and evaluated with the highest level of fidelity $\MaxLevFid$, and the utility functions $\alpha_1$, $\alpha_2$ and $\alpha_3$ are defined as follows: 

\begin{equation}\label{e:MFEI1}
    \alpha_1 (\DesignVar, \LevFid) = corr \left[ \ObjFun^{(\LevFid)}, \ObjFun^{(\MaxLevFid)} \right] 
\end{equation}  
\begin{equation} \label{e:MFEI2}
    \alpha_2 (\DesignVar, \LevFid) = 1 - \frac{\StandDevNoise}{\sqrt{\StandDevGP^{2(\LevFid)} (\DesignVar) + \StandDevNoise^{2}}}
\end{equation}
\begin{equation} \label{e:MFEI3} 
    \alpha_3 (\LevFid) = \frac{\CompCost^{(\MaxLevFid)}}{\CompCost^{(\LevFid)}}.
\end{equation}

The first element $\alpha_1$ is the posterior correlation coefficient between the level of fidelity $\LevFid$ and the high-fidelity level $\MaxLevFid$, and accounts for the reduction of the expected improvement when a sample is evaluated with a low fidelity model. This term reflects a measure of the informativeness of the $\LevFid$-th source of information in the location $\DesignVar$, and balances the amount of improvement achievable evaluating the high-fidelity level $\MaxLevFid$ with the reliability of the prediction associated with the level of fidelity $\LevFid$. Accordingly, $\alpha_1$ modifies the learning scheme by adding a penalty in the formulation that reduces the $\AF_{\MFEI}$ when $1 \leq \LevFid<\MaxLevFid$: this includes awareness about the increase of uncertainty associated with a low-fidelity prediction. The second element $\alpha_2$ is conceived to adjust the expected improvement when the output evaluated with the $\LevFid$-th level of fidelity contains random errors. This is equivalent to consider the reduction of the uncertainty of the Gaussian process prediction after a new evaluation of the objective function is added to the dataset $\Dataset$. This function allows to improve the robustness of $\AF_{\MFEI}$ when the representation of $\ObjFun^{(\LevFid)}$ at different levels of fidelity is affected by noise in the measurements. The third element $\alpha_3$ is formulated as the ratio between the computational cost of the high-fidelity level $\MaxLevFid$ and the $\LevFid$-th level of fidelity. This permits to balance the informative contributions of high- and a lower-fidelity observations and the related computational resources required for the evaluation. The effect of this term is to encourage the use of low-fidelity representations if almost the same expected improvement can be achieved with a high-fidelity evaluation. This wisely directs the use of computational resources to achieve the representativeness/diversity of samples, and prevents a massive use of expensive accurate queries during the exploration phases.

\subsubsection{Multifidelity Probability of Improvement}
\label{s:MFPI}

The Multifidelity Probability of Improvement (MFPI) acquisition function provides an extended formulation of the probability of improvement suitable for the multifidelity scenario as follows \cite{RuanAl2020}: 

\begin{equation} \label{eq:MFPI}
 \AF_{\MFPI}(\DesignVar,\LevFid)= \AF_{\PI}(\DesignVar, \MaxLevFid) \eta_{1}(\DesignVar,\LevFid) \eta_{2}(\LevFid) \eta_{3}(\DesignVar,\LevFid) 
\end{equation}

\noindent where the PI acquisition function (Equation \eqref{e:PI}) is computed considering the highest-fidelity level $\MaxLevFid$ available, and the utility function $\eta_1$, $\eta_2$ and $\eta_3$ are defined as follows: 

\begin{equation}\label{e:MFPI1}
    \eta_1 (\DesignVar, \LevFid) = corr \left[ \ObjFun^{(\LevFid)}, \ObjFun^{(\MaxLevFid)} \right] 
\end{equation}  

\begin{equation} \label{e:MFPI2}
    \eta_2 (\LevFid) = \frac{\CompCost^{(\MaxLevFid)}}{\CompCost^{(\LevFid)}}
\end{equation}
\begin{equation} \label{e:MFPI3} 
    \eta_3 (\DesignVar,\LevFid) = \prod_{i=1}^{\IndNumObs_{\LevFid}} \left[1-R\left(\DesignVar,\DesignVar_i^{(\LevFid)}\right)\right].
\end{equation}

The first term $\eta_1$ shares the same formalization of the utility function $\alpha_1$ in Equation \eqref{e:MFEI1}, and accounts for the increase of uncertainty associated with low-fidelity representations $1 \leq \LevFid < \MaxLevFid$ if compared with the high-fidelity output $\MaxLevFid$. This reduces the probability of improvement if a low-fidelity representation is queried in a specific location of the input space $\DesignVar$. As already highlighted in Section \ref{s: MFEI}, $\eta_1$ incentivizes a form of informativeness learning where the information source is selected according to its capability to accurately represent the objective function. Similarly, the second utility function $\eta_2$ is also included in the multifidelity expected improvement in Equation \eqref{e:MFEI3} as the $\alpha_3$ term. This element balances the computational costs and the informative contributions achieved through the $\LevFid$-th level of fidelity. This prevents the rise of computational demand produced by the over-exploitative nature of the probability of improvement (Section \ref{s:PI}): $\eta_2$ encourages the use of fast low-fidelity data if the discrepancy between the $\LevFid$-th level of fidelity and the high-fidelity $\MaxLevFid$ -- quantified by $\eta_1$ -- is not significant. The third element $\eta_3$ is the sample density function computed as the product of the complement to unity of the spatial correlation function $R\left(\cdot\right)$ \cite{liu2018sequential} evaluated for the $\IndNumObs_{\LevFid}$ samples considering the $\LevFid$-th level of fidelity. This term reduces the probability of improvement in locations with an high sampling density -- over exploitation of the domain -- to prevent the clustering of data. Accordingly, $\eta_3$ promotes a form of representativeness/diversity learning scheme and encourages the exploration to augment the awareness about the domain structure.

\subsubsection{Multifidelity Entropy Search and Multifidelity Max-Value Entropy Search}

\label{s:MFMES}

The Multifidelity Entropy Search (MFES) acquisition function is formulated extending the entropy search acquisition function to query multiple sources of information \cite{ZhangAl2017mf}

 \begin{equation} \label{e:MFES}
 \begin{split}
     \AF_{\MFES} (\DesignVar)  =  \Entropy (\Probability(\MinDesignVar | \Dataset)) - \Expectation_{\ObjFun^{(\LevFid)}(\DesignVar) | \Dataset} [ \Entropy(\Probability(\MinDesignVar | \ObjFun^{(\LevFid)}(\DesignVar), \Dataset)) ]
 \end{split}
 \end{equation}
 
 \noindent where the expectation term $\Expectation_{\ObjFun^{(\LevFid)}(\DesignVar)} [\cdot]$ considers multiple levels of fidelity $\LevFid = 1,...,\MaxLevFid$. Similarly to the entropy search acquisition function, the computation of the expectation in Equation \eqref{e:MFES} is not possible in closed-form and requires an intensive procedure to provide a reliable approximation. 
 
 The Multifidelity Max-Value Entropy Search (MFMES) acquisition function can be formulated extending the max-value entropy search to a multifidelity setting as follows \cite{TakenoAl2020}: 

  \begin{equation} \label{e:MFMES}
  \begin{split}
     \AF_{\MFMES} (\DesignVar) = [ \Entropy (\Probability(\ObjFun^{(\LevFid)} | \Dataset)) - \Expectation_{\ObjFun^{(\LevFid)}(\DesignVar) | \Dataset} [ \Entropy(\Probability(\ObjFun^{(\LevFid)} | \ObjFun^{*(\MaxLevFid)}, \Dataset)) ]]/\CompCost^{(\LevFid)}
 \end{split}
 \end{equation}
 
 \noindent where the differential entropy is measured for the minimum value of the objective function $\ObjFun^{*(\MaxLevFid)}$ considering the high-fidelity representation $\MaxLevFid$. In this case, the approximation of the expectation term in Equation \eqref{e:MFMES} relies on a Monte Carlo strategy that allows to contain the computational cost if compared with the procedure used for the MFES acquisition function \cite{TakenoAl2020}.
 
In the multifidelity scenario, the MFMES acquisition function measures the information gain obtained evaluating the objective function $\ObjFun^{(\LevFid)}(\DesignVar)$ in a certain location $\DesignVar$ and associated level of fidelity $\LevFid$ with respect to the global minimum of the objective function. This can be interpreted as an informativeness-driven learning based on the reduction of the uncertainty associated with the minimum value of the objective $\ObjFun^{*(\MaxLevFid)}$ through the observation $\ObjFun^{(\LevFid)}(\DesignVar)$, where this uncertainty is measured as the differential entropy associated with the $\LevFid$-th level of fidelity. At the same time, the information gain is also sensitive to the accuracy of the surrogate predictive framework, and realizes a form of representativeness/diversity balance to improve the awareness about the distribution of the objective function over the domain. The sensitivity to the computational cost $\CompCost^{(\LevFid)}$ of the $\LevFid$-th level of fidelity is introduced in Equation \eqref{e:MFMES} to balance the quality of the source -- quantified by the information gain -- and the demand for computational resources.

\section{Experiments} \label{s:Experiments}

This section investigates and compares the performance of the acquisition functions for both single-fidelity and multifidelity Bayesian optimization considering a set of benchmark problems conceived to stress those algorithms. The objective is to highlight advantages and opportunities offered by different learning principles over challenging mathematical properties of the objective function, which are frequently encountered in real-world engineering and scientific problems. \cite{mainini2022analytical}. In particular, this comparative study considers the expected improvement (Section \ref{s:EI}), probability of improvement (PI) (Section \ref{s:PI}), and Max-Value Entropy Search (MES) (Section \ref{s: MES}) for the single-fidelity frameworks, and their multifidelity counterparts Multifidelity Expected Improvement (MFEI) (Section \ref{s: MFEI}), Multifidelity Probability of Improvement (MFPI) (Section \ref{s:MFPI}) and Multifidelity Max-Value Entropy Search (MFMES) (Section \ref{s:MFMES}).

We impose the same initialization conditions for both the single-fidelity and the multifidelity algorithms. This initial setting includes: (i) the initial dataset of $\NumObs_0^{(\LevFid)}$ samples for each level of fidelity $\LevFid$ to compute the prior surrogate model of the objective function, (ii) the computational cost assigned to each level of fidelity $\CompCost^{(\LevFid)}$, and (iii) the maximum computational budget $\Budget_{max}$ allocated for each benchmark problem defined linearly with the dimensionality $\Dim$ of the problem $\Budget_{max}= 100 \Dim$. The initial dataset $\NumObs_0^{(\LevFid)}$ is obtained through Latin hypercube sampling for all the numerical experiments \cite{mckay1992latin} to ensure the full coverage of the range of the optimization variables. The computational budget $\Budget = \sum \CompCost_{\IterOpt}^{(\LevFid)}$ is quantified as the cumulative computational cost used during the optimization at each iteration $\IterOpt$. 

All the methods are based on the Gaussian processes surrogate model and its extension to the multifidelity setting. We implement the squared exponential kernels for all the GP covariances, and use the maximum likelihood estimation approach to optimize the hyperparameters of the kernel and the mean function of the GP \cite{sobester2008engineering}.

\subsection{Benchmark Problems}
\label{s: Benchmarks}

The following set of benchmark problems is specifically conceived to investigate the capabilities of different learning criteria over challenging mathematical properties of the objective function \cite{mainini2022analytical}. In particular, the experimental settings include a variety of attributes that can be traced in real-world optimization problems, namely local and global behaviours, non-linearities and discontinuities, multimodality and noise. The set of problems consist of several objective functions such as the Forrester continuous and discontinuous, the Rosenbrock increasing the domain dimensionality, the Rastrigin shifted and rotated, the Agglomeration of Locally Optimized Surrogate (ALOS), a coupled spring-mass optimization problem and the noisy Paciorek function.

\subsubsection{Forrester Function}
\label{s:Forrester}

\begin{figure*}[t!]
    \centering
     \subfigure[Forrester]{%
        \includegraphics[width=0.35\linewidth,trim=220 0 245 0, clip]{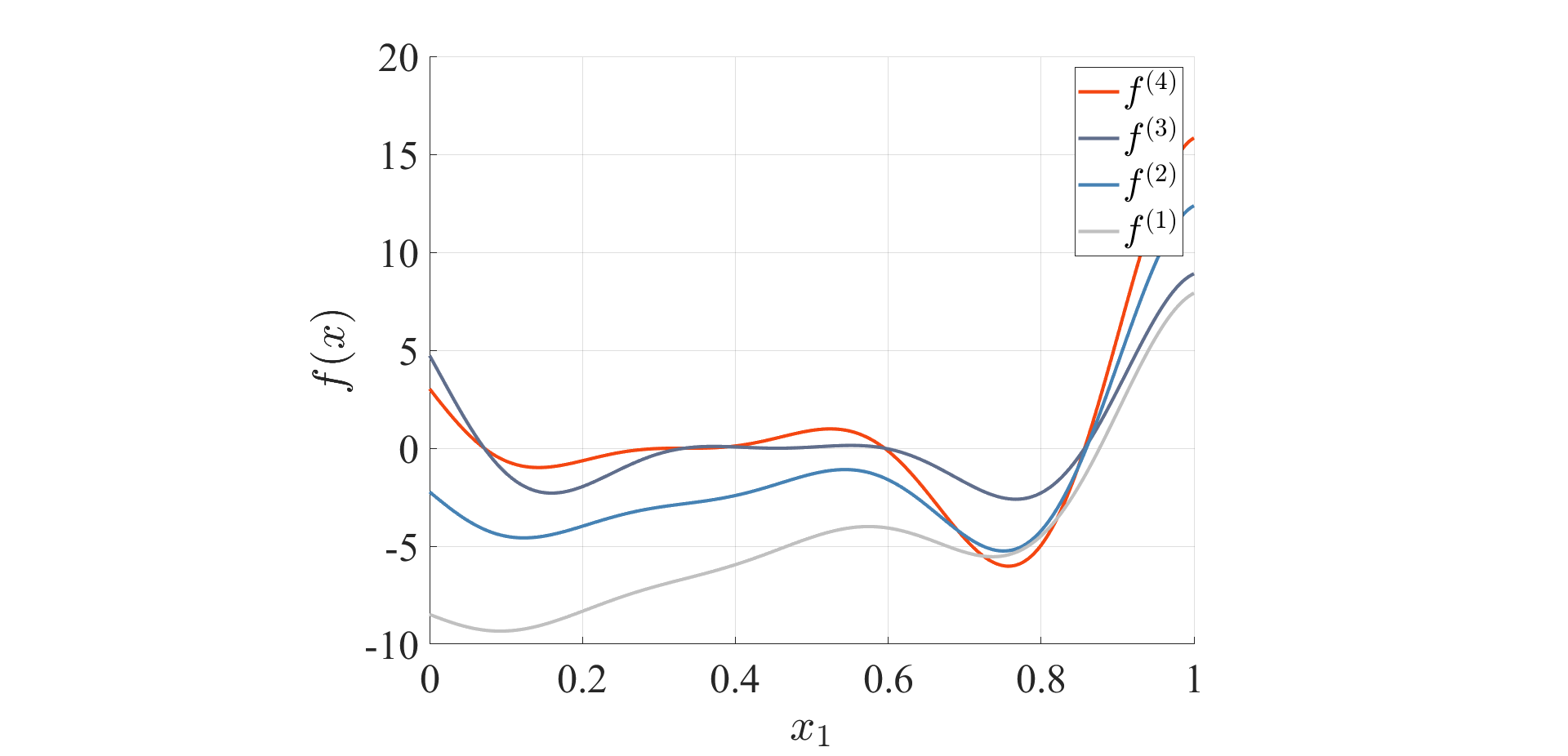} 
        \label{fig:ForrPlot}}
        \subfigure[Jump Forrester]{%
        \includegraphics[width=0.35\linewidth,trim=220 0 245 0, clip]{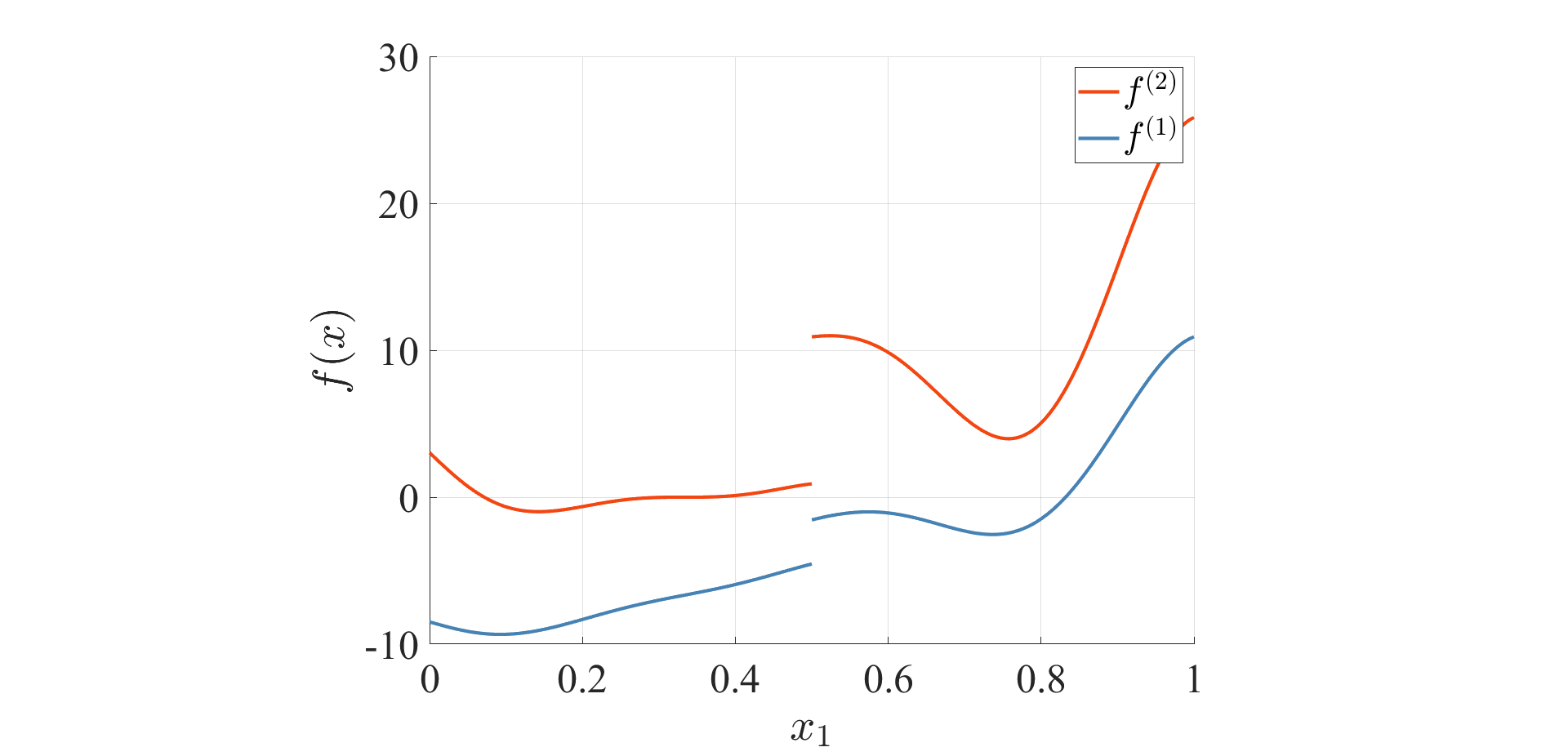}
        \label{fig:JumpForrPlot}}
        
    \caption{Forrester function benchmark problems}\label{fig: ForrBenchPlot}
\end{figure*}

The Forrester function is a popular test-case to investigate the performance of different learning strategies over a non-linear one-dimensional distribution characterized by local behaviours. This benchmark problem guarantees an high interpretability of the results thanks to the one-dimensional nature of the objective function. The search domain is bounded between $ \DesignSpace = [0,1]$ and four levels of fidelity are available during the optimization:


\begin{equation}
    \ObjFun^{(4)}(\DesignVar)= (6\DesignVar-2)^2\sin(12\DesignVar-4)
\end{equation}

\begin{equation}
     \ObjFun^{(3)}(\DesignVar)=(5.5\DesignVar-2.5)^2\sin(12\DesignVar-4)
\end{equation}

\begin{equation}
    \ObjFun^{(2)}(\DesignVar)=0.75 \ObjFun^{(4)}(\DesignVar)+5(\DesignVar-0.5)-2
\end{equation}

\begin{equation}
    \ObjFun^{(1)}(\DesignVar)=0.5 \ObjFun^{(4)}(\DesignVar)+10(\DesignVar-0.5)-5
\end{equation}

\noindent where $\ObjFun^{(4)}$ is the high-fidelity function and the levels of fidelity $\LevFid=1,2,3,4$ increase with the accuracy of the representations. Figure \ref{fig:ForrPlot} reports the four levels of fidelity for the Forrester function over the search domain. The analytical minimum of the Forrester function is equal to $\ObjFun^{*(4)} = -6.0207$ and it is located at the domain point $\MinDesignVar= 0.7572$.

\subsubsection{Jump Forrester Function}
\label{s: disc forrester}

The jump Forrester function introduces a discontinuity in the formulation of the Forrester function to investigate the capabilities of the learning schemes to refine the surrogate model and capture the instantaneous variation of the objective function over the domain. This scenario can often occur in real problems where the phenomena of interest -- e.g. physical quantity of interest in engineering -- evolves over the domain and determines large variations of the objective function values. Figure \ref{fig:JumpForrPlot} reports the two levels of fidelity that are available during the search procedure:


\begin{equation} \label{e:forrester_disc}
\begin{split}
    \ObjFun^{(2)}(\DesignVar) =  \left\{ \begin{array}{lc}
    (6\DesignVar-2)^2sin(12\DesignVar-4), & 0\leq \DesignVar \leq 0.5 \\
    (6\DesignVar-2)^2sin(12\DesignVar-4)+10, & 0.5<\DesignVar \leq 1
   \end{array} \right. \\
 \end{split}
 \end{equation}

 \begin{equation}
 \begin{split}
   \ObjFun^{(1)}(\DesignVar)&= \left\{ \begin{array}{lc} 0.5\ObjFun^{(2)}(\DesignVar)+10(\DesignVar-0.5)-5, & 0\leq \DesignVar\leq 0.5\\
   0.5\ObjFun^{(2)}(\DesignVar)+10(\DesignVar-0.5)-2 & 0.5< \DesignVar\leq 1 \end{array} \right.
\end{split}
\end{equation}

\noindent where $\ObjFun^{(2)}$ is the high-fidelity information source. The optimum is located at $\MinDesignVar=0.75724876$ corresponding to a value of the objective equal to $\ObjFun^{*(2)} = -0.9863$.

\subsubsection{Rosenbrock Function}
\label{s: Rosenbrock function}

\begin{figure*}[t!]
    \centering
     \subfigure[Rosenbrock $f^{(3)}$, $f^{(2)}$]{%
        \includegraphics[width=0.35\linewidth,trim=0 0 0 0, clip]{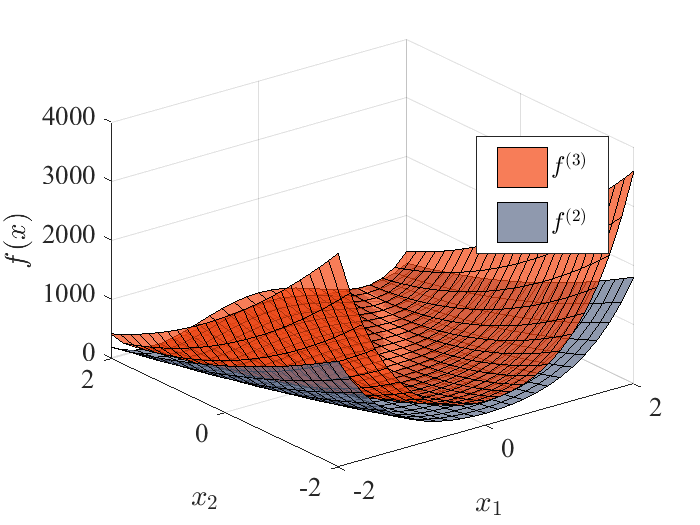} 
        \label{fig:Ros23}}
        \subfigure[Rosenbrock $f^{(3)}$, $f^{(1)}$]{%
        \includegraphics[width=0.35\linewidth,trim=0 0 0 0, clip]{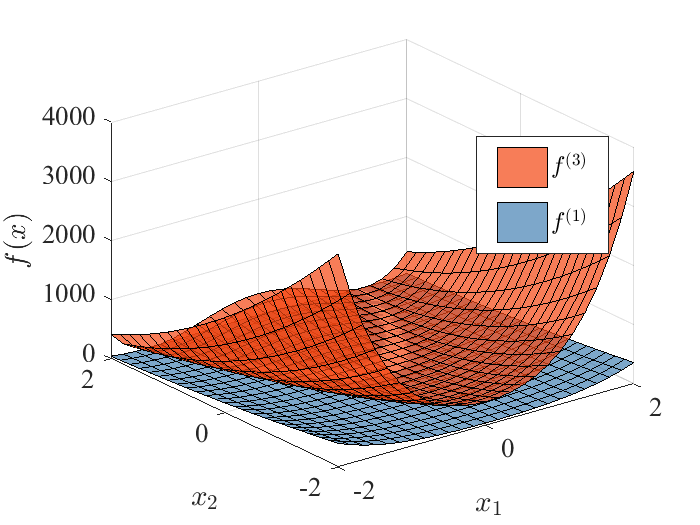}
        \label{fig:Ros13}}
        
    \caption{Rosenbrock function benchmark problem over the $\Dim=2$ dimensional domain}\label{fig: RosenbrockPlot}
\end{figure*}

The Rosenbrock function permits to investigate the learning criteria over a non-convex objective function that allows for parametric scalability over the domain $\DesignSpace = [-2,2]^\Dim$ where $\Dim$ is the dimensionality of the input space. A library of three levels of fidelity is available (Figure \ref{fig: RosenbrockPlot}):     

\begin{equation}
    \ObjFun^{(3)}(\DesignVar)= \sum_{i=1}^{\Dim-1}100(\DesignVar_{i+1}-\DesignVar_i^2)^2+(1-\DesignVar_i)^2
\end{equation}

\begin{equation}
\begin{split}
    \ObjFun^{(2)}(\DesignVar) = \sum_{i=1}^{\Dim -1} 50(\DesignVar_{i+1}-\DesignVar_i^2)^2  +(-2-\DesignVar_i)^2-\sum_{i=1}^\Dim 0.5\DesignVar_i
\end{split}
\end{equation}

\begin{equation}
    \ObjFun^{(1)}(\DesignVar)= \frac{\ObjFun^{(3)}(\DesignVar)-4-\sum_{i=1}^{\Dim} 0.5 \DesignVar_i}{10+\sum_{i=1}^{\Dim} 0.25 \DesignVar_i}
\end{equation}


\noindent where the high-fidelity function is $\ObjFun^{(3)}$ and the lower-fidelities are obtained using a transformation of $\ObjFun^{(3)}$ based on linear additive and multiplicative factors. The analytical minimum is located at $\MinDesignVar = [1,1]^\Dim$ and corresponds to a value of the objective function $\ObjFun^{*(3)} = 0$. The scalability of the Rosenbrock function allows to test the performance of the methods increasing the dimensionality of the input space. In this study, we consider the cases $\Dim=2,5,10$.

\subsubsection{ALOS Functions}
\label{s: ALOS}

The Agglomeration of Locally Optimized Surrogate (ALOS) is a heterogeneous and non-polynomial function defined on unit hypercubes up to three dimensions useful to assess the accuracy of surrogate models in presence of localized behaviours. In particular, the ALOS function reproduces a real-world scenario where the objective function is characterized by oscillatory phenomena at different frequency distributed along the domain. We consider two levels of fidelity and increasing dimensionality of the input space $\Dim = 1,2,3$. For $\Dim = 1$ the ALOS function is formalized as follows:

\begin{align}\left\{ \begin{array}{ll}
\label{e:heterogeneous1D}
    \ObjFun^{(2)}(\DesignVar) &= \sin[30(\DesignVar-0.9)^4]\cos[2(\DesignVar-0.9)] +(\DesignVar-0.9)/2\\
     \ObjFun^{(1)}(\DesignVar)&= (\ObjFun^{(2)}(\DesignVar)-1.0+\DesignVar)/(1.0+0.25\DesignVar) 
    \end{array} \right. 
\end{align}

\noindent and for $\Dim = 2,3$ is formulated as:   

\begin{align} \left\{ \begin{array}{ll}
\label{e:heterogeneous_nD}
   \ObjFun^{(2)}(\DesignVar)&=
   \sin[21(\DesignVar_1-0.9)^4]\cos[2(\DesignVar_1-0.9)]
   +(\DesignVar_1-0.7)/2+\sum_{i=2}^\Dim i\DesignVar_i^i\sin\left(\prod_{j=1}^i\DesignVar_j\right)\\
    \ObjFun^{(1)}(\DesignVar)&=
    (\ObjFun^{(2)}(\DesignVar)-2.0+\sum_{i=1}^\Dim\DesignVar_i)/(5.0 +\sum_{i=1}^2 0.25i\DesignVar_i-\sum_{i=3}^\Dim 0.25i\DesignVar_i)
  \end{array} \right. 
\end{align}

\begin{figure*}[t!]
    \centering
     \subfigure[ALOS $\Dim=1$]{%
        \includegraphics[width=0.35\linewidth,trim=220 0 245 0, clip]{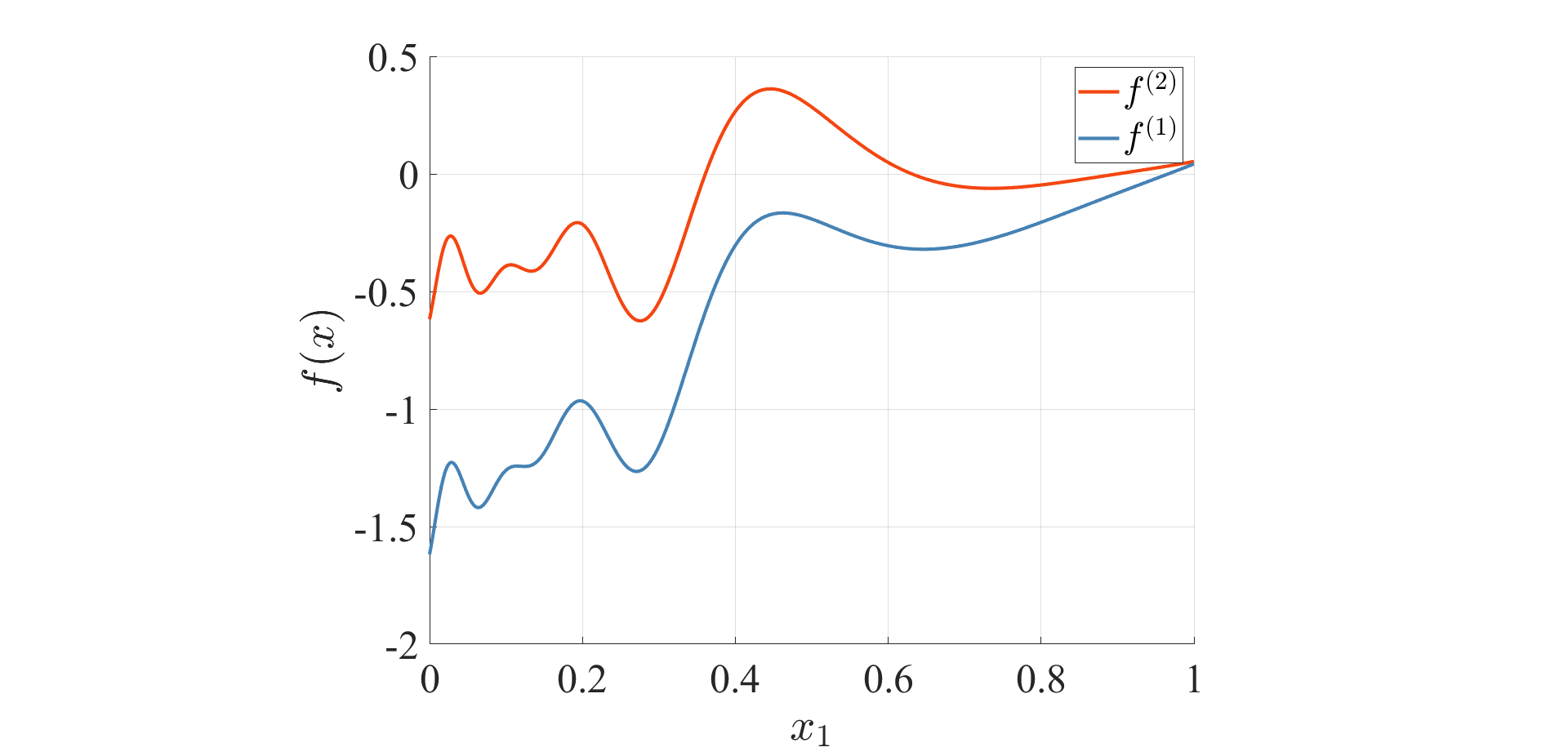} 
        \label{fig:alos1plot}}
        \subfigure[ALOS $\Dim=2$]{%
        \includegraphics[width=0.35\linewidth,trim=0 0 0 0, clip]{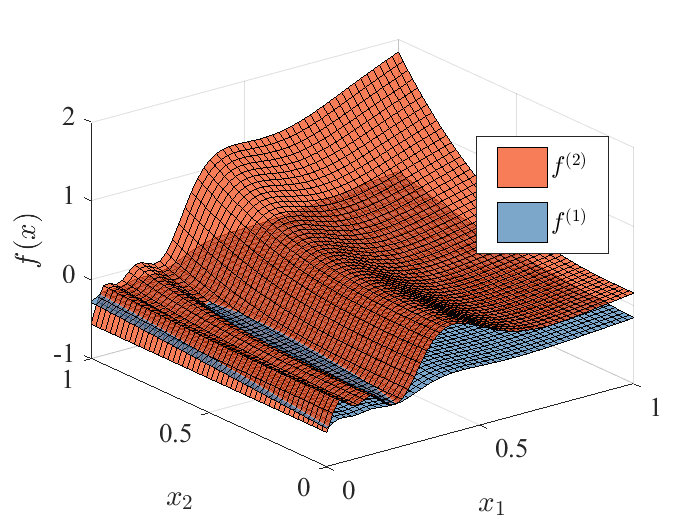}
        \label{fig:alos2plot}}
        
    \caption{ALOS function benchmark problems over the $\Dim=1$ and $\Dim=2$ dimensional domain}\label{fig: AlosBenchPlot}
\end{figure*}

For $\Dim=1$, the analytical optimum is located at $\MinDesignVar= 0.2755$ corresponding to $\ObjFun^{*(2)}=-0.6250$ while for $\Dim \geq 2$ the minimum is located at $\MinDesignVar=[0,0]^{\Dim}$ with value of the objective function $\ObjFun^{*(2)}=-0.5627123$. Figure \ref{fig: AlosBenchPlot} illustrates the high and low-fidelity ALOS function for $\Dim = 1$ (Figure \ref{fig:alos1plot}) and $\Dim = 2$ (Figure \ref{fig:alos2plot}).

\begin{figure*}[t!]
    \centering
     \subfigure[Rastrigin $f^{(3)}$, $f^{(2)}$]{%
        \includegraphics[width=0.35\linewidth,trim=0 0 0 0, clip]{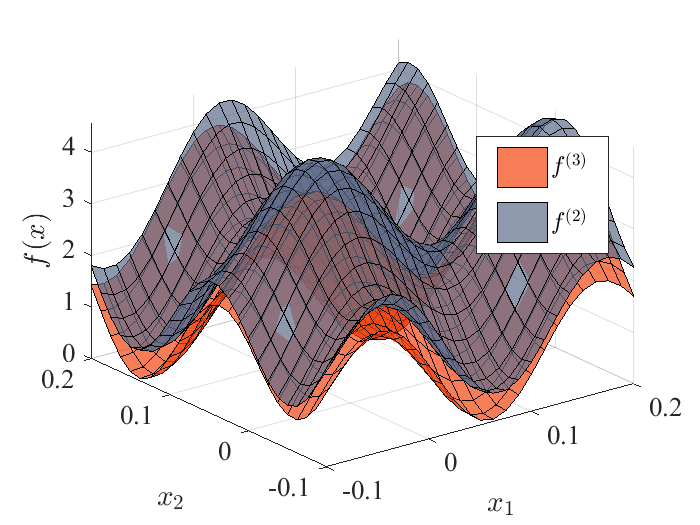} 
        \label{fig:Ras23}}
        \subfigure[Rastrigin $f^{(3)}$, $f^{(1)}$]{%
        \includegraphics[width=0.35\linewidth,trim=0 0 0 0, clip]{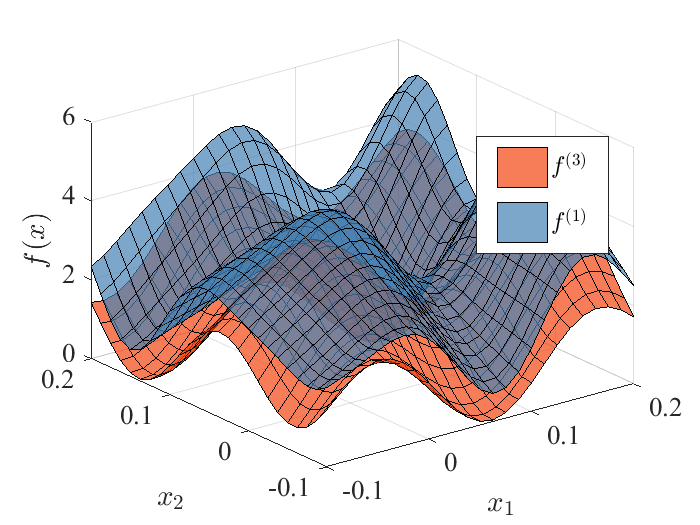}
        \label{fig:Ras13}}
        
    \caption{Rastrigin function shifted and rotated benchmark problem}\label{fig: RastriginPlot}
\end{figure*}

\subsubsection{Shifted-Rotated Rastrigin Function}
\label{s: Rastrigin}

The Rastrigin function is commonly used as test function to represent real-world applications where the objective function might present a high multimodal behaviour. We adopt a benchmark problem based on the original formulation of the Rastrigin function shifted and rotated as follows (Figure \ref{fig: RastriginPlot}): 

\begin{equation} \label{e:rastrigin}
\ObjFun (\pmb{z})=\sum_{i=1}^{\Dim}(z_i^2+1-\cos(10\pi z_i)),
\end{equation}

where: $\pmb{z}= R(\theta)(\DesignVar-\MinDesignVar)$ and $R(\theta)=\begin{bmatrix} \cos\theta & -\sin\theta \\
\sin\theta & \cos\theta\end{bmatrix}$ is the rotation matrix with the rotation angle fixed at $\theta=0.2$. We define three levels of fidelity for this benchmark problem as follows:

\begin{equation} \label{e:fidlevel_rastrigin}
    \ObjFun^{(\LevFid)}(\pmb{z},\phi)=\ObjFun(\pmb{z})+e_r(\pmb{z},\phi_i) 
\end{equation}

\noindent where $e_r(\pmb{z},\phi_i) $ is the resolution error:

\begin{equation} \label{e:res_error}
e_r(\pmb{z},\phi)=\sum_{i=1}^{2}a(\phi)\cos^2(w(\phi)z_i+b(\phi)+\pi).
\end{equation}

\noindent with $\Theta(\phi)=1-0.0001\phi$, $a(\phi)= \Theta(\phi)$, $ w(\phi)=10\pi\Theta(\phi)$, and $b(\phi)=0.5\pi\Theta(\phi)$. Thus, we define the high-fidelity function $\ObjFun^{(3)}(\phi = 10000)$, the intermediate fidelity function $\ObjFun^{(2)}(\phi = 5000)$ and the low-fidelity function $\ObjFun^{(1)}(\phi = 2500)$. For this benchmark, the input variables are defined within the interval $\DesignSpace=[-0.1,0.2]^2$ and the analytical optimum is $\ObjFun^{*(3)}=0$ located at $\MinDesignVar=[0.1,0.1]$.

\begin{figure*}[t!]
    \centering
        \includegraphics[width=0.35\linewidth,trim=0 0 0 0, clip]{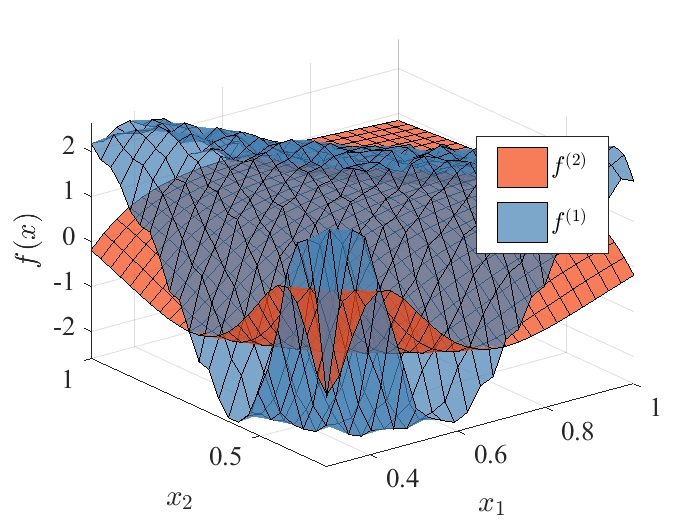} 
        \label{fig:Pac}
    \caption{Paciorek function benchmark problem}\label{fig: PaciorekPlot}
\end{figure*}

\subsubsection{Spring-Mass System}
\label{s: MK4}

This benchmark problem consists of a coupled spring mass system composed of two masses connected by two springs. The challenges associated with this simple physical optimization problem are related to the intrinsic multimodality induced by the elastic behaviour of the system dynamics. We consider the masses $m_1$ and $m_2$ concentrated at their center of gravity and the elastic behaviour of the two spring modeled through the Hooke's law and characterized by the Hooke's constants $k_1$ and $k_2$, respectively. Considering a friction-less dynamics, it is possible to define the equations of motion as follows

\begin{align} \label{e:MK4_1}
  m_1 \ddot{h}_1(t) &= (-k_1 -k_2)  \, h_1(t)+k_2 h_2(t)\\
  m_2 \ddot{h}_2(t) &= k_2 h_1(t)+ (-k_1-k_2) \,h_2(t).
\end{align}

\noindent where $h_1(t)$ and $h_2(t)$ are the positions of the masses as a function of time $t$. 

Equation \eqref{e:MK4_1} can be solved using the fourth-order accurate Runge-Kutta time-marching method and varying the time-step $dt$ to define two fidelity levels. Specifically, we define the high-fidelity model $\ObjFun^{(2)}(dt=0.01)$ and the low-fidelity model $\ObjFun^{(1)}(dt=0.6)$. The benchmark problem requires the identification of the combination of masses and Hooke's constants of springs $\DesignVar=[m_1, m_2, k_1, k_2]$ that minimizes $h_1(t=6)$ considering the domain $\DesignSpace = [1,4]^4$ and the initial conditions of motion $h_1=h_2=0$ and $\dot{h}_1 = \dot{h}_2 = 0$.

\subsubsection{Paciorek Function with Noise}
\label{s: paciorek}

The Paciorek function reproduces an optimization setting where the objective function is affected by measurement noise and localized multimodal behaviour. This scenario is replicated through a random noise term in the low-fidelity Paciorek function as follows (Figure \ref{fig: PaciorekPlot}): 


\begin{equation}
    \ObjFun^{(2)}(\DesignVar) =sin\left(\prod_{i=1}^\Dim\DesignVar_i\right)^{-1}
\end{equation}

\begin{equation}
\begin{split}
     \ObjFun^{(1)}(\DesignVar) &= \ObjFun^{(2)}(\DesignVar)-9A^2\cos\left(\prod_{i=1}^\Dim\DesignVar_i\right)^{-1} + rand.norm(0,\alpha)
\end{split}
\end{equation}

where $\ObjFun^{(2)}$ is the Paciorek function, $A = 0.5$, $\alpha=0.2$, and the input variable is defined across the input domain $\DesignSpace = [0,3,1]^2$.

\subsection{Results and Discussion}
\label{s: results}

First, we define the following evaluation metrics to assess the performances of the Bayesian schemes \cite{mainini2022analytical}: 

\begin{equation}
  \epsilon_\DesignVar = \frac{\| \DesignVar^*- \hat{\DesignVar}^*\|}{\sqrt{N}}
\end{equation}

\begin{equation}
    \epsilon_f = \frac{f(\hat{\DesignVar}^*)-\MinObjFun}{f_{max}-\MinObjFun}
\end{equation}


\noindent where ${\DesignVar}^*$ is the location of the analytical optimum, $\hat{\DesignVar}^*$ is the optimum identified by the algorithm, and $\ObjFun_{max}$ and $\MinObjFun$ are the maximum and minimum of the objective function, respectively. The first metric $\epsilon_\DesignVar$ quantifies the search error in the domain of the objective function, while the second metric $\epsilon_\ObjFun$ evaluates the error associated with the learning goal -- minimum of the objective function \cite{serani2016parameter}. We evaluate the metrics $\epsilon_\DesignVar$ and $\epsilon_\ObjFun$ as functions of the computational budget $\Budget$ defined as the cumulative computational cost associated with observations of the objective function considering the $\LevFid$-th level of fidelity. We run 10 trails for each benchmark problem presented in Section \ref{s: Benchmarks} to compensate the influence of the random initial design of experiments, and to verify the sensitivity and robustness of the algorithms to the initialization setting. The results for all the experiments are reported in terms of median values of $\epsilon_\DesignVar$ and $\epsilon_\ObjFun$.

\begin{figure*}[t!]
    \centering
     \subfigure[Forrester]{%
        \includegraphics[width=0.3\linewidth,trim=220 0 245 0, clip]{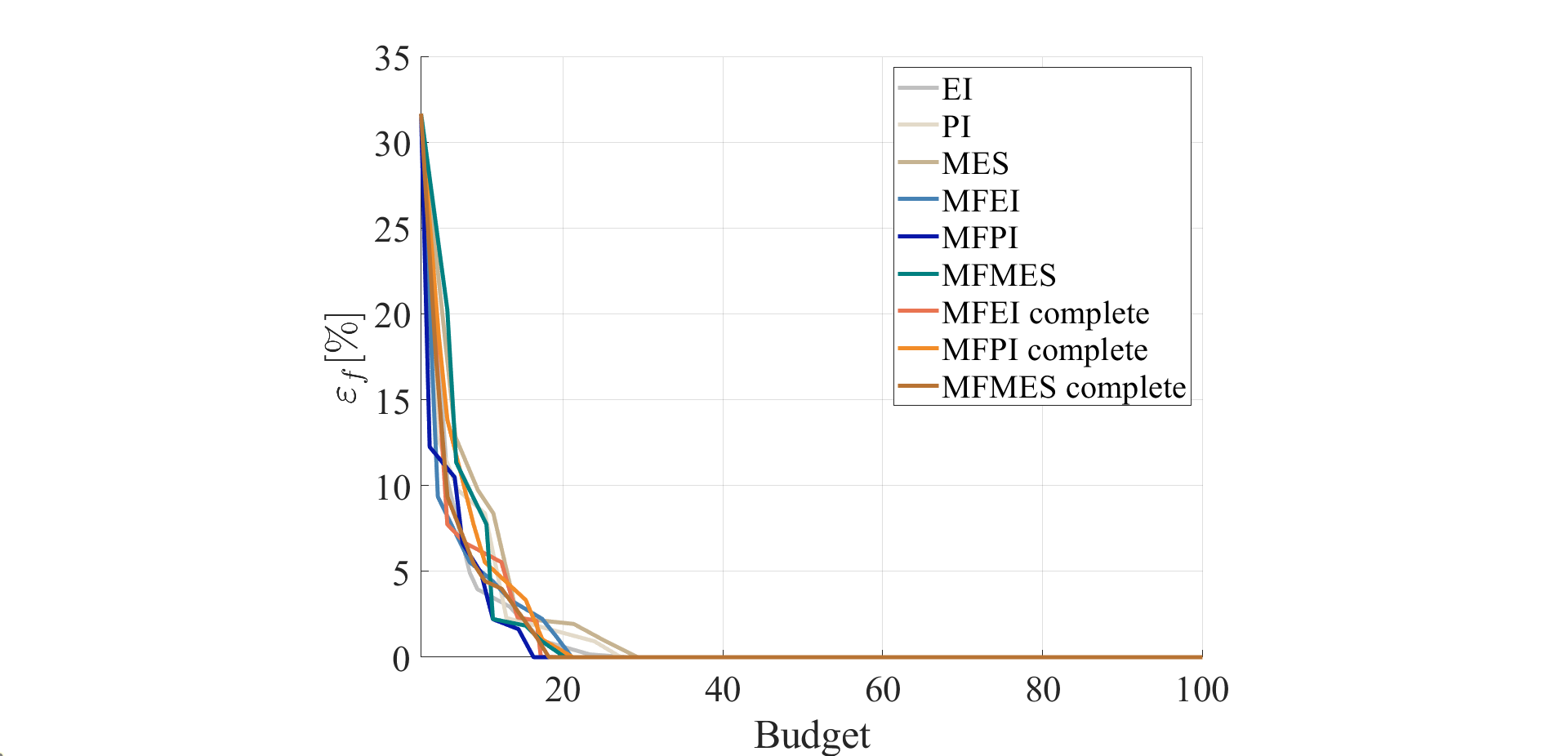} 
        \label{fig:Forresterf}}
        \subfigure[Jump Forrester]{%
        \includegraphics[width=0.3\linewidth,trim=220 0 245 0, clip]{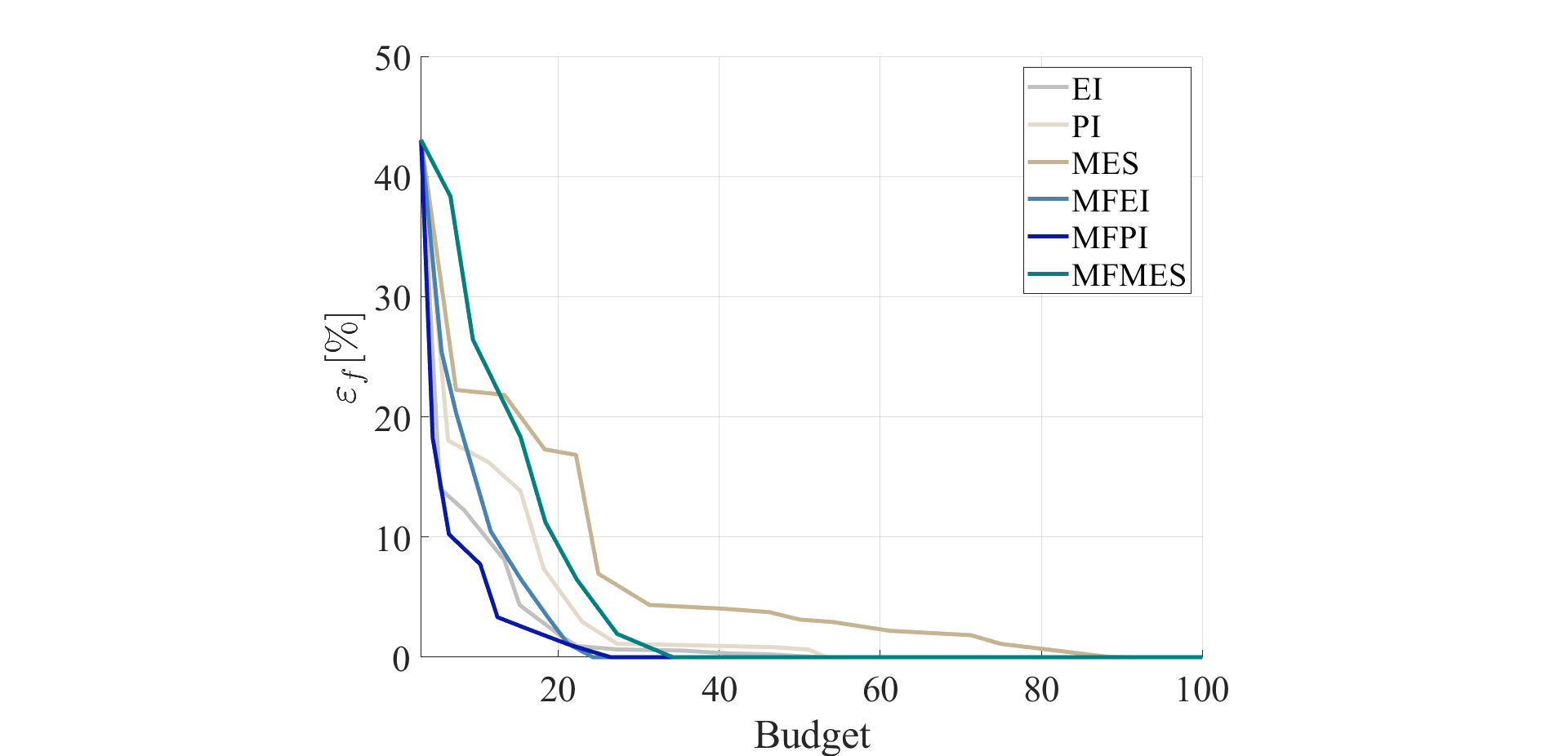}
        \label{fig:JumpForresterf}}

        \subfigure[Forrester]{%
        \includegraphics[width=0.3\linewidth,trim=220 0 245 0, clip]{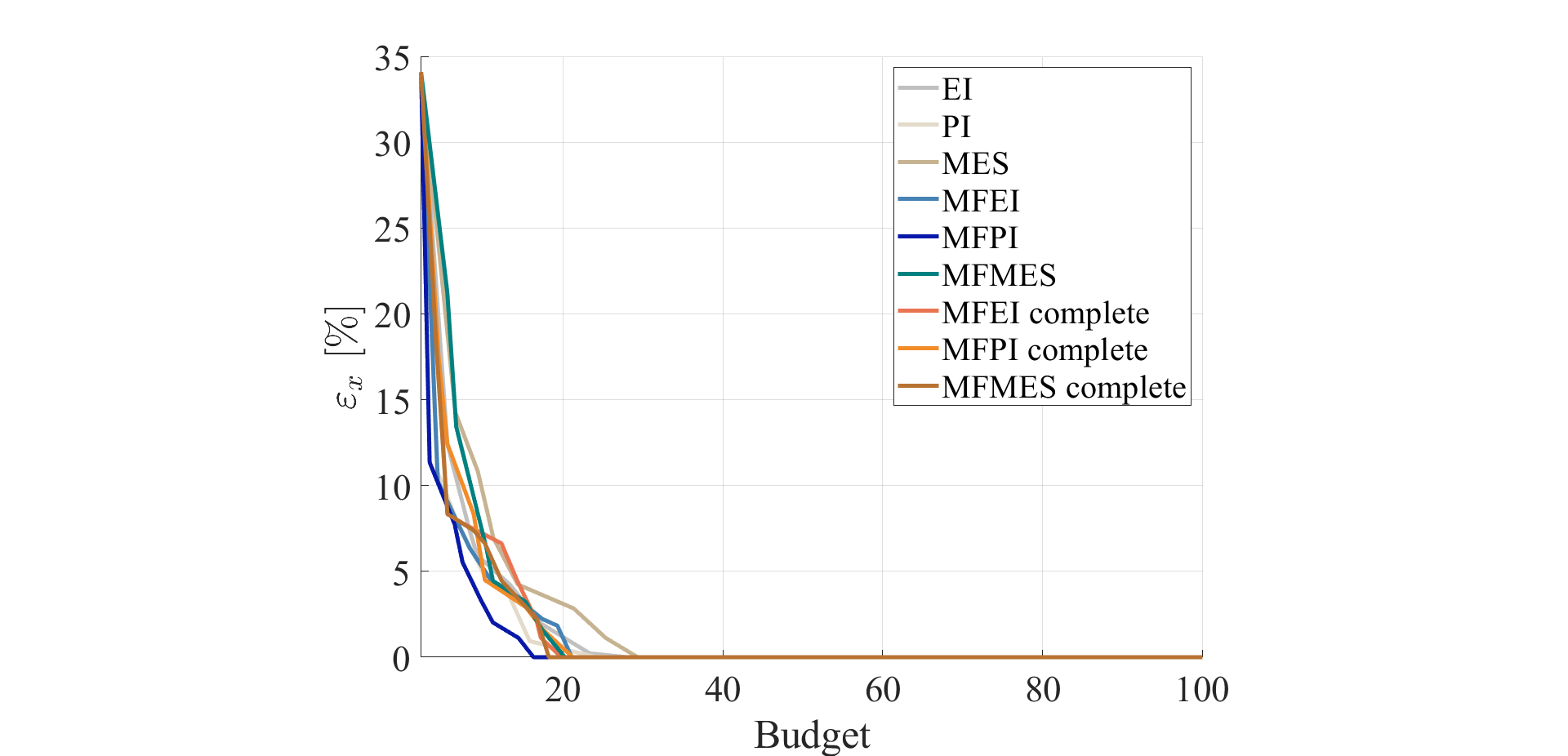} 
        \label{fig:Forresterx}}
        \subfigure[Jump Forrester]{%
        \includegraphics[width=0.3\linewidth,trim=220 0 245 0, clip]{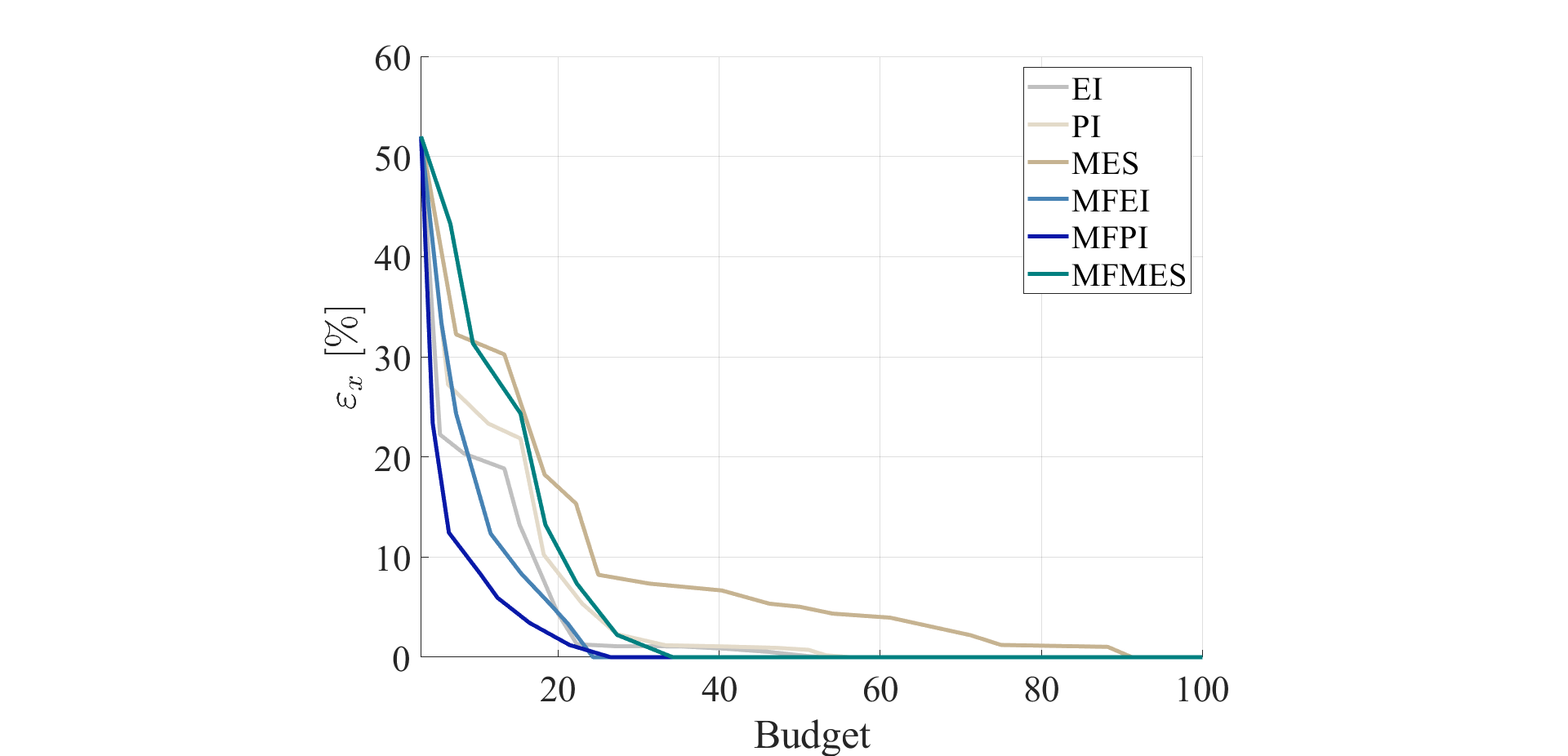}
        \label{fig:JumpForresterx}}
        
    \caption{Performances of the competing algorithms for the Forrester and Jump Forrester benchmarks.}\label{fig: ForresterResults}
\end{figure*}

Figure \ref{fig: ForresterResults} summarizes the outcomes obtained for the Forrester function and discontinuous Forrester function. The results for the Forrester benchmark (Figure \ref{fig:Forresterf} and Figure \ref{fig:Forresterx}) show that the multifidelity algorithms identify the optimum solution with a significant reduction of the computational budget if compared with the single fidelity counterparts. The best performing algorithm is the MFPI learner considering only the high-fidelity $\LevFid = 4$ and the lower-fidelity $\LevFid = 1$ levels, while the second best is the MFEI acquisition function considering available the complete spectrum of fidelities $\LevFid = 1,2,3,4$. These outcomes suggest that multifidelity learning paradigms driven majorly by informativeness -- MFPI acquisition function -- are capable of efficiently directing the computational resources toward the optimum of low-dimensional objective functions in presence of continuous localized behaviours. Moreover, it should be noted that the MFEI capitalizes from all the information sources available and leverages the balance between informativeness and representativeness/diversity to effectively search toward the analytical optimum. The single fidelity Bayesian frameworks exhibit a lower convergence rate with respect to the multifidelity algorithms. The EI and PI use almost the same computational budget to identify the optimum solution, while the MES adopts more evaluations of the objective function. This confirms the observations for the multifidelity experiments. PI takes advantage from the purely exploitation of high-fidelity samples in the surrounding of the surrogate minimum to reach the optimum. This can be explained with the computation of an accurate surrogate model -- at least close to the optimum -- for low-dimensional objective functions. In contrast, EI balances an exploration phase to improve the overall accuracy of the surrogate with the exploitation toward the believed optimum. Particular attention should be dedicated to the MES and MFMES outcomes. In the single-fidelity frameworks, MES scores slightly worst both in terms of convergence rate and budget expenditure. This can be interpreted with an overall over-exploration behavior: MES distributes computational resources to explore the domain and refine the surrogate model, and directs lately efforts toward the optimum. This trend is considerably dampened in the multifidelity scenario, where MFMES shows good capabilities especially when all the sources of information are available during the search. In this case, cheap low-fidelity observations are used to explore the domain with contained computational expenditure, and high-fidelity data are mostly adopted to search toward the prescribed optimum location. 

The discontinuous Forrester problem introduces a discontinuous local property of the objective function that further stresses the learning schemes. This can be explicitly observed with the average improvement of the budget required to achieve the optimum. Overall, it is possible to identify the same trends observed for the continuous Forrester function (Figure \ref{fig:JumpForresterf} and Figure \ref{fig:JumpForresterx}): either balancing exploration and exploitation -- EI and MFEI -- or a major exploitation search -- PI and MFPI -- lead to an efficient identification of the analytical optimum. In contrast, the over-exploration of MES and MFMES decelerates the optimization procedure with respect to the counterpart competing methods. This can be observed majorly for the MES which uses almost all the budget available to explore the domain and finally reach the optimum.

\begin{figure*}[t!]
    \centering
     \subfigure[Rosenbrock D=2]{%
        \includegraphics[width=0.3\linewidth,trim=220 0 245 0, clip]{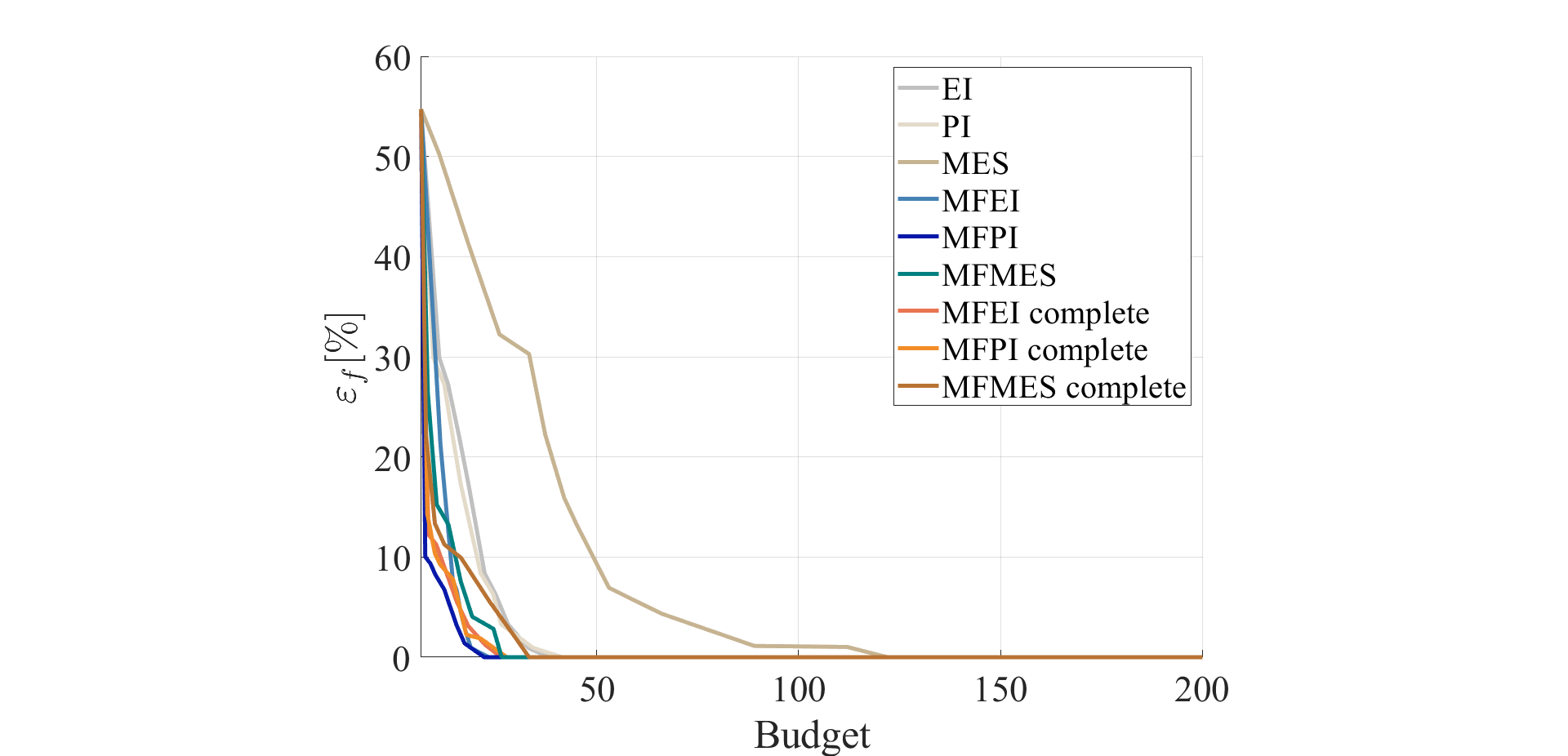} 
        \label{fig:ros2f}}
        \subfigure[Rosenbrock D=5]{%
        \includegraphics[width=0.3\linewidth,trim=220 0 245 0, clip]{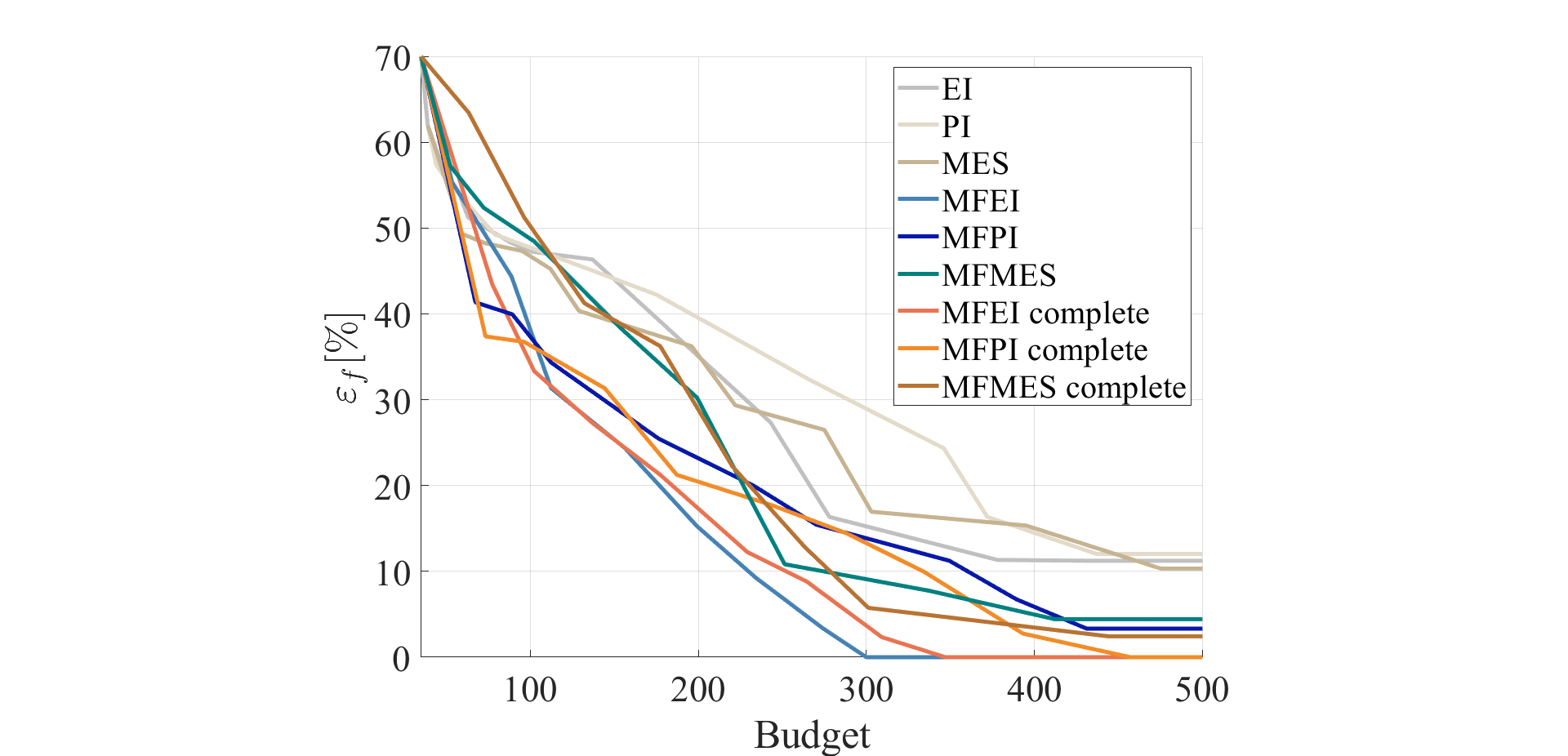}
        \label{fig:ros5f}}
        \subfigure[Rosenbrock D=10]{%
        \includegraphics[width=0.304\linewidth,trim=220 0 235 0, clip]{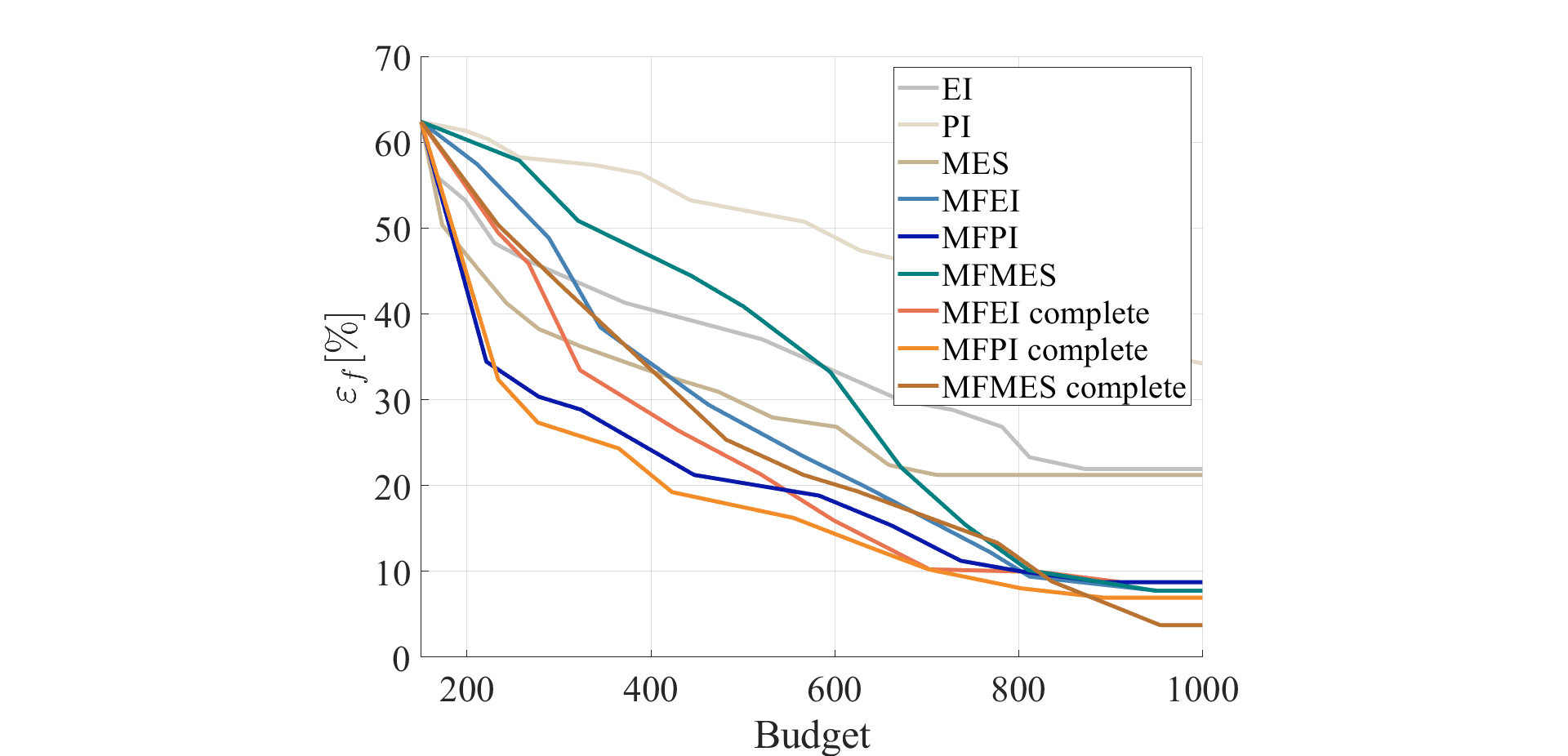}
        \label{fig:ros10f}}

         \subfigure[Rosenbrock D=2]{%
        \includegraphics[width=0.3\linewidth,trim=220 0 245 0, clip]{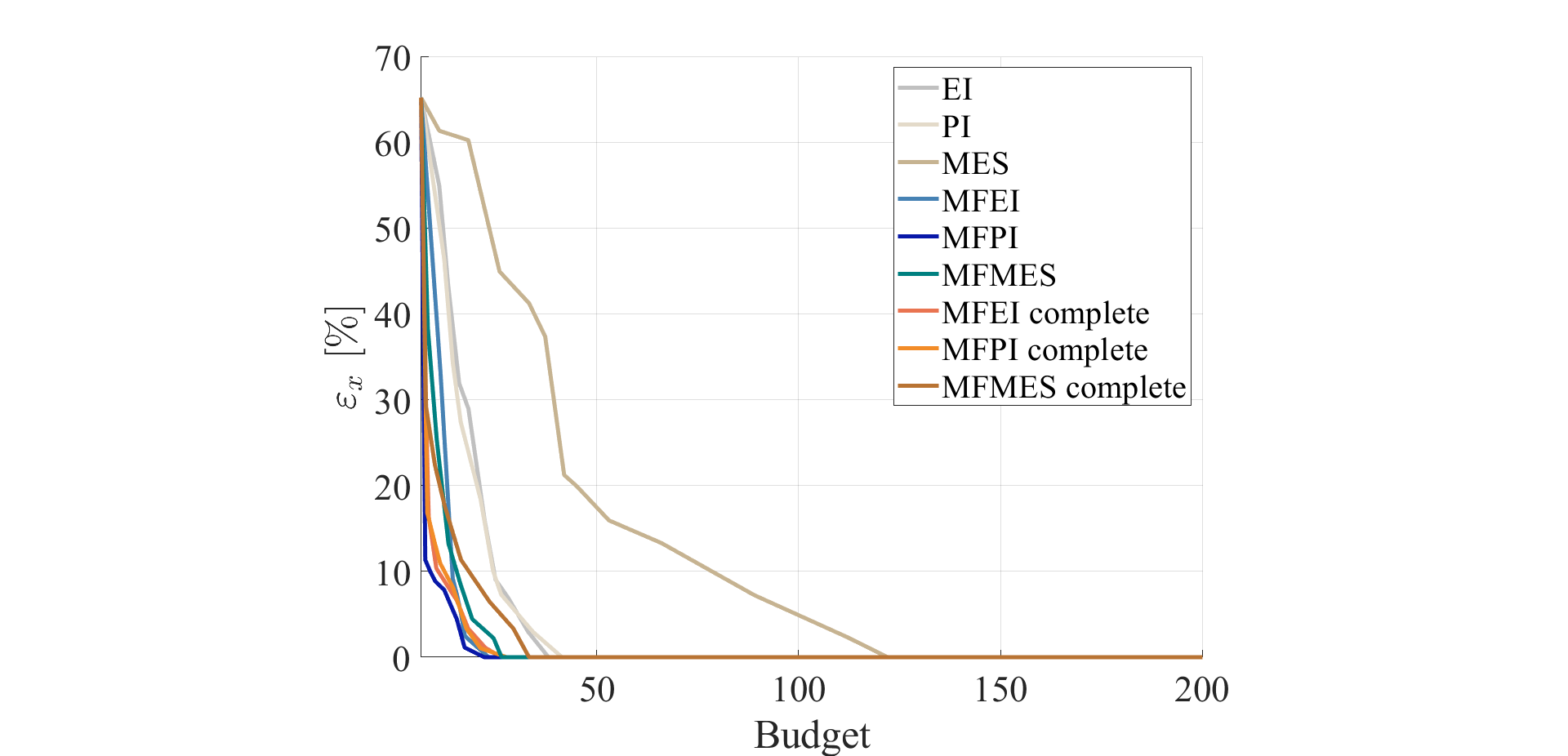} 
        \label{fig:ros2x}}
        \subfigure[Rosenbrock D=5]{%
        \includegraphics[width=0.3\linewidth,trim=220 0 245 0, clip]{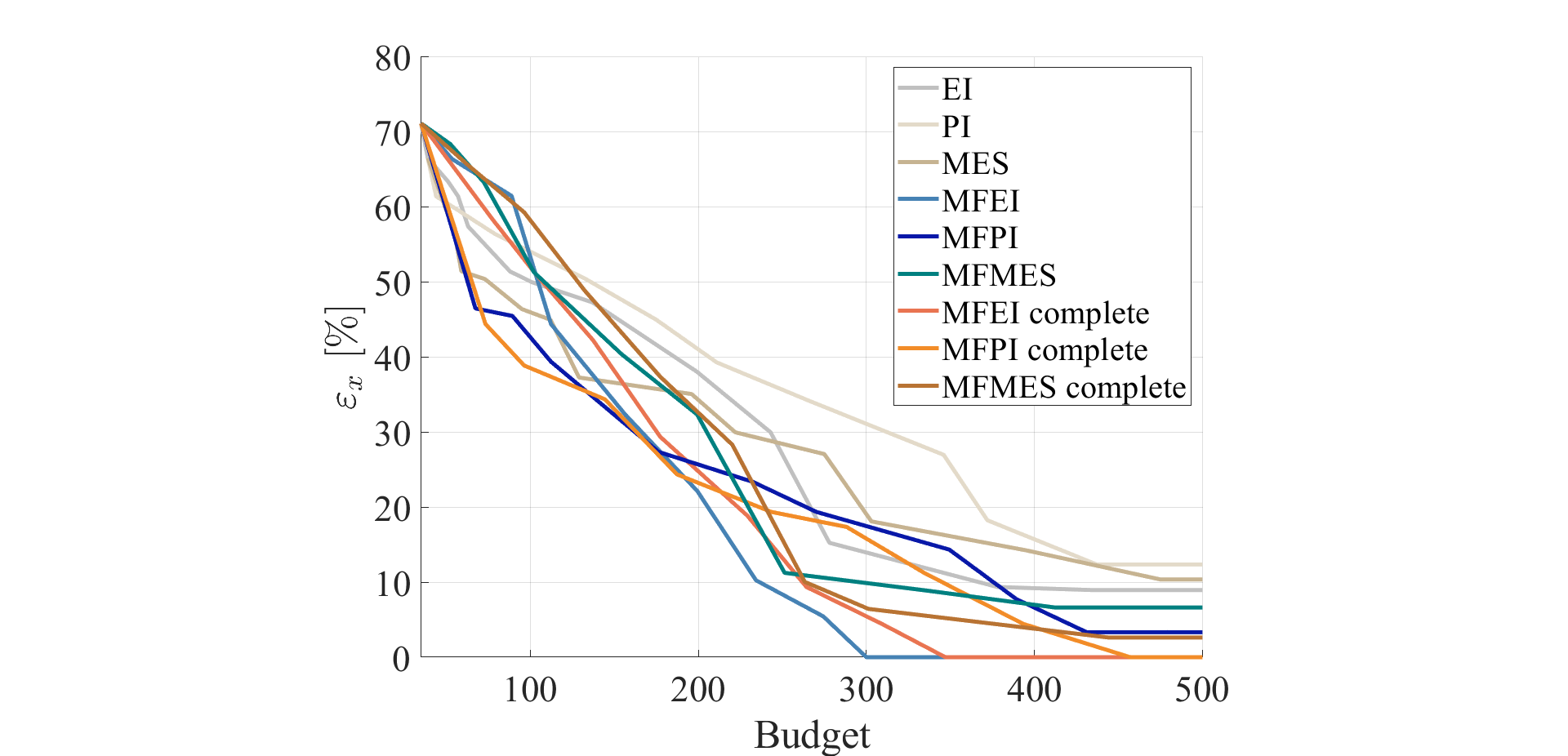}
        \label{fig:ros5x}}
        \subfigure[Rosenbrock D=10]{%
        \includegraphics[width=0.304\linewidth,trim=220 0 235 0, clip]{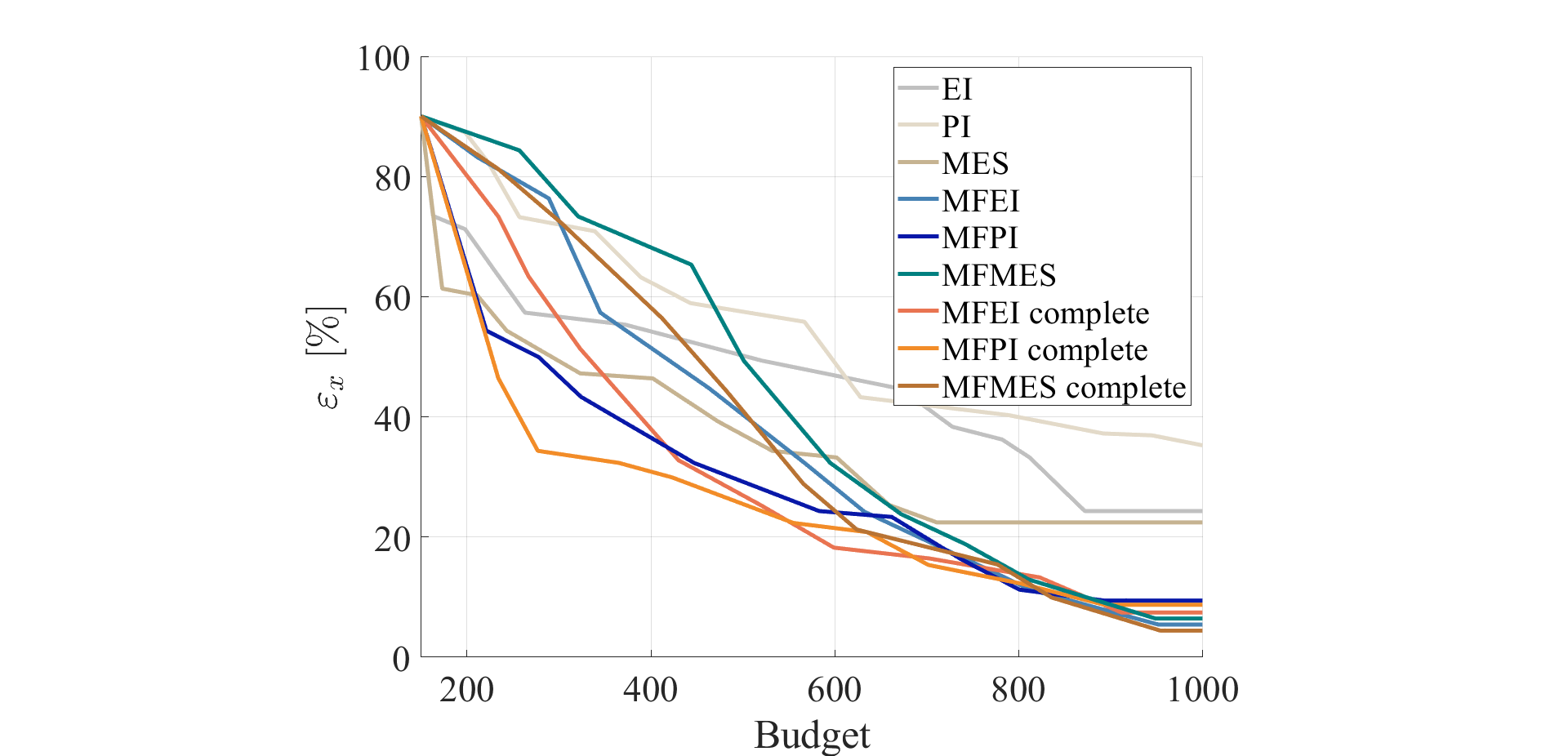}
        \label{fig:ros10x}}
    \caption{Performances of the competing algorithms for the Rosenbrock benchmarks.}\label{fig:RosResults}
\end{figure*}

Figure \ref{fig:RosResults} illustrates the experiments conducted on the Rosenbrock benchmark function increasing the dimensionality $\Dim$ of the domain. This allows to investigate the performance of the learning scheme as the number of parameters to optimize increases. Overall, the multifidelity schemes deliver better convergences with a fraction of the computational budget required by single-fidelity algorithms for all the dimensions of the domain -- $\Dim=2,5,10$. 

For $\Dim=2$ (Figure \ref{fig:ros2f} and Figure \ref{fig:ros2x}), MFEI and MFPI implementing only the highest and lower levels of fidelity $\LevFid = 1,3$ are the best performing algorithms, followed by the counterpart considering all the fidelities spectrum and the MFMES also learning from $\LevFid = 1,3$. Two major observations can be made in this experimental setting. First, multifidelity learners are not capable of making advantage of the intermediate fidelity $\LevFid=2$ during exploration leading to an increase of the computational expenditure. A possible explanation to these outcomes is the local behaviour of the intermediate fidelity that pushes the exploration in regions far from the optimum. Second, pure exploitation or a balanced search between exploration and exploitation are advantageous in low-dimensional domains, while pure exploration sacrifices valuable computational resources to improve the awareness about the global distribution of the objective instead of searching the optimum.

Increasing the dimension of the input space to $\Dim=5$ (Figure \ref{fig:ros5f} and Figure \ref{fig:ros5x}), only the MFEI and the MFPI using all the fidelities available are capable of identifying the optimum solution, while the other competing algorithms converge to suboptimal solutions. However, it should be noted the much faster convergence of MFEI with respect to the MFPI in the complete setting. These outcomes indicate that as the number of optimization variables increases, both exploration and exploitation are required for an efficient learning procedure. In particular, the exploration improves the accuracy of the surrogate over the domain which permits to better inform the learner during the exploitation phase. The utility of purely exploitation -- MFPI -- also continues to be observed, but the effectiveness is limited by the dimensionality of the domain that requires an exploration phase to better capture the distribution of the objective function.

Pushing further the dimensionality of the domain at $\Dim=10$ (Figure \ref{fig:ros10f} and Figure \ref{fig:ros10x}), all the algorithms are not capable of reaching the analytical optimum with the allocated budget. This can be explained with the unreliable prediction of the surrogate model that is not capable of correctly informing the learner with limited amount of data -- limited allocated budget. However, the multifidelity paradigms achieve larger reductions of both the error in the domain $\epsilon_\DesignVar$ and the goal error $\epsilon_\ObjFun$ if compared with the single-fidelity outcomes. This suggests that learners capable of leveraging multiple information sources might produce higher gains in a limited budget scenario thanks to the massive use of cheap low-fidelity models to learn the objective function. Among the competing strategies, MFMES exhibits remarkable outcomes in terms of convergence values of the errors when all the library of fidelities is available. These results can be justified with the over-exploration properties of the MFMES acquisition function: the learner uses massive low-fidelity data to refine the approximation of the surrogate model and augment its predictive capabilities. This permits to better inform the procedure and direct computational resources toward the optimum.

\begin{figure*}[t!]
    \centering
     \subfigure[ALOS D=1]{%
        \includegraphics[width=0.3\linewidth,trim=220 0 245 0, clip]{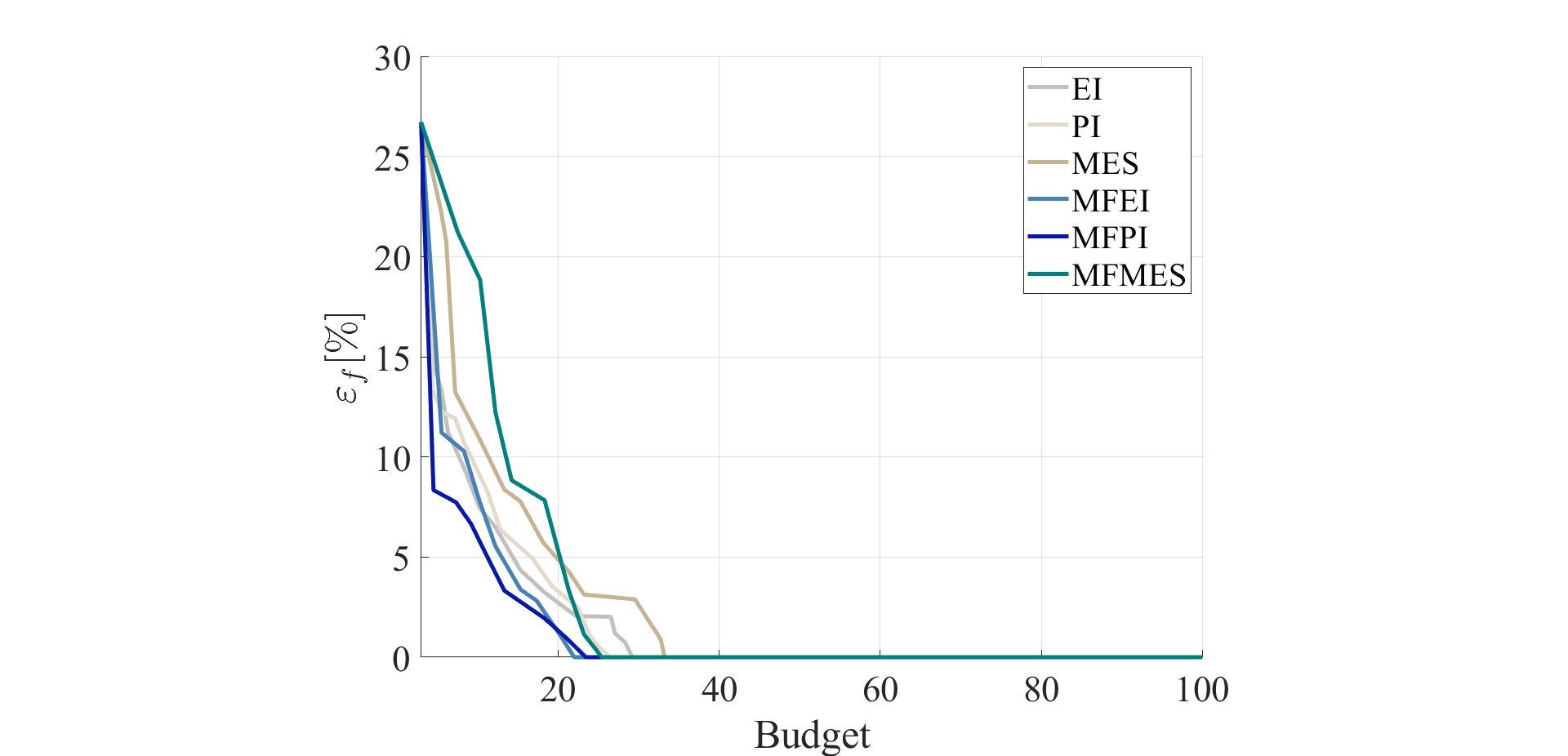} 
        \label{fig:alos1f}}
        \subfigure[ALOS D=2]{%
        \includegraphics[width=0.3\linewidth,trim=220 0 245 0, clip]{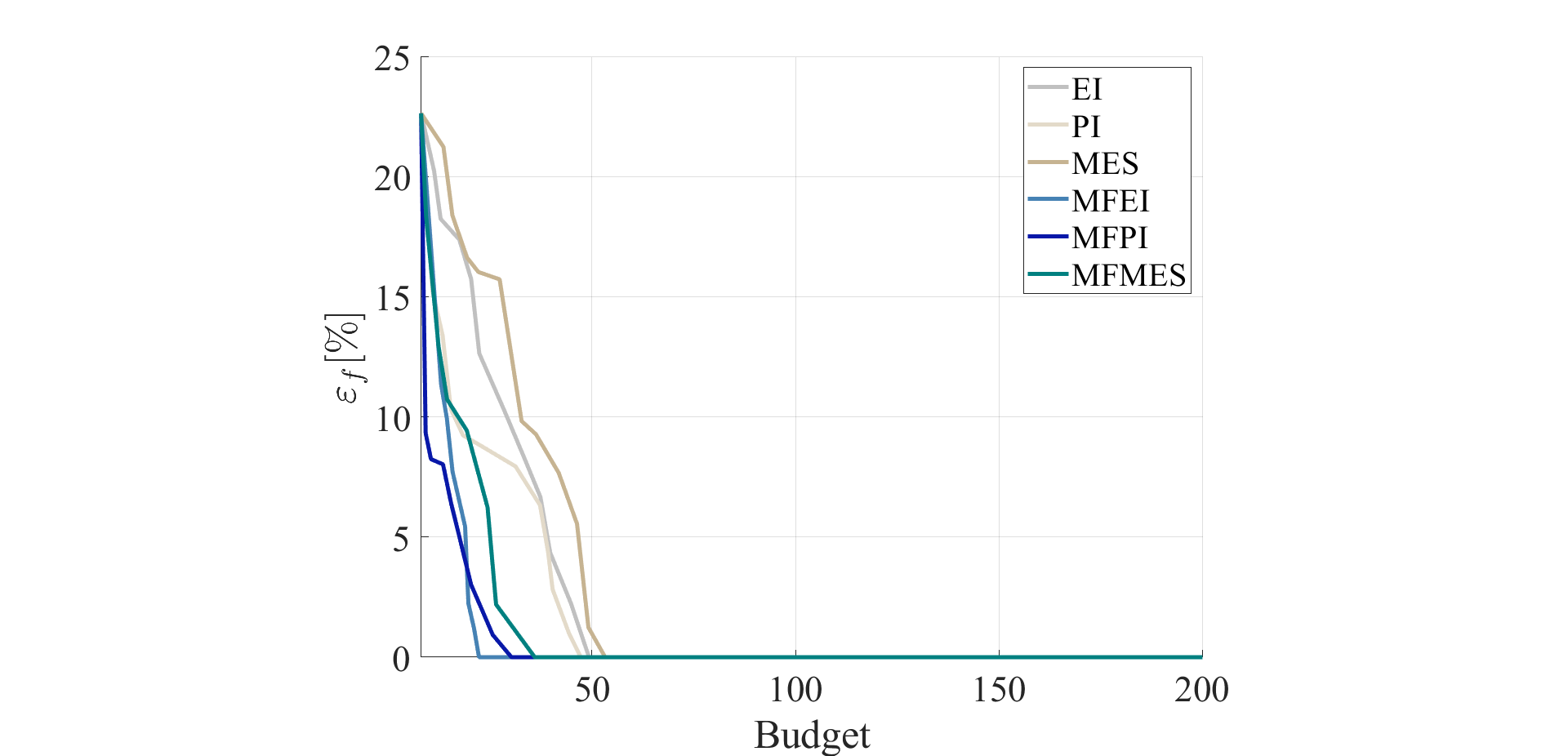}
        \label{fig:alos2f}}
        \subfigure[ALOS D=3]{%
        \includegraphics[width=0.3\linewidth,trim=220 0 245 0, clip]{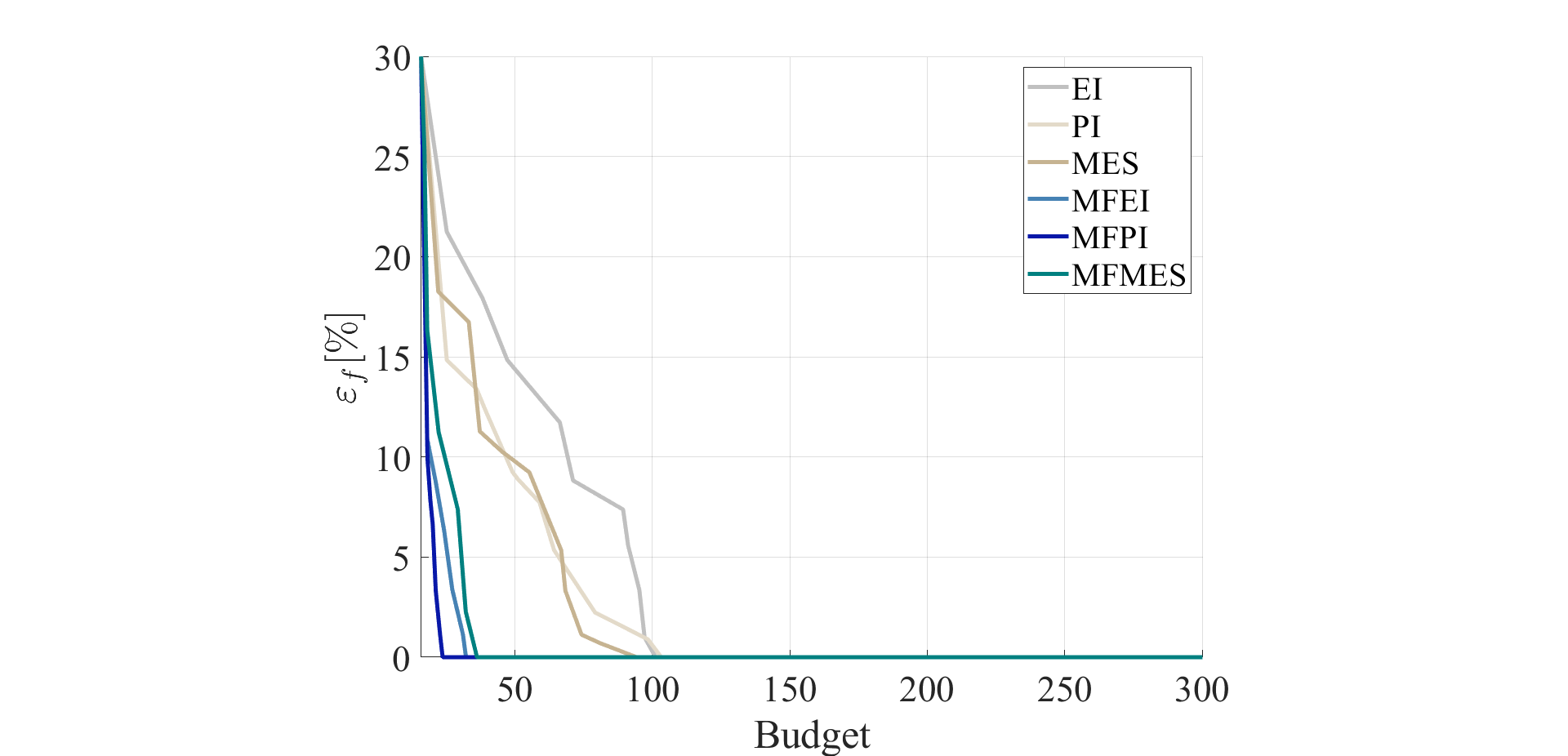}
        \label{fig:alos3f}}

        \subfigure[ALOS D=1]{%
        \includegraphics[width=0.3\linewidth,trim=220 0 245 0, clip]{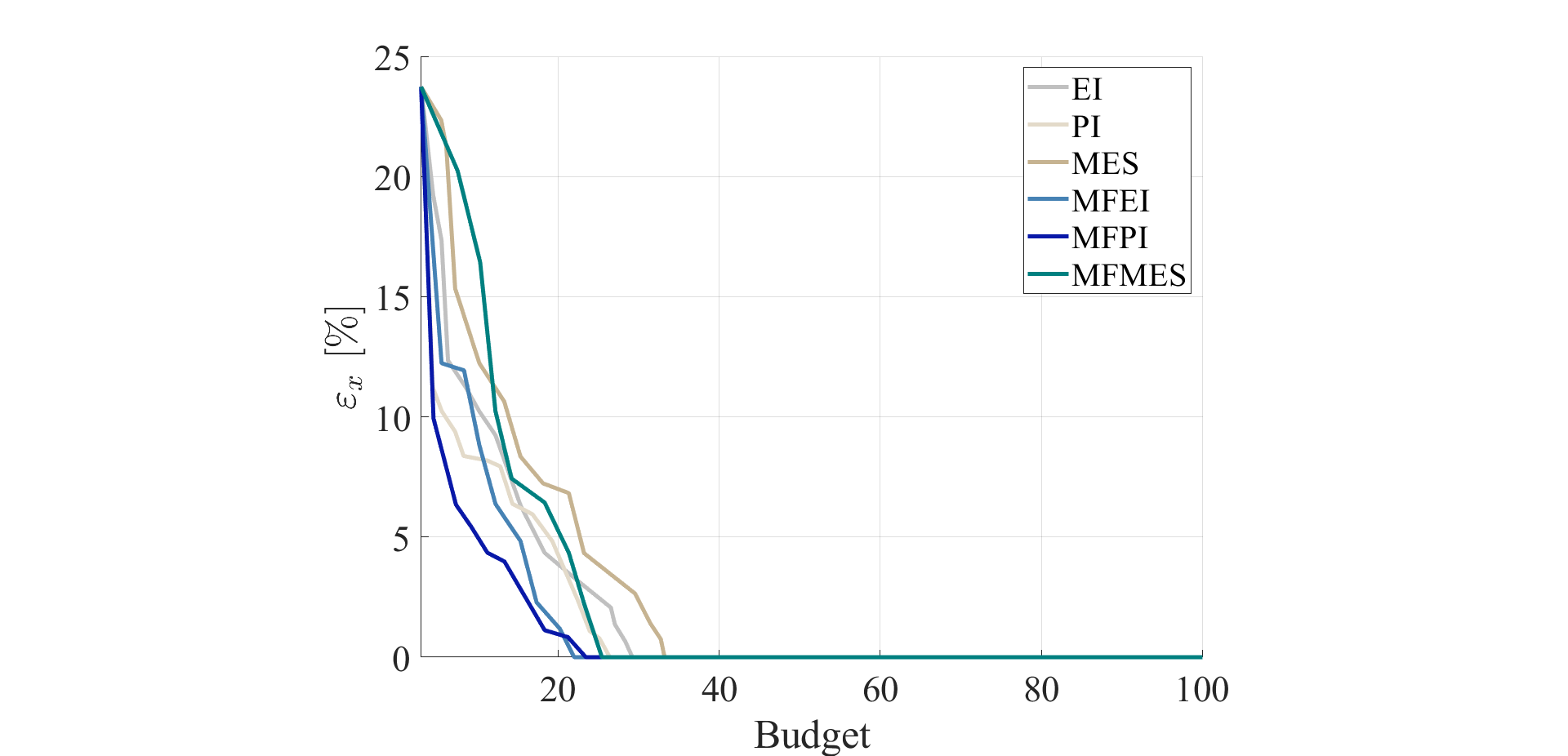} 
        \label{fig:alos1x}}
        \subfigure[ALOS D=2]{%
        \includegraphics[width=0.3\linewidth,trim=220 0 245 0, clip]{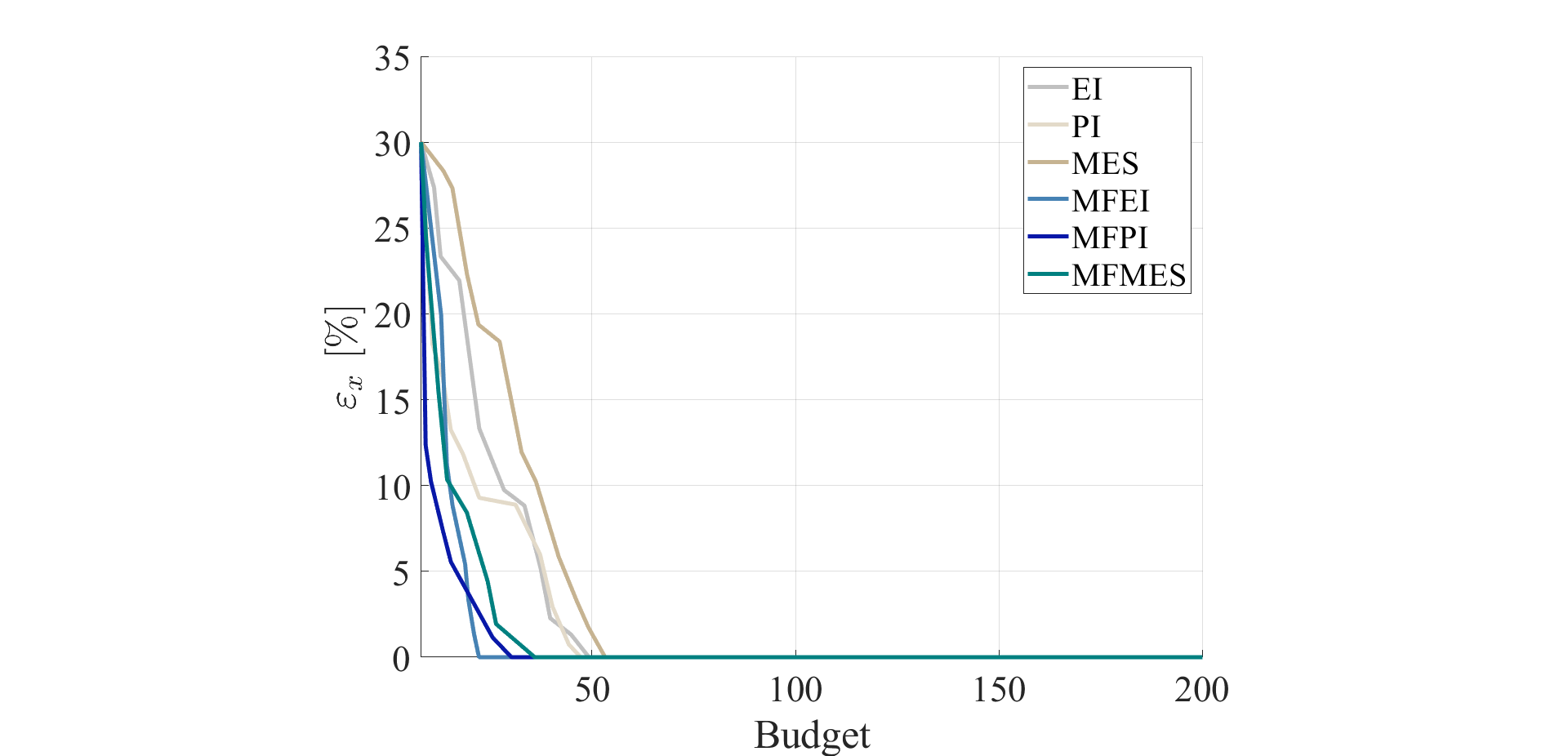}
        \label{fig:alos2x}}
        \subfigure[ALOS D=3]{%
        \includegraphics[width=0.3\linewidth,trim=220 0 245 0, clip]{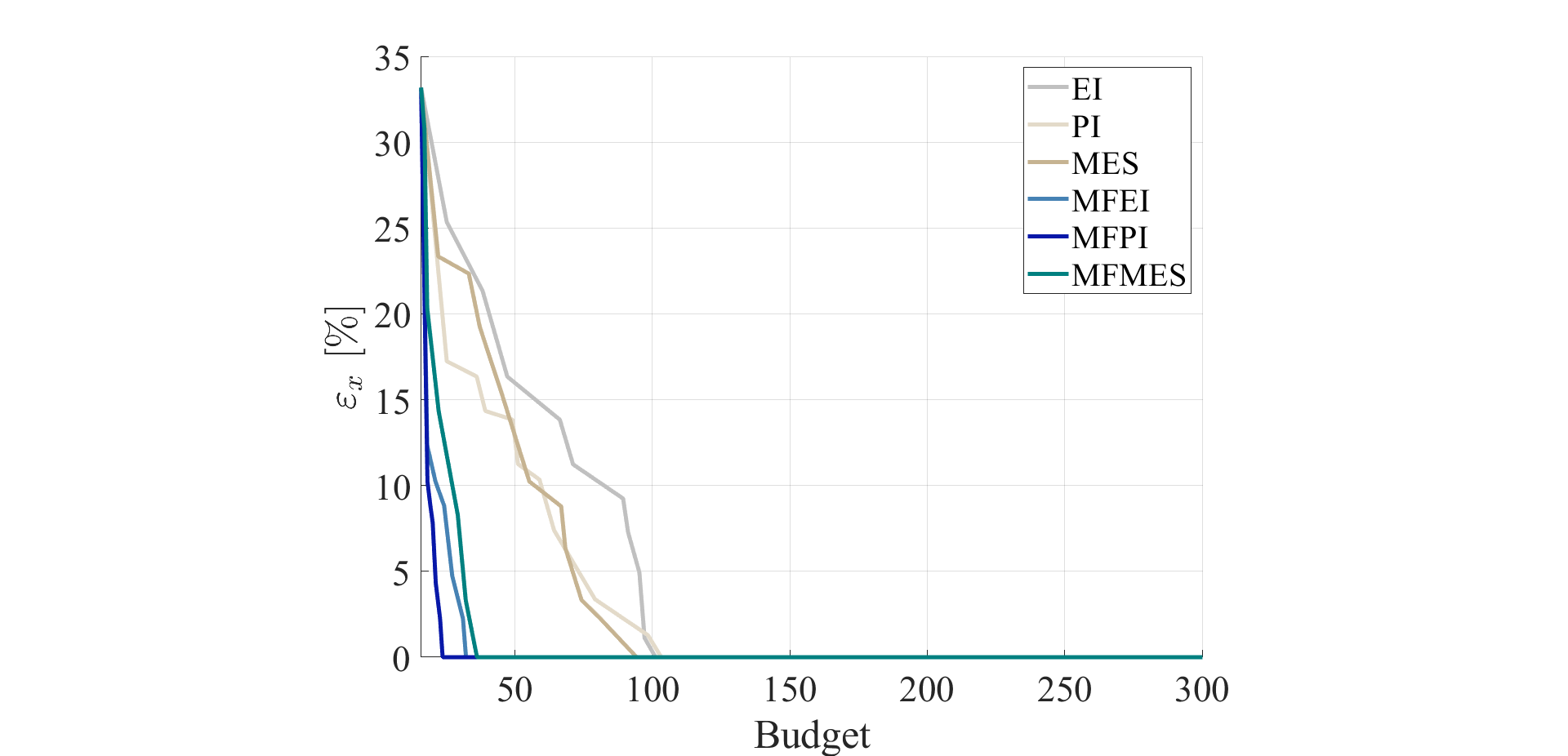}
        \label{fig:alos3x}}
        
    \caption{Performances of the competing algorithms for the ALOS benchmarks.}\label{fig:alosresults}
\end{figure*}

The results obtained for the ALOS benchmark problem in Figure \ref{fig:alosresults} confirm the previous observations about the different effectiveness of the learning schemes. In particular, the multifidelity strategies provide larger accelerations of the optimization procedure in presence of oscillations at different frequencies of the objective function for the one- (Figure \ref{fig:alos1f} and Figure \ref{fig:alos1x}), two- (Figure \ref{fig:alos2f} and Figure \ref{fig:alos2x}) and three- (Figure \ref{fig:alos3f} and Figure \ref{fig:alos3x}) dimensional ALOS problem. We observe that the best performances are delivered by either learners based on the balance between informativeness and representativeness/diversity -- MFEI and EI -- or a purely informativeness-driven -- MFPI and PI --, while over-exploration performs relatively poorly -- MFMES and MES. These results are justified with the low-dimensionality of the objective function.

\begin{figure*}[t!]
    \centering
     \subfigure[Rastrigin]{%
        \includegraphics[width=0.3\linewidth,trim=220 0 245 0, clip]{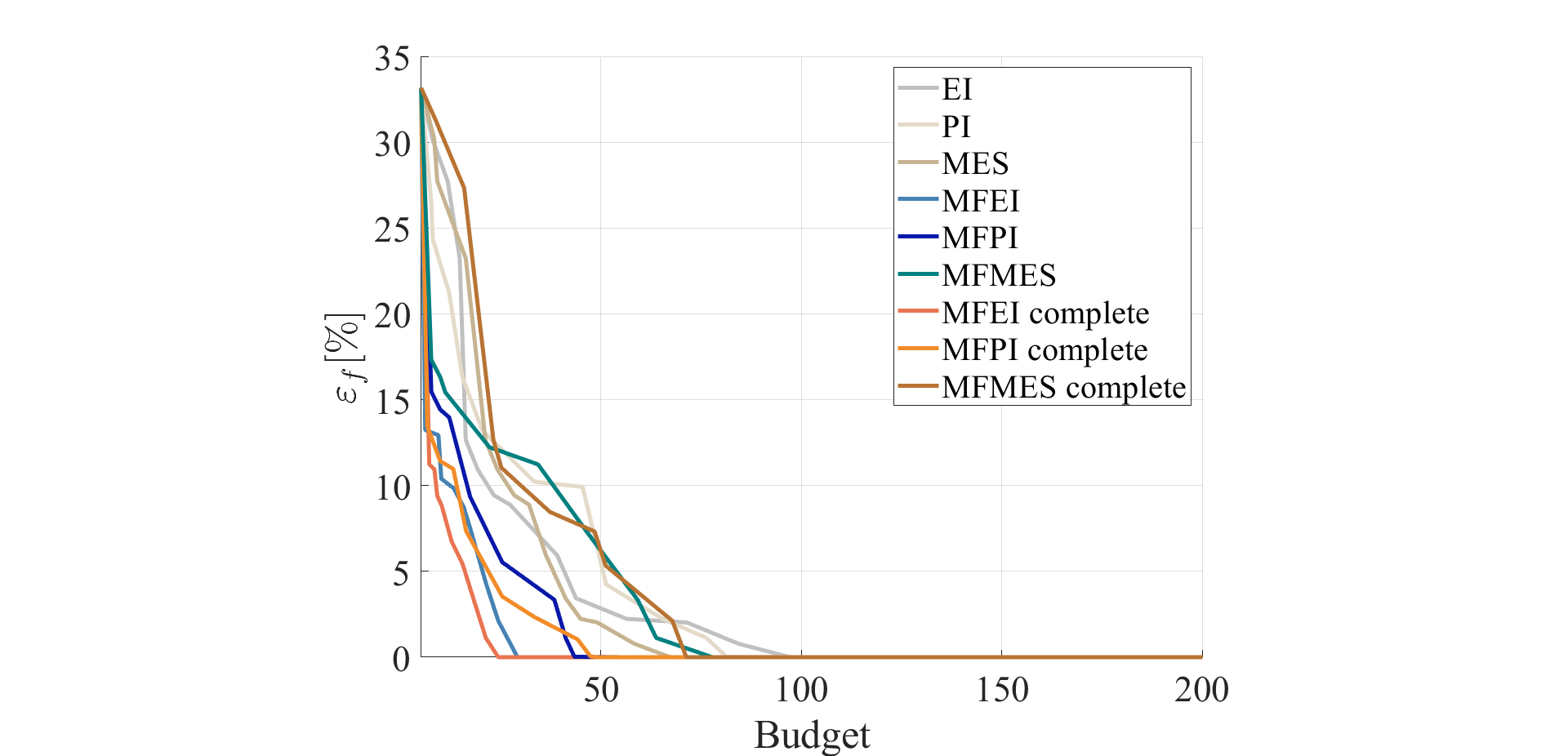} 
        \label{fig:rasf}}
        \subfigure[Spring-Mass System]{%
        \includegraphics[width=0.3\linewidth,trim=220 0 245 0, clip]{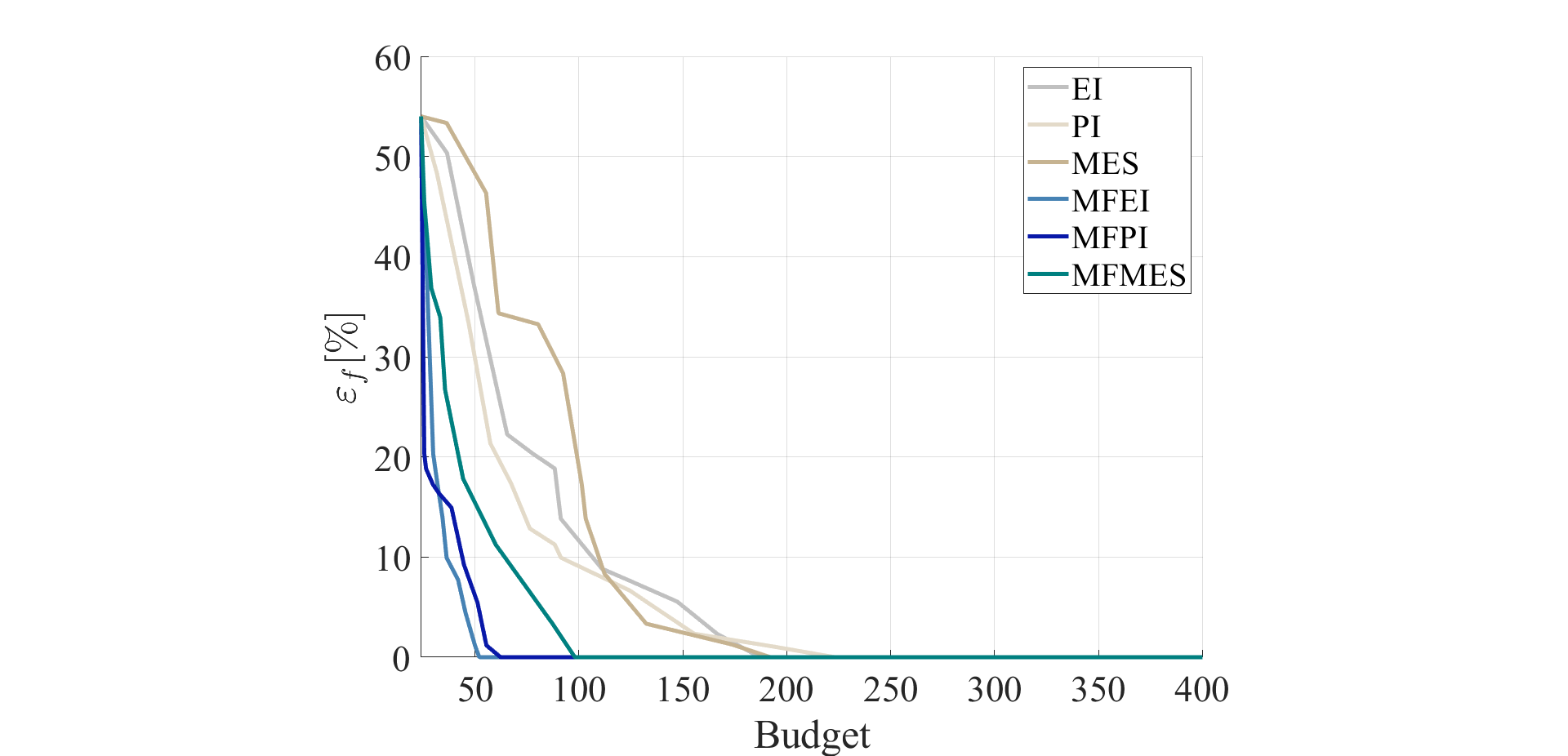}
        \label{fig:mk4f}}
        \subfigure[Paciorek]{%
        \includegraphics[width=0.3\linewidth,trim=220 0 245 0, clip]{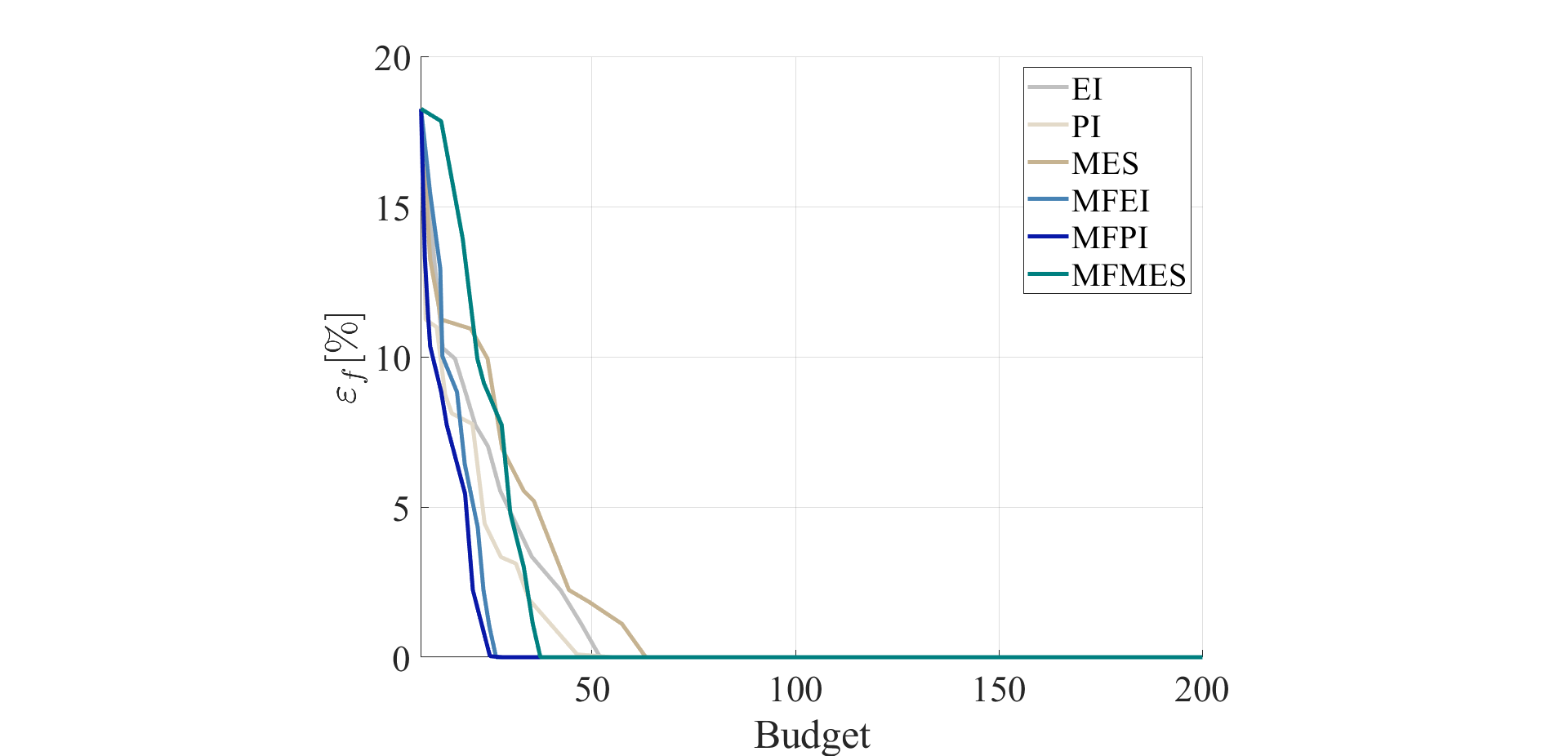}
        \label{fig:pacf}}

        \subfigure[Rastrigin]{%
        \includegraphics[width=0.3\linewidth,trim=220 0 245 0, clip]{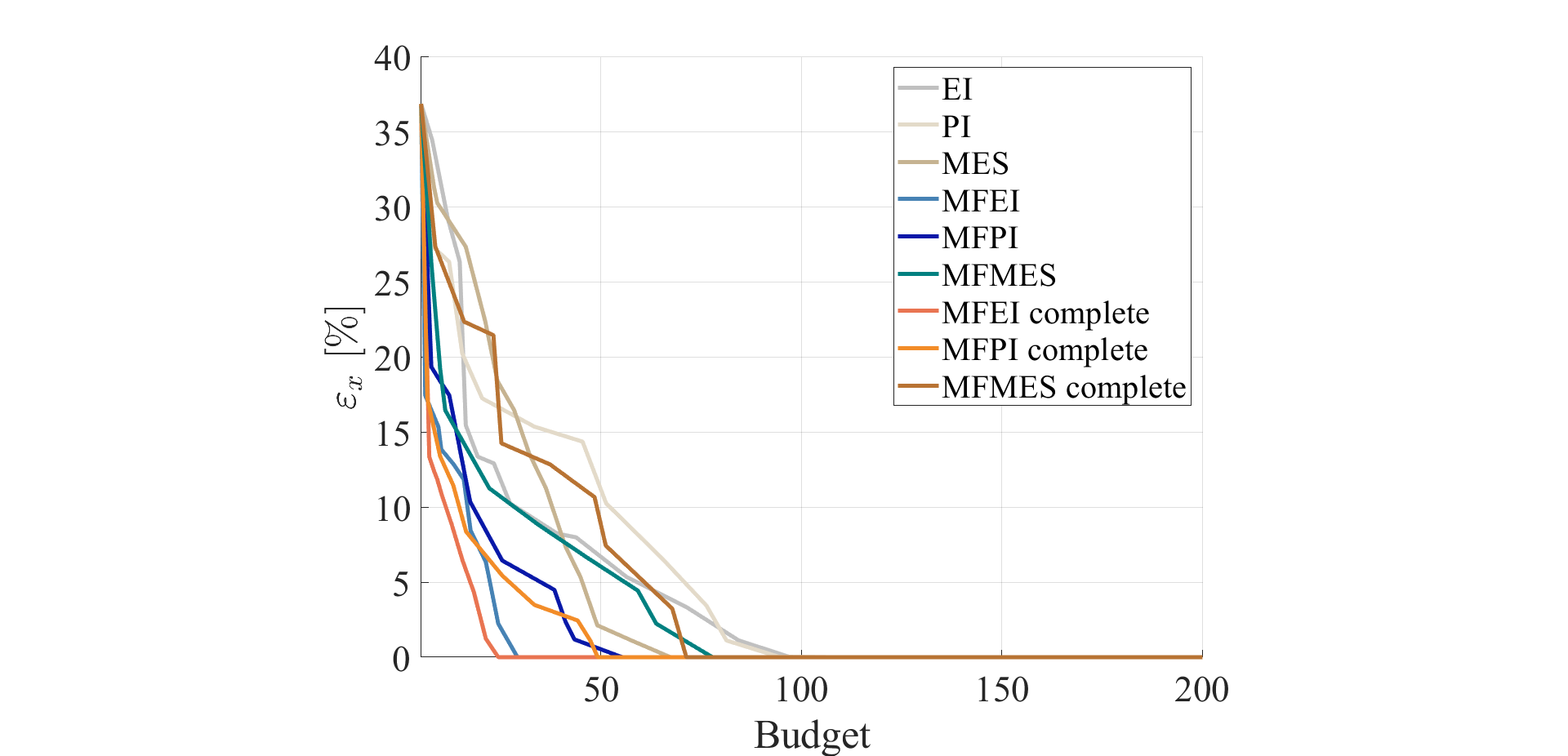} 
        \label{fig:rasx}}
        \subfigure[Spring-Mass System]{%
        \includegraphics[width=0.3\linewidth,trim=220 0 245 0, clip]{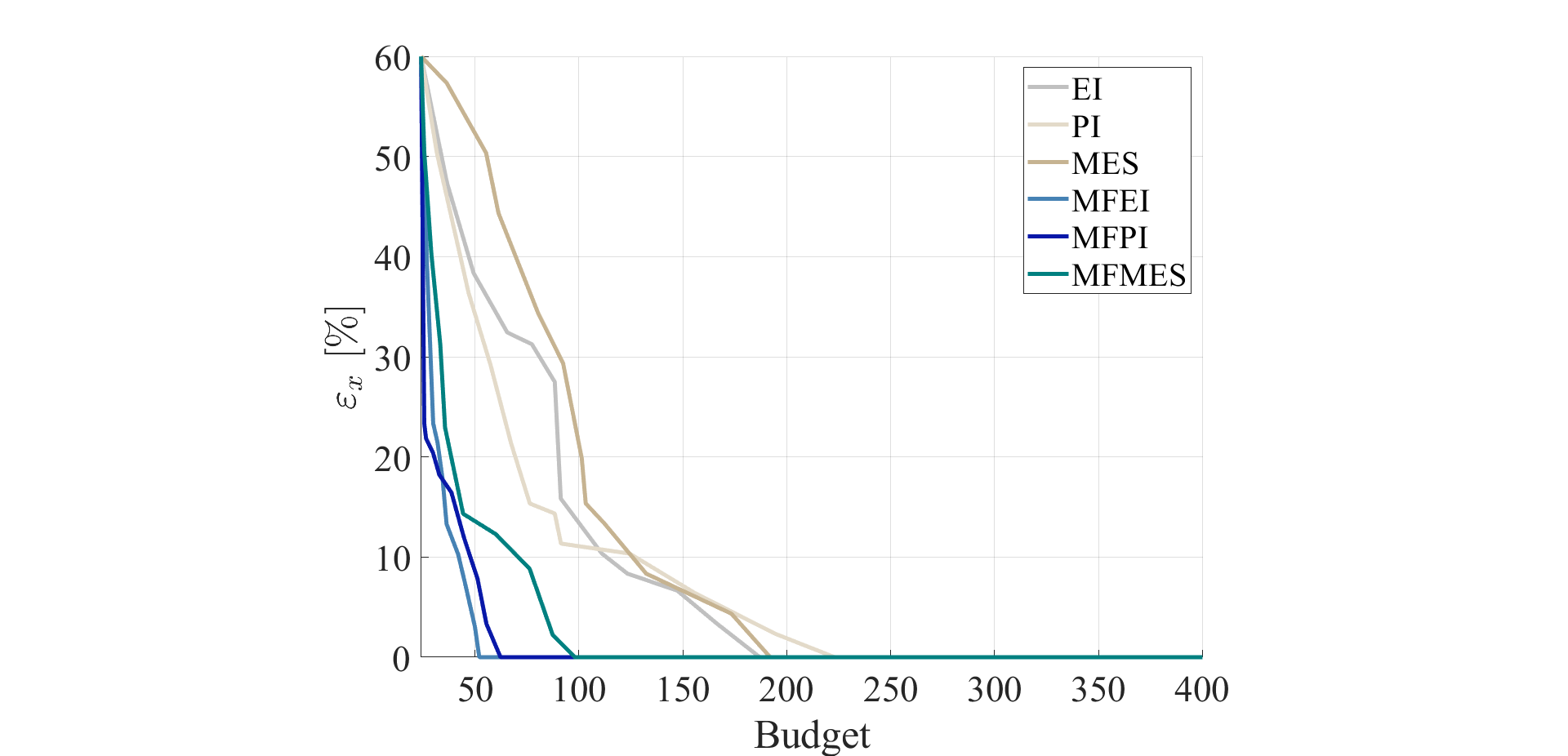}
        \label{fig:mk4x}}
        \subfigure[Paciorek]{%
        \includegraphics[width=0.3\linewidth,trim=220 0 245 0, clip]{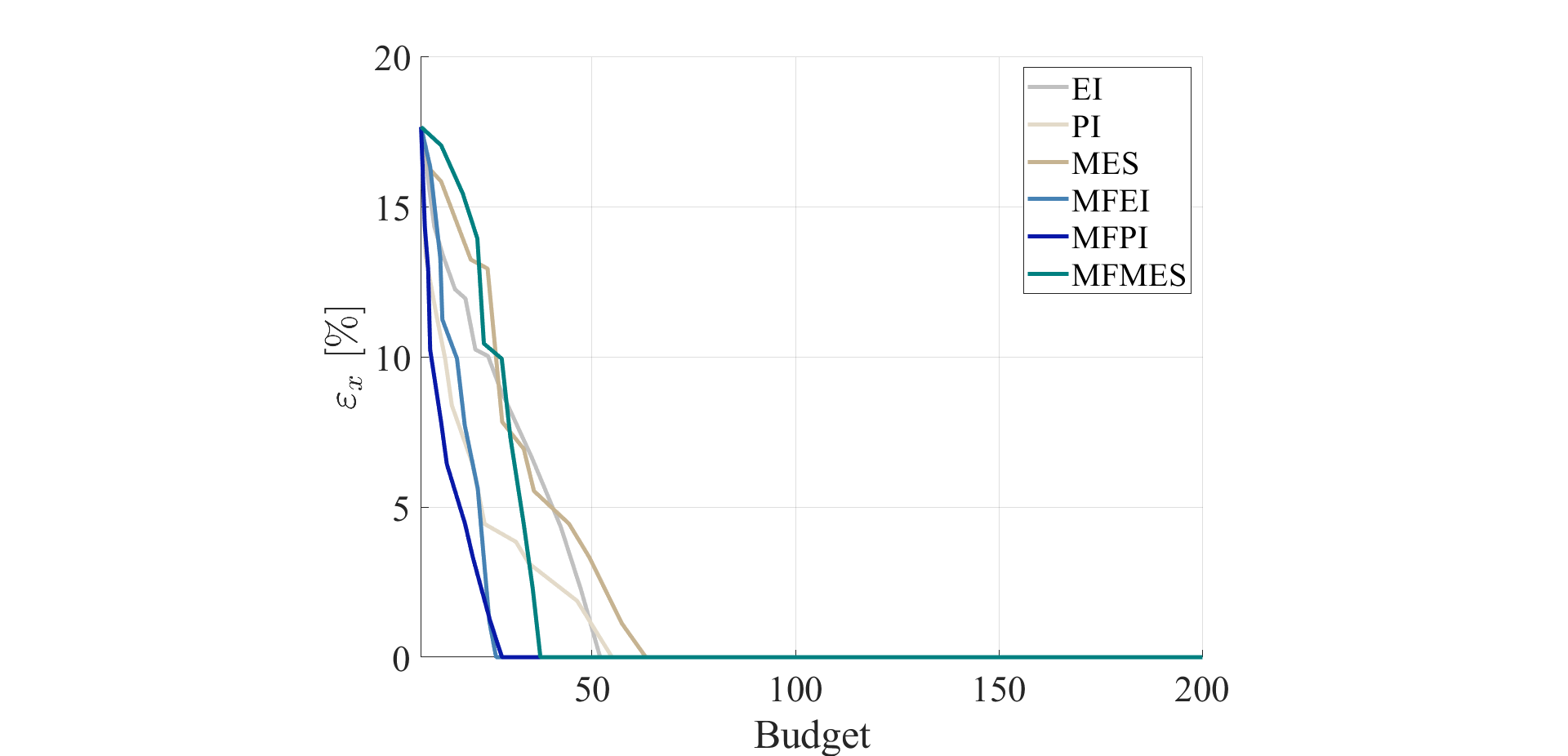}
        \label{fig:pacx}}
        
    \caption{Performances of the competing algorithms for the multimodal benchmarks.}\label{fig:multimodalresults}
\end{figure*}

The outcomes related to the multimodal benchmarks are reported in Figure \ref{fig:multimodalresults}. The multifidelity algorithms are capable of converging toward the analytical optimum with a fraction of the computational cost, if compared with the single-fidelity results. For the Rastrigin function (Figure \ref{fig:rasf} and Figure \ref{fig:rasx}), the multifidelity methods implementing all the levels of fidelity $\LevFid=1,2,3$ outperform the multifidelity methods with $\LevFid=1,3$: the intermediate level of fidelity $\LevFid=2$ is more accurate if compared with the low-fidelity output $\LevFid=3$ and allows to improve the reliability of the Gaussian process in presence of a strong multimodal behaviour. The best performing method is MFPI using $\LevFid=1,2,3$ denoting that the over-exploitation of the input space with lower-fidelity levels $\LevFid=2,3$ allows to take full advantage from low-fidelity data, improving the performance of the learning process. In contrast, we observe that the MES algorithm exhibits a more efficient convergence of the MFMES counterpart. This is related to the already noticed over-exploration of the domain: the MES uses accurate high-fidelity observations to refine the surrogate during the exploration, while the MFMES systematically adopts lower-levels of fidelity to massively query the domain and retard the exploitation with more accurate information sources. The results achieved for the mass spring benchmark problem (Figure \ref{fig:mk4f} and Figure \ref{fig:mk4x}) confirm the superior convergence performance of the multifidelity algorithms in presence of marked multimodal objective functions. In particular, the balance between exploration and exploitation delivered by the MFEI allows for superior accelerations and contained demand for computational resources. Similar results can be observed for the Paciorek benchmark problem (Figure \ref{fig:pacf} and Figure \ref{fig:pacx}): the multifidelity learning delivers efficient optimization procedures even in the simultaneous presence of multi-modality and noise. It should be noticed that in presence of noise both the MES and MFMES show an attenuation of the exploratory behaviour and a greater exploitation of the domain. This result is in agreement with what observed by Nguyen et al. \cite{NguyenAl2022}. The overall outcomes for this subset of benchmark functions demonstrate that a learning scheme characterized by a balanced exploration and exploitation phase is essential in presence of multimodal behaviour and noise in the measurements of the objective function.

\subsection{Advice on using Learning Criteria}

Throughout the experiments in this paper and in our research experience, we can summarize several recommendations that are intended to provide a guideline to apply the different learning criteria in real-world optimization problems. Although these advice may not be suitable in general due to the vast and natural heterogeneity of the applications where optimization is relevant, we believe that these guidelines can be useful in directing researchers toward the effective use of learning schemes.

\begin{enumerate}

    \item Pure exploitation/informativeness learning schemes could be potentially beneficial for low-dimensional optimization problems. In our experience, the direct exploitation of data at the beginning of the optimization procedure can produce significant improvement in the solution with relatively contained computational resources. The reason behind this behaviour is due to the accurate prediction of the emulator with contained amount of data in low-dimensional domains. This contributes to better inform the learner and effectively direct resources toward the optimum.

    \item Pure exploration/representativeness-diversity could impact considerably the optimization results for high-dimensional optimization problems. The exploration reduces the uncertainty of the emulator over all the domain and leads to a more reliable predictive framework. This would better inform the learner and help directing the computational resources in regions of the domain where is more likely to achieve benefits in terms of solution.

    \item The balance between exploration and exploitation guarantees consistent and satisfactory optimization performances over different mathematical properties of the objective function. In particular, our experiments suggest that pursuing the trade-off between exploration and exploitation often leads to satisfactory and in many cases better performance than implementing the learning criteria individually. Although the well performing behaviour in general, it should be privileged mainly in cases when there is no prior knowledge about the specific optimization problem considered to increase the chances of success.  

    \item When the computational resources are severely limited -- e.g. engineering preliminary design phases or trade-off analysis --, there is a clear advantage of using multifidelity learning criteria and leverage a spectrum of information sources at different levels of fidelity. Indeed, the wise combination of fast low-fidelity data with expensive high-fidelity evaluations reduces the overall demand for computational resources, and shows more robust performance for challenging mathematical properties of the objective function such as local/global behaviours, non-linearities and discontinuities, multimodality, and noisy measurements. 

\end{enumerate}

\section{Concluding Remarks}
\label{s:concluding remarks}

This paper proposes an original unified perspective of Bayesian optimization and active learning as adaptive sampling schemes guided by common learning principles toward a given optimization goal. Our arguments are based on the recognition of Bayesian optimization and active learning as goal-driven learning procedures characterized by the mutual information exchange between the learner and the surrogate model: the learner makes a decision based on the surrogate information to maximize the sampling utility with respect to the given goal, while the emulator is constantly updated through the results of this decision. Accordingly, we clarify and support our discussion through a general classification of adaptive sampling methodologies, and recognize Bayesian optimization as the logic intersection between active learning and adaptive sampling. This lays the foundations for the explicit formalization of the synergy between Bayesian optimization and active learning considering both a single information source and when a library of representations at different levels of fidelity is available to the learner. This unified perspective is based on the dualism between the active learning criteria of informativeness and representativeness/diversity, and the Bayesian infill criteria of exploration and exploitation as the driving elements to achieve the learning goal. To support our perspective, we reviewed and analysed popular formulations of the acquisition function for Bayesian optimization considering both single-fidelity and multifidelity settings. Accordingly, we formalize this synergy mapping the informativeness learning criterion with the exploitation infill criterion as driving components that direct the selection of samples toward the learning goal. Similarly, we formulate the substantial analogy between representativeness-diversity learning criterion and the exploration infill criterion as sampling policies that improve the awareness about the objective function over the domain. Through stressfull analytical benchmark problems, the authors demonstrate the benefits of each learning/infill criteria over challenging mathematical properties of the objective function typically encountered in real-world applications. The results reveal that the balance between the learning/infill criteria ensures good performances and computational efficiency over all the benchmark problems. In addition, multifidelity learning schemes deliver significant accelerations of the learning procedure making them particularly attractive when the available computational resources are limited. The authors also include some advice and guidelines on the use of the different learning criteria based on the experimental results and their own experience in the field.

\section*{References}
\renewcommand{\bibsection}{}
\bibliographystyle{unsrt}
\bibliography{references}

\end{document}